\documentclass[preprint,12pt]{elsarticle}

\usepackage{amssymb}

\usepackage{xcolor}
\usepackage{subcaption}
\usepackage{amsmath}
\usepackage{comment}
\usepackage{orcidlink}
\usepackage{multirow}
\usepackage{makecell}

\usepackage[absolute,overlay]{textpos}
\usepackage{atbegshi}

\AtBeginShipout{%
  \ifnum\value{page}>0 
  \ifnum\value{page}<2
    \AtBeginShipoutAddToBox{%
      \begin{textblock*}{\textwidth}(6cm,2cm)
        \textit{Preprint submitted to Array. January 29, 2026}
      \end{textblock*}
    }%
  \fi
  \fi
}

\journal{Array}

\begin{document}

\begin{frontmatter}



\title{Visual Localization via Semantic Structures in Autonomous Photovoltaic Power Plant Inspection}

\newcommand{\orcidauthorA}{0000-0001-8405-269X} 
\newcommand{\orcidauthorB}{0000-0002-6362-4254} 
\newcommand{\orcidauthorD}{0000-0002-0997-5889} 




\author[label1,label2]{Viktor Kozák\corref{cor1}\orcidlink{\orcidauthorA}}
\cortext[cor1]{Corresponding author.
}
\ead{viktor.kozak@cvut.cz}
\author[label1]{Karel Košnar\orcidlink{\orcidauthorB}}
\author[label1]{Jan Chudoba}
\author[label1]{Miroslav Kulich\orcidlink{\orcidauthorD}}
\author[label1]{Libor Přeučil}

\affiliation[label1]{organization={Czech Institute of Informatics, Robotics, and Cybernetics, Czech Technical University in Prague},
             addressline={Jugoslávských partyzánů 1580/3},
             city={Prague 6},
             postcode={160 00},
            country={Czech Republic}}

\affiliation[label2]{organization={Department of Cybernetics, Faculty of Electrical Engineering, Czech Technical University in Prague},
            addressline={Karlovo náměstí 13},
            city={Prague 2},
            postcode={121 35},
            country={Czech Republic}}

\begin{abstract}

Inspection systems utilizing unmanned aerial vehicles (UAVs) equipped with thermal cameras are increasingly popular for the maintenance of photovoltaic (PV) power plants. However, automation of the inspection task is a challenging problem as it requires precise navigation to capture images from optimal distances and viewing angles.

This paper presents a novel localization pipeline that directly integrates PV module detection with UAV navigation, allowing precise positioning during inspection. The detections are used to identify the power plant structures in the image. These are associated with the power plant model and used to infer the UAV’s position relative to the inspected PV installation. 
We define visually recognizable anchor points for the initial association and use object tracking to discern global associations. 
Additionally, we present three different methods for visual segmentation of PV modules and evaluate their performance in relation to the proposed localization pipeline. 

The presented methods were verified and evaluated using custom aerial inspection data sets, demonstrating their robustness and applicability for real-time navigation. Additionally, we evaluate the influence of the power plant model's precision on the localization methods.

\end{abstract}

\begin{graphicalabstract}
\includegraphics[width=1\textwidth]{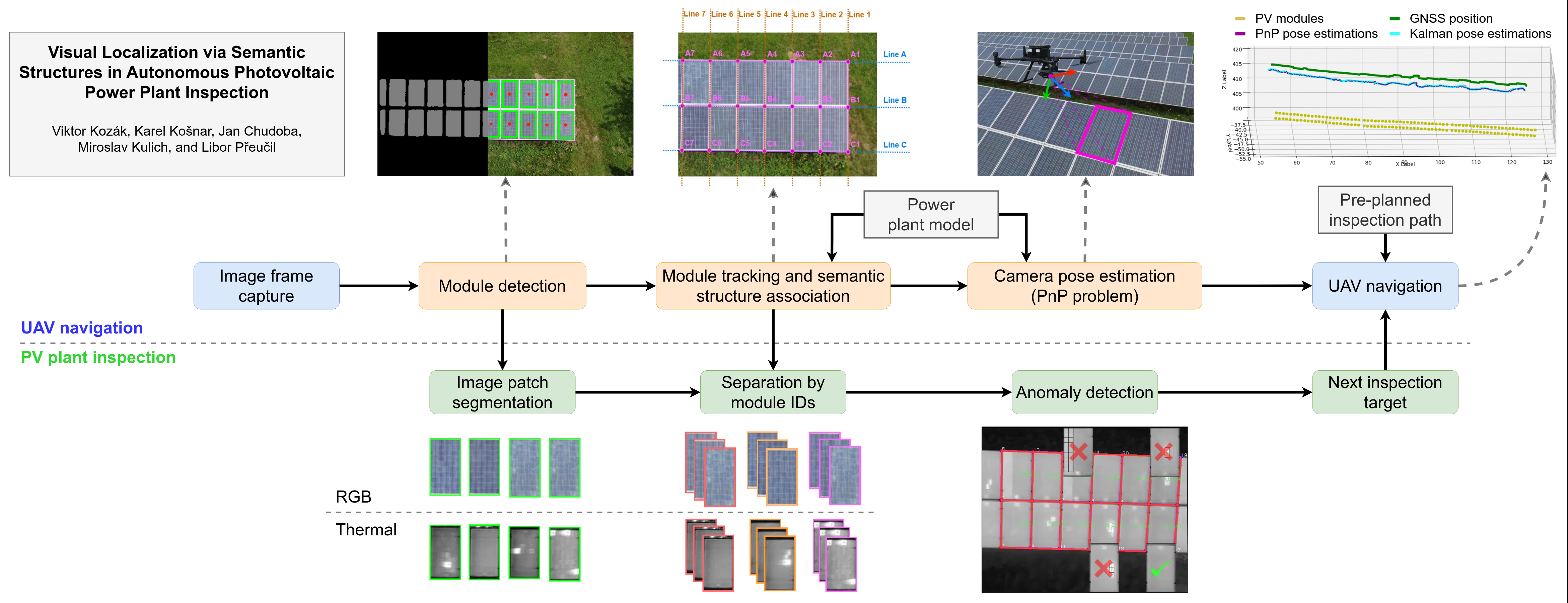}
\end{graphicalabstract}

\begin{highlights}
\item Direct integration of photovoltaic module detections into the localization method
\item Real-time 6-DoF localization relative to the inspected power plant
\item Precise aerial navigation and data acquisition from optimal viewing positions
\item Simultaneous extraction of image patches used for thermal defectoscopy
\end{highlights}


\end{frontmatter}








\section{Introduction}
\label{introduction}

The increasing deployment of photovoltaic (PV) power plants in last decades has caused a great demand for economically viable maintenance procedures. Standard PV modules have a limited lifespan and are prone to defects, which can greatly reduce their power output or even pose safety hazards. Such defects may occur during manufacturing or develop due to aging, incorrect handling, or environmental influences \citep{Mastny2015}. Thus, a regular inspection of PV plants is essential.

The task involves collecting operational data and periodic acquisition of images with individual PV modules. These are essential for evaluating the condition, performance, and faults of the system. With the increasing size of PV power plants, the need to automate the inspection procedure is paramount.

Systems that use unmanned aerial vehicles (UAVs) equipped with thermal infrared (IR) cameras have gained significant popularity in recent years. The use of UAVs can greatly reduce the necessary time and cost of the inspection procedure, and the acquired thermal images allow us to detect anomalies such as open or short circuits and hot spots. However, a fully autonomous UAV inspection of PV plants is a complex task that generally includes mission and path planning, robust control, and navigation. 
It has been the topic of many recent works concerning the challenges involved in the inspection process, the economical analysis, and the available camera and drone technologies \citep{Kumar2018, OverviewOnUAVPVInspection2021}. The main focus lies on computer vision techniques for autonomous detection and localization of PV modules, their registration and segmentation from IR images, and subsequent anomaly detection and analysis. The effort resulted in inspection systems with varying degrees of autonomy. 

A comprehensive review of the studied techniques and developed systems can be found in \citep{BommesPVinspectionReview2022, AutomaticPVinspectionReview2022, MICHAIL2024}. 
A significant number of developed systems and available market software packages are based on automatic mission control, where the UAV flies over a set of waypoints that cover all modules of the PV plant. These systems generally rely on the use of GNSS (global navigation satellite system) and IMU (inertial measurement unit) sensors and are prone to measurement errors as no additional visual cues are considered. 
In order to acquire detailed thermal images suitable for anomaly detection, the data must be acquired under an optimal viewing angle and distance \citep{PVResolution2018}. This requires high precision of both the navigation system and the power plant model. Since these cannot be easily achieved (or guaranteed) even with current top-shelf drone technologies, there is the necessity of incorporating camera feedback into the navigation system. Although several systems incorporate PV plant segmentation or line tracking into their control algorithms, 
most of the functionality is limited to flight path correction along a row of PV panels. 
In contrast, the localization method proposed in this paper directly produces a 6-DoF position estimate relative to the inspected power plant installation.

\subsection{Objectives and Contributions}

\begin{figure}[h!]
    \centering    
    \begin{subfigure}[t]{0.4\textwidth}
      \centering
      \includegraphics[width=1\textwidth]{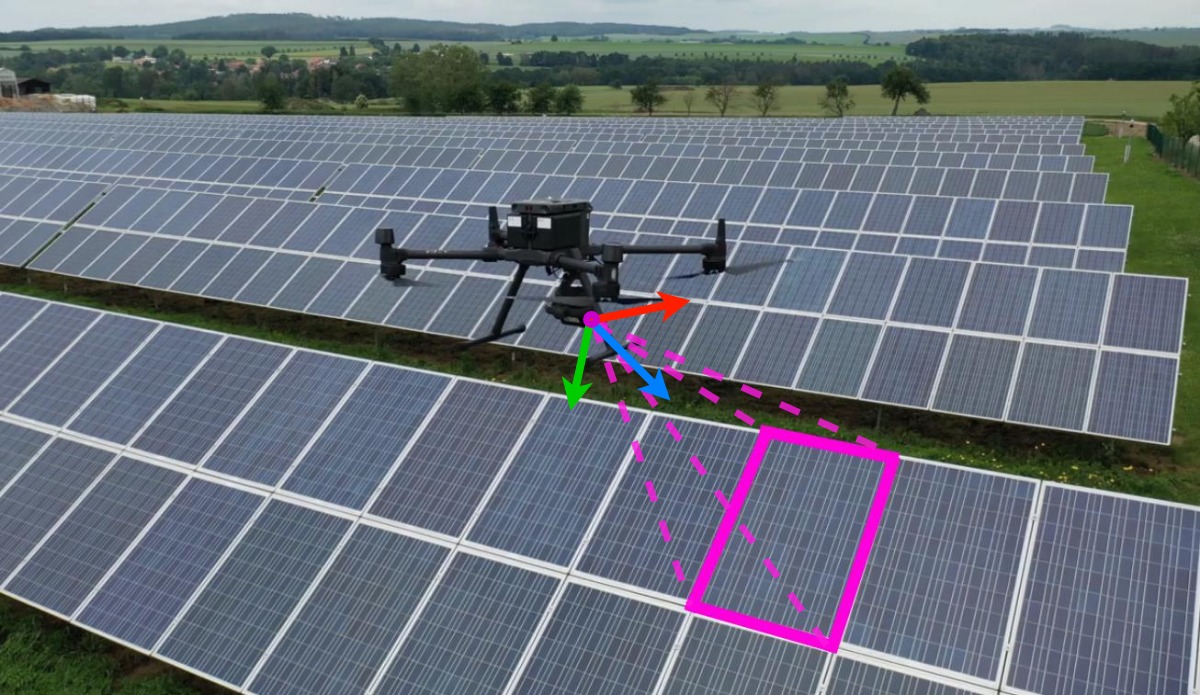}
      \caption{}
      \label{fig:drone_img}
    \end{subfigure}\hspace{1mm}%
    \begin{subfigure}[t]{0.58\textwidth}
      \centering
      \includegraphics[width=1\textwidth]{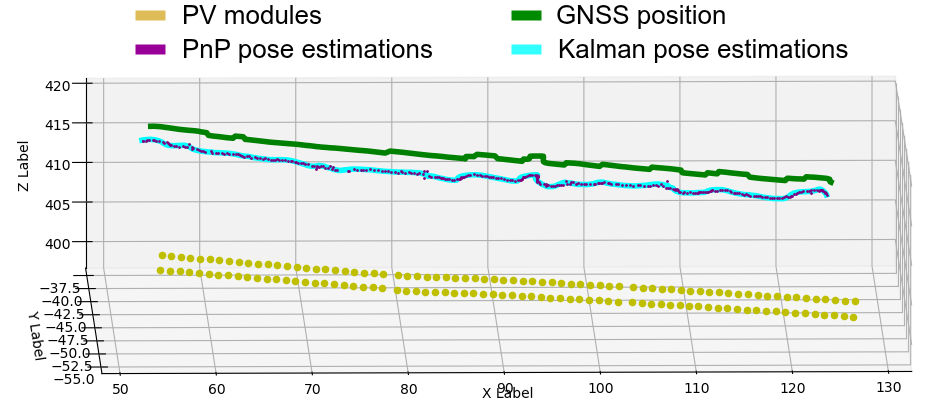}
      \caption{}
      \label{fig:kalman_3d_graph}
    \end{subfigure}

    \caption{\textbf{(a)} The DJI Matrice 300 drone during PV power plant inspection. A single PV module is highlighted in purple. \textbf{(b)} An example trajectory from an inspection flight depicting the position of individual PV modules, GNSS-based UAV position measurements, and UAV positions estimated by our proposed visual localization pipeline.}
    \label{fig:drone_pv_illustration}
\end{figure}

We present a localization pipeline that directly integrates the detection of PV modules to estimate the position of the drone (Figure \ref{fig:drone_pv_illustration}). 
The main contributions of our work are:

\begin{itemize} \setlength\itemsep{0em}
    \item A method for visual localization of UAVs capable of direct drone-to-module pose estimation. This allows for precise autonomous navigation and data acquisition from optimal viewing angles and distances. 
    \item The proposal and evaluation of several fundamentally different visual segmentation methods used for the detection of PV modules. All methods can be used for both localization and defectoscopy.
    \item  Introduction of visually recognizable anchor points, used to initialize navigation and associate detections with the power plant model.
    \item Experimental verification in different flight scenarios, demonstrating the robustness, precision, and real-time capabilities of the proposed methods.
\end{itemize}

The localization pipeline was tested on a custom dataset obtained from a manual aerial inspection of two PV power plants. We also created an annotated dataset for the segmentation of PV modules, and we have made both the training dataset and the trained segmentation models publicly available. 
In addition, we use two distinct approaches to the generation of compact power plant models and identify the limitations of the model and the localization method.


\subsection{Paper Outline}

The rest of this paper is organized as follows. Section \ref{rel_work} introduces related work. In Section \ref{methodology}, we present the proposed system and the methodology. Section \ref{results} describes the experimental setup and results, whereas Section \ref{conclusions} is devoted to conclusions and future work.

\section{Related Work}
\label{rel_work}

The use of UAVs has become increasingly popular in various applications, including PV power plant inspection and maintenance. 
While traditional UAV systems relying on GNSS and IMU sensors can provide adequate information for general navigation, they lack visual feedback, which limits their capabilities. 


Vision-based localization systems can provide a practical tool for precise navigation. 
However, most current vision-based methods rely on low-level visual features represented by
local descriptors, which generally provide limited contextual information \citep{Azzam2020FeaturebasedVS}. 
Practical object-oriented applications require the integration of high-level visual features (e.g., objects or structures) into the navigation pipeline. This can be achieved by direct estimation of the relative position of the detected object of interest or by directly incorporating such features into the navigation method \citep{semanticSLAMsurvey2020}. Another challenging aspect of visual navigation is long-term autonomy, as the environment can exhibit significant appearance changes caused by illumination or weather variations \citep{krajnik_longterm_survey2018}. Thus, there arises a need for robust feature detection and navigation algorithms.


\subsection{Aerial Photovoltaic Power Plant Inspection}

The technical and economic feasibility of drone-based inspection systems is analyzed in \citep{UseOfDronesPV2019}, the work estimates that the use of drones is cost-efficient for plants with an installed capacity of more than 50 MW. An overview of photovoltaic inspection using drones that emphasizes the importance of automating the inspection process to increase efficiency and reduce costs is provided in \citep{OverviewOnUAVPVInspection2021}. The study also points out the need for standardization of inspection procedures. 
A general overview of different technologies and sensors is provided in \citep{Kumar2018}.
A data management system for digital mapping and monitoring of PV power plants was presented in \citep{pvManagement2019}. Other studies on autonomous aerial monitoring of PV power plants address optimal planning of the UAV flight path and execution of the inspection task \citep{RoboPV2022}.

Thermal cameras have been widely used to detect temperature anomalies and identify defective PV cells and modules. Comprehensive reviews on PV power plant inspection based on aerial thermography are given in \citep{AutomaticPVinspectionReview2022, BommesPVinspectionReview2022, MICHAIL2024}. The inspection methodology presented in \citep{Lee2019} uses infrared sensors on board of UAVs to identify defective PV modules that are comparably hotter than other modules. A similar approach is adopted in \citep{PVFaultDetection2020}, where a combination of CNNs and support vector machines (SVMs) is used for fault detection. 
In \citep{sunmap2023} and \citep{cardoso2024}, SfM and multi-view stereo techniques were used to create terrain and surface models of the area and generate orthoimages. In both works, the detection of defective PV modules was performed on these orthoimages by applying different contour detection and thresholding operations. 
However, these systems rely on pre-planned flight paths at higher altitudes, they suffer from limited autonomy, and fail to acquire detailed thermal images.


A computer vision tool for the semi-automatic extraction of PV modules from thermographic UAV videos was presented in \citep{BommesPVHawk2021}. The tool is used to collect a large-scale dataset that contains 4.3 million IR images of 107,842 PV modules from thermographic videos of seven different PV plants. The dataset is used to train a deep convolutional neural network classifier ResNet-50 \citep{resnet50_2016} to detect and classify ten common module anomalies.  
In their other work \citep{BommesGeoreferencing2022}, the authors use Mask R-CNN detections \citep{maskRCNN2017} from infrared images for mapping and georeferencing of PV modules using structure-from-motion. However, the presented works do not address real-time UAV localization or autonomous navigation, as only videos captured from manual drone operation are processed.

Autonomous navigation was addressed in \citep{PVLineTracking2020} and \citep{PVLineFollowing2020}. Both works present line-tracking algorithms that rely on edge detection in RGB images for the segmentation of photovoltaic installations and principal line-based navigation. 
In \citep{drones6110347}, navigation was based on the estimation of lines that go through the middle of PV module rows detected by thermal and color thresholding. 
These works enable real-time flight trajectory correction along installations in row layouts commonly found in PV plants, without the adjustment of position or viewing angles in regards to individual PV modules.



\subsection{Visual Detection and Segmentation of Photovoltaic Modules}

Methods for PV module detection in images can generally be divided into techniques using traditional computer vision (CV) algorithms, and the increasingly popular deep learning based methods. Traditional CV techniques based on binary thresholding of image intensities are often used for PV module detection in both visual and thermography images \citep{PVThresholding2017, PVThresholding2018}. Other works are based on the use of edge detection, the Hough transform, and morphological operations \citep{PVMosaic2016, HougPVdetection2017}. A method based on the detection of rectangular candidate contours through thresholding and subsequent SVM classification of individual segments was presented in \citep{PVComplex2020}. A model-based approach using edge detectors and structural regularity of the PV infrastructures was presented in \citep{EdgeBasedPVSytem2020}. The method was tested on several large datasets and enables real-time performance. In \citep{EdgesAndPeaksSegmentation2021}, edge detections were used for boundary detection and subsequent use of peak identification on lines of pixels. 

Deep learning is increasingly common in practical applications; however, its use is highly dependent on available training data. This is especially burdensome in atypical domains, such as industrial applications, where the use of standard datasets or pre-trained network models is inherently difficult. A multi-resolution dataset for segmentation of PV installations in RGB images was presented in \citep{MultiresolutionSegmentation2021} together with the comparison of several CNN segmentation models. A combination of ResNet \citep{resnet50_2016} and U-Net \citep{unet2015} CNN models was presented in \citep{ResnetUnet2019}. However, both works address segmentation only on the level of PV installations, not individual modules. 
In \citep{BommesPVHawk2021}, a Mask R-CNN model \citep{maskRCNN2017} was trained on infrared images and successfully used for PV module detection on a large-scale dataset. A study presented in \citep{PVComplex2020} compares a traditional CV segmentation, deep learning detection (using Mask R-CNN), and the influence of a post-processing step in thermal images with complex backgrounds. Both approaches achieved comparable precision, further increased by the post-processing step. 
Several versions of the YOLO detector \citep{yolov8_ultralytics} were used for PV module detection in thermal images in \citep{PVYolo2020}.

The use of traditional CV-based methods is highly dependent on manual priors and parameter adjustments. Although it is possible to mitigate such shortcomings, these techniques often generalize poorly to changing conditions and unseen imagery. Conversely, it might be relatively easy for an expert to fine-tune the algorithms on limited data without the need for a new training dataset. In comparison, CNN-based techniques display greater robustness; however, this is dependent on the provision of a significantly large dataset. Traditional CV techniques can also be relatively fast and computationally inexpensive, without the need for dedicated hardware.

To the best of our knowledge, the method proposed in our paper is the first to directly utilize PV module detections for UAV localization at the level of individual modules and to provide 6-DoF position estimation. This enables optimal UAV positioning during inspections. Furthermore, we present three distinct segmentation methods, utilizing both traditional CV and CNN models, and compare their performance within the localization pipeline.

\section{Methodology}
\label{methodology}

Sections \ref{problem_definition} and \ref{pipeline_overview} provide an overview of the aerial power plant inspection task and the proposed UAV localization pipeline. The visual detection of PV modules is described in detail in Section \ref{pv_detection}. This is followed by Section \ref{pv_tracking} which describes the 
initial association of the detected modules with the power plant model and subsequent module tracking. Lastly, \mbox{Section \ref{uav_localization}} is focused on UAV pose estimation.

The methods and results presented in this work are focused on the use of RGB camera images, since these offer richer feature space, wider viewing angle, higher resolution, and exhibit lower variation of image intensities compared to thermal images. However, the presented methods can also be modified for use with thermal images (see \ref{apdx_thermal}) or paired with detections from RGB images through accurate temporal synchronization and spatial registration. 

\subsection{Power Plant Inspection}
\label{problem_definition}

The objective of the developed inspection system is both visual and thermographic defectoscopy of PV power plants. Power plant installations typically consist of long rows, each comprising multiple parallel lines of PV modules (Figure \ref{fig:drone_img}). 
The task of the UAV is to navigate alongside these rows to acquire detailed images of the PV modules.

An example of an inspection flight over a row containing two lines with PV modules can be seen in Figure \ref{fig:drone_pv_illustration}. During flight, it is necessary to maintain an ideal position of the drone relative to the PV installation, to capture the images from optimal distances and viewing angles \citep{sunmap2023}.

\begin{figure}[h!]
\centering
\begin{subfigure}[t]{.37\textwidth}
  \centering
  \includegraphics[width=1\textwidth]{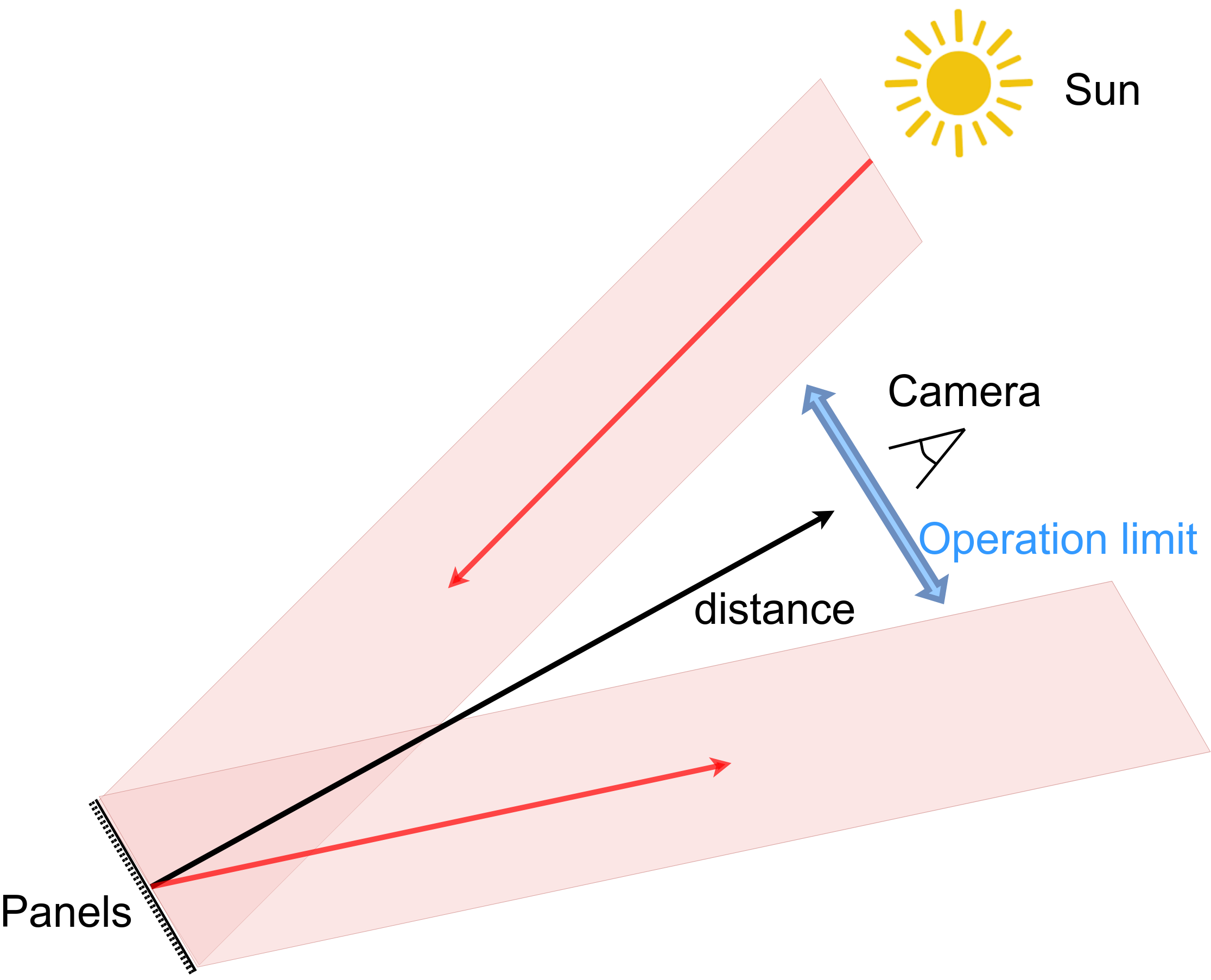}
\end{subfigure}\hspace{10mm}%
\begin{subfigure}[t]{.38\textwidth}
  \centering
  \includegraphics[width=1\textwidth]{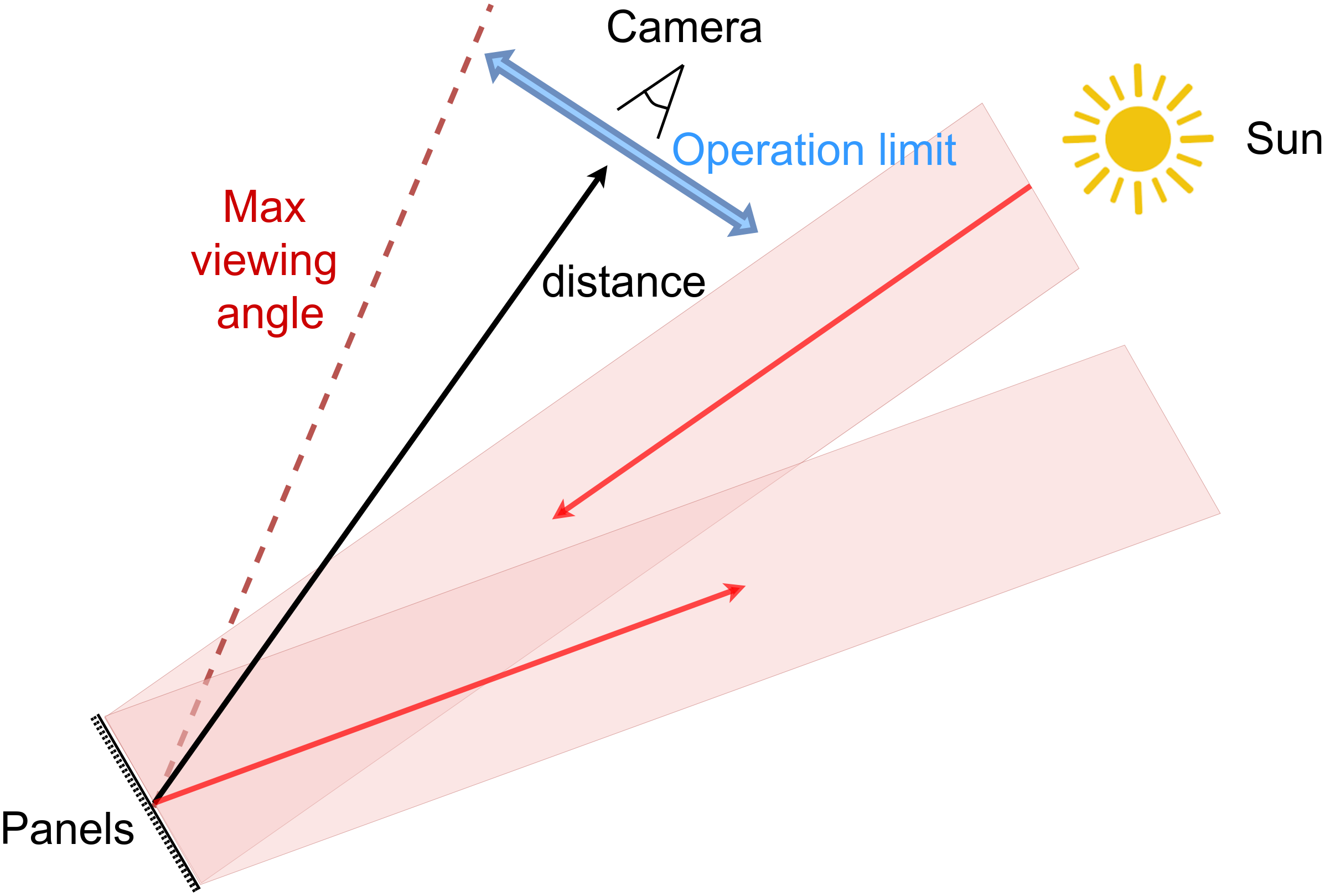}
\end{subfigure}

\caption{Influence of Sun reflection in varying daytime.}
\label{fig:sun_reflection_limits}
\end{figure}

The normal of the PV modules and the current position of the Sun should be considered during trajectory planning. 
The UAV must avoid solar reflection from the modules while also minimizing interference caused by its own shadow (Figure \ref{fig:sun_reflection_limits}). 
It is also limited by a maximum viewing angle. 
Detailed inspections are typically conducted at distances ranging from 10 to 30 meters from the PV modules. The operation area depends on the specific power plant, and the tolerance for position error increases at larger distances. Our proposed localization pipeline provides precise position information that supports navigation along the desired trajectories.

\subsection{Localization Pipeline Overview}
\label{pipeline_overview}

\begin{figure}[h!]
    \centering
    \includegraphics[width=1\textwidth]{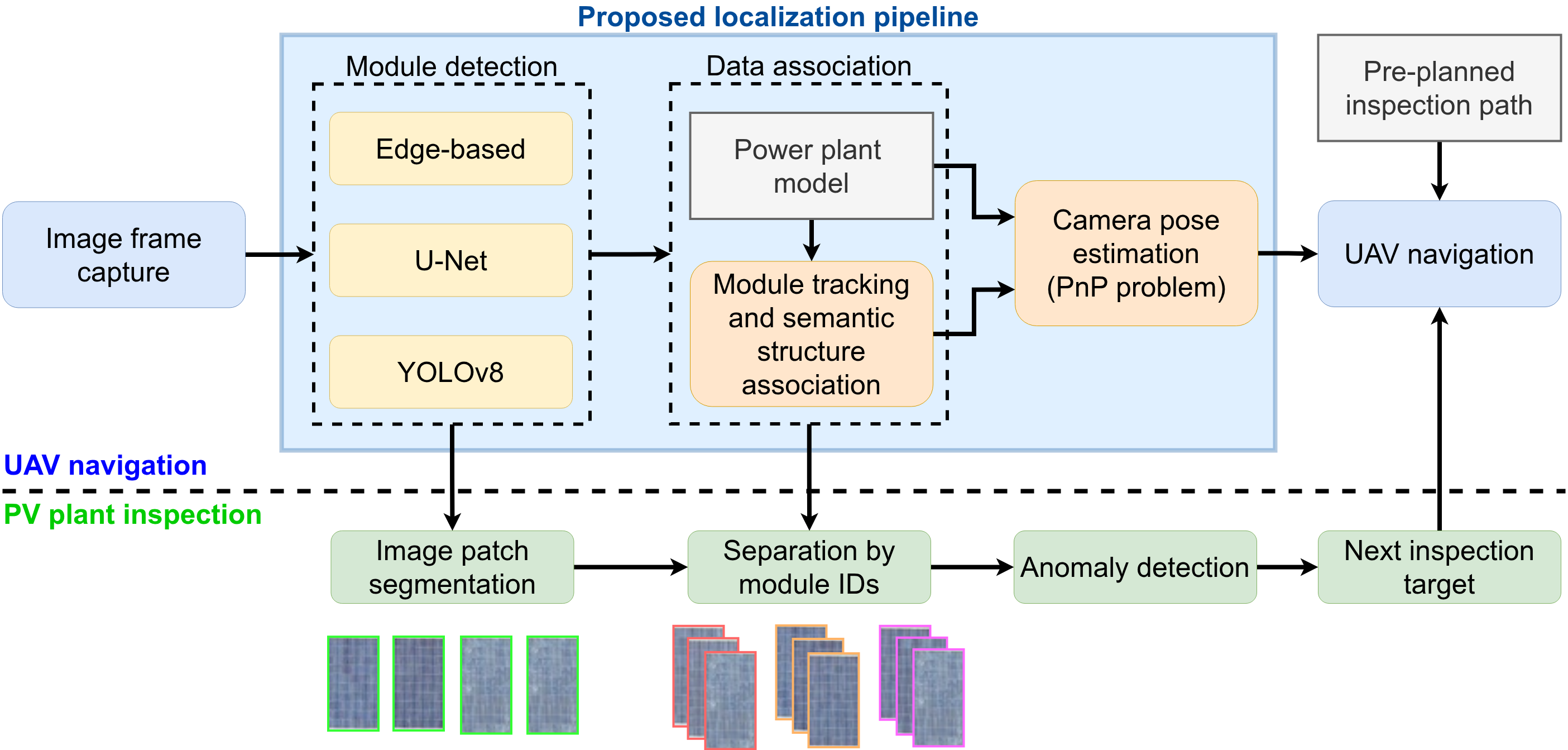}
    \caption{The proposed localization pipeline and its role within the developed inspection system. The pipeline supports different PV module detection methods (e.g., YOLO, \mbox{U-Net}).}    
    \label{fig:pipeline}
\end{figure}

The overview of the developed system is illustrated in Figure \ref{fig:pipeline}. 
The core of the system is the proposed localization method, in which individual PV module detections are associated with the model of the power plant and are used for UAV pose estimation. The pose estimation is then used for UAV navigation during the inspection. The detections can also be directly reused for defectoscopy, either for real-time anomaly detection or processed offline after inspection.

The visual localization pipeline can be divided into 3 main steps:

\begin{itemize} \setlength\itemsep{0em}
    \item \textbf{Module detection:} \\
    Visual detection and segmentation of individual PV modules.
    \item \textbf{Data association:} \\    
    Individual module detections are tracked in an image sequence from the camera stream, preserving unique detection IDs. 
    The detections are used to infer semantic structures in the image and associate these with the power plant model.
    \item \textbf{Camera pose estimation:} \\
    The association of detected structures with the power plant model and its real-word coordinates is used to estimate the relative position of the camera by solving the perspective-n-point problem.
\end{itemize}

The pipeline requires a georeferenced model with the positions, orientations, and dimensions of individual PV modules. 
Since the appearance of each module is nearly identical, we define a set of visually recognizable anchor points that are used to initialize the localization pipeline (Section \ref{pv_tracking}). A typical example of an anchor point is the beginning of a row with PV modules. The inspection process begins with a GNSS-based flight above the selected anchor point and its visual confirmation. This is followed by initial detection of the PV modules and their association with the power plant model.

\subsection{PV Module Detection and Segmentation}
\label{pv_detection}

The presented segmentation methods are designed to detect significant points and structures that can be linked to the model of the power plant and associated with their respective real-world coordinates. 
This general representation enables the use of different methods and reference points in the localization pipeline. 
We present three different segmentation approaches, in order to properly evaluate the capabilities of the proposed pipeline. These differ in their prerequisites, applicability, parameter tuning, computational requirements, and the resulting detection representations. Each of these is used and evaluated independently within the localization pipeline.


\begin{figure}[thpb]
\centering
\begin{subfigure}[t]{0.47\textwidth}
  \centering
  \includegraphics[width=1\textwidth]{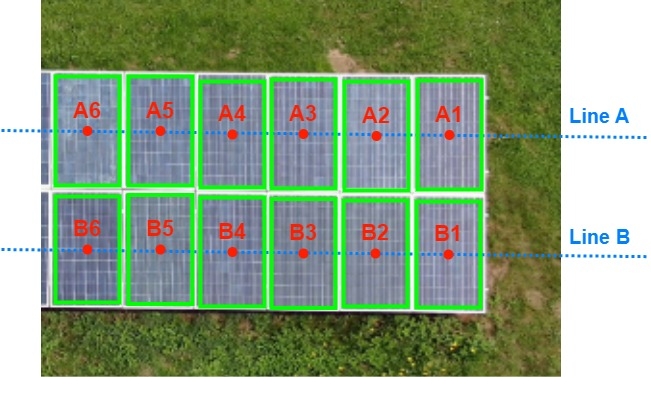}
  \caption{}
  \label{fig:structure_CNN0}
\end{subfigure}\hspace{6mm}%
\begin{subfigure}[t]{0.47\textwidth}
  \centering
  \includegraphics[width=1\textwidth]{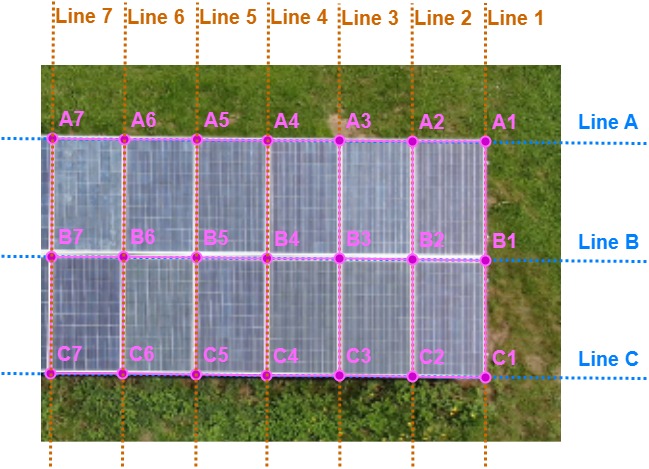}
  \caption{}
  \label{fig:structure_Edge0}
\end{subfigure}

\caption{Semantic structure illustrations. \textbf{(a)} CNN-based PV module detections, aligned to their respective rows (lines A and B). The same representation can
be used for both YOLO and U-Net. \textbf{(b)} Edge-based line intersection graph.}
\label{fig:semantic_structure0}
\end{figure}

The first two methods are based on convolutional neural networks (CNNs). We employ the YOLOv8 \citep{yolov8_ultralytics} model designed for instance segmentation, and we propose a new method using the U-Net model \citep{unet2015}. Both methods produce a set of bounding boxes, which can be used to determine the centers of their respective PV modules (\mbox{Figure \ref{fig:structure_CNN0}}).

The third method introduces a novel PV module segmentation algorithm based on edge detection and traditional CV. The output of the edge-based method (Figure \ref{fig:structure_Edge0}) is a graph structure created from the intersections of the detected lines. These represent the corners of their associated PV modules.

\subsubsection{CNN-Based PV Module Detection Methods}
\label{CNN_based}

The presented CNN-based methods take an RGB image as input and return minimum bounding rectangles (oriented bounding boxes) as output. The inference of semantic structures from the detected bounding boxes is identical for both methods. For this reason, we describe the YOLO and U-Net segmentation methods together.

\vspace{0.3cm}
\textit{YOLOv8 detection}

\begin{figure}[thpb]
\centering
\begin{subfigure}[t]{.45\textwidth}
  \centering
  \includegraphics[width=1\textwidth]{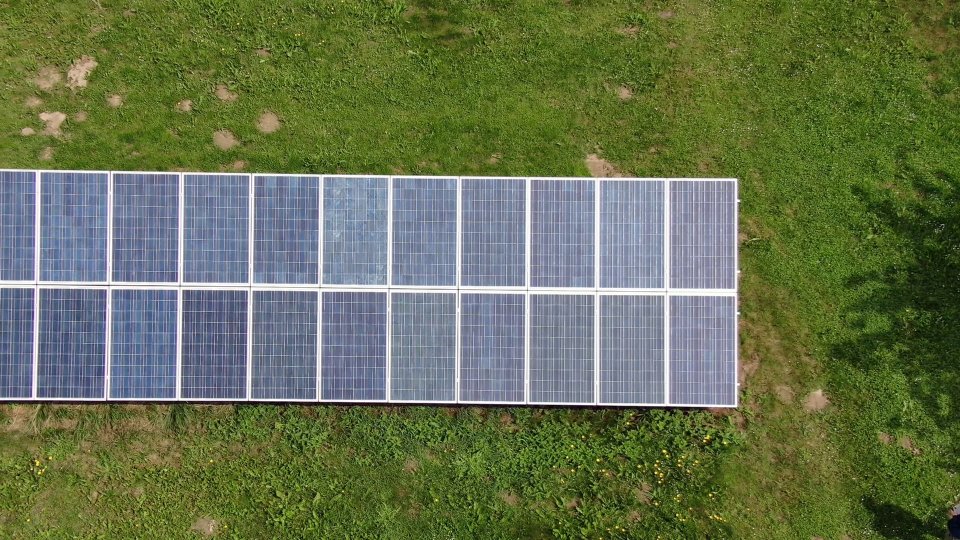}
  \caption{}
  \label{fig:Ydet0}
\end{subfigure}\hspace{3mm}%
\begin{subfigure}[t]{.45\textwidth}
  \centering
  \includegraphics[width=1\textwidth]{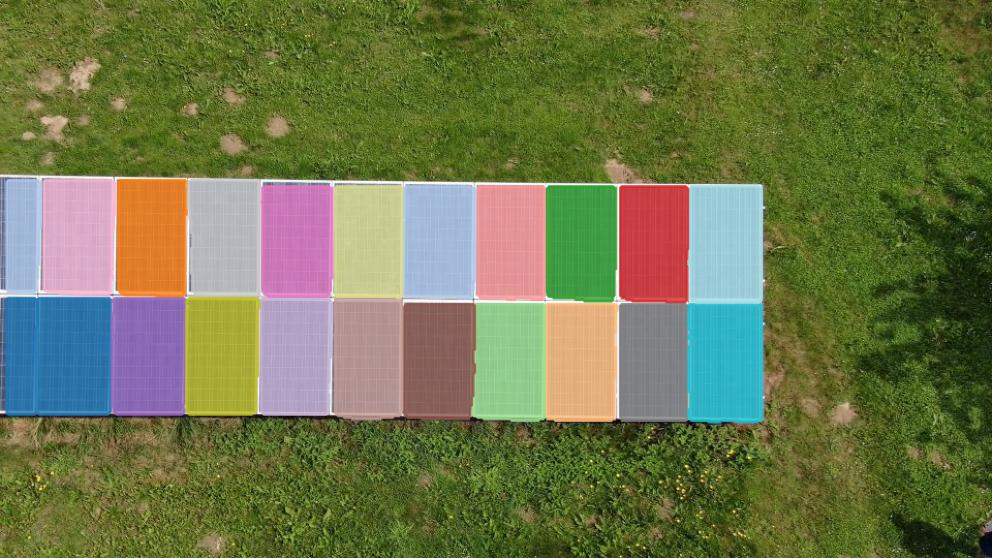}
  \caption{}
  \label{fig:Ydet2}
\end{subfigure}%

\vskip\floatsep%
\vspace{-0.1cm}

\begin{subfigure}[t]{.45\textwidth}
  \centering
  \includegraphics[width=1\textwidth]{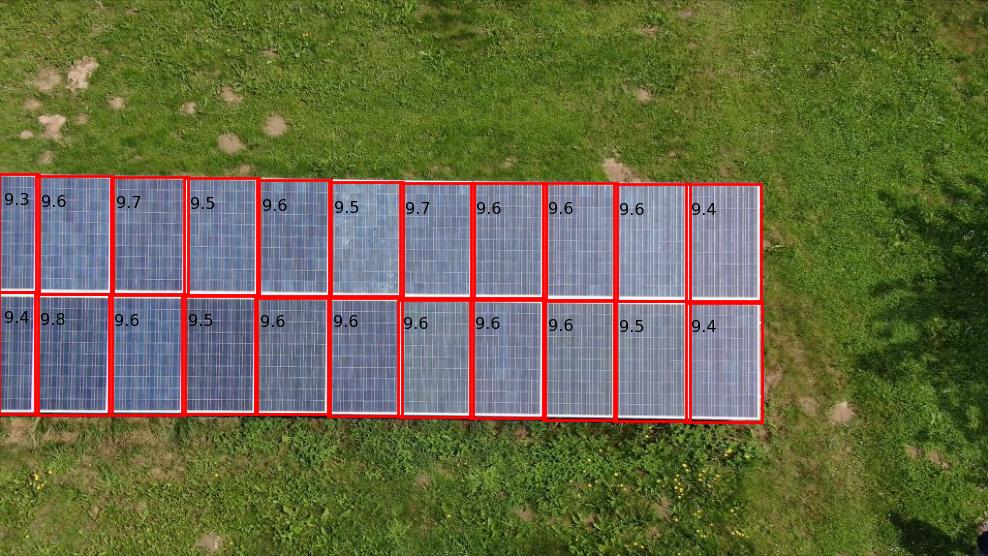}
  \caption{}
  \label{fig:Ydet1}
\end{subfigure}

\caption{YOLO PV module detection and segmentation. \textbf{(a)} Original image. \textbf{(b)} Mask instance segmentation. \textbf{(c)} Bounding boxes with PV module detections.}
\label{fig:yolo_segmentation}
\end{figure}

The YOLOv8 model \citep{yolov8_ultralytics} is designed for instance segmentation. The model outputs bounding boxes and segmentation masks with detected PV modules (Figure \ref{fig:yolo_segmentation}). The masks are given as a set of contour points. We compute a convex hull for each of the detected objects and calculate bounding rectangles aligned with each of the edges of the convex hull. The rectangle with the minimum area is determined to be the minimum bounding rectangle.

The resulting oriented bounding box representation is used in the proposed localization pipeline. 
Its parameters are given in Equation \ref{eq:b_boxe}, where $x_c$ and $y_c$ are the pixel coordinates of its center, $w$ and $h$ are its width and height in pixels, and $\alpha$ represents the orientation.

\begin{equation}
\label{eq:b_boxe}
    obb = \begin{bmatrix}
     x_c & y_c & w & h & \alpha
    \end{bmatrix}
\end{equation}

\vspace{0.3cm}
\textit{U-Net detection}

U-Net \citep{unet2015} is a scalable Fully Convolutional Network (FCN). Its architecture consists of a contracting encoder network used to capture context and a symmetric expanding decoder network that allows pixel-wise segmentation. The use of FCNs enables segmentation maps to be translated directly into output images of any size. The architecture also exhibits faster performance than patch classification approaches.

\begin{figure}[h!]
\centering
\begin{subfigure}[t]{.48\textwidth}
  \centering
  \includegraphics[width=1\textwidth]{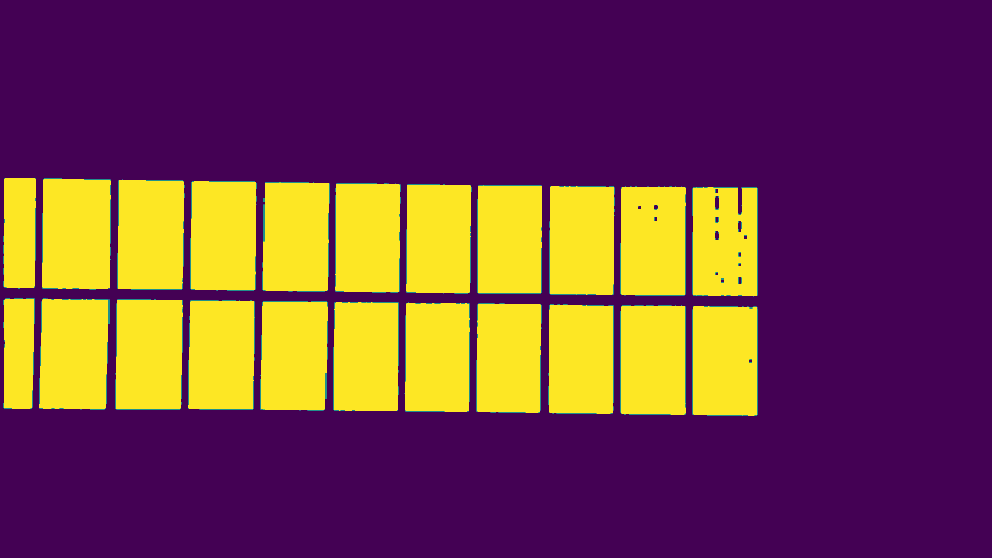}
  \caption{}
  \label{fig:det1}
\end{subfigure}\hspace{3mm}%
\begin{subfigure}[t]{.48\textwidth}
  \centering
  \includegraphics[width=1\textwidth]{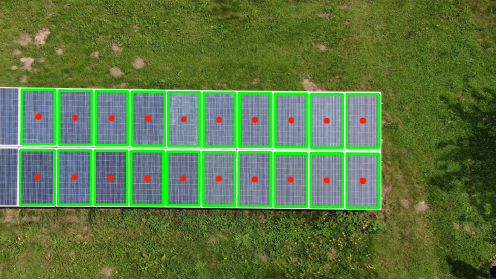}
  \caption{}
  \label{fig:det4}
\end{subfigure}

\caption{U-Net PV module detection and segmentation. \textbf{(a)} Mask segmentation. \textbf{(b)} Bounding boxes with PV module detections.}
\label{fig:Ucnn_segmentation}
\end{figure}

The U-Net model outputs a binary segmentation image mask (Figure \ref{fig:det1}), which has to be processed to achieve instance segmentation. We use erosion to eliminate undesirable false detections, which occasionally occur in the form of minute false positive segments. This also increases margins between masks of individual modules and prevents situations where the masks of two modules are connected by stray pixels. Subsequent dilation is performed to restore the original size of the detected masks. 
Finally, we extract individual image components and find their minimum bounding rectangles (Figure \ref{fig:det4}).

\vspace{0.3cm}
\textit{Inference of semantic structure from bounding box detections}

An additional filtering procedure is performed for both the YOLO and the U-Net bounding box detections. Detections on the edge of the image are discarded to remove incomplete PV modules. Detections with more than a 20\% overlap are also discarded.

A representative bounding box is selected from the 75th percentile of the detections sorted by size, favoring larger bounding boxes to maintain focus on the primary objects of interest. 
Bounding boxes with abnormal dimensions are filtered using the height and width of the representative bounding box and an acceptable size divergence value. 
An additional check 
can be performed using the known approximate distance from the camera on the pre-planned flight path.  

\begin{figure}[thpb]
\centering
\begin{subfigure}[t]{0.48\textwidth}
  \centering
  \includegraphics[width=1\textwidth]{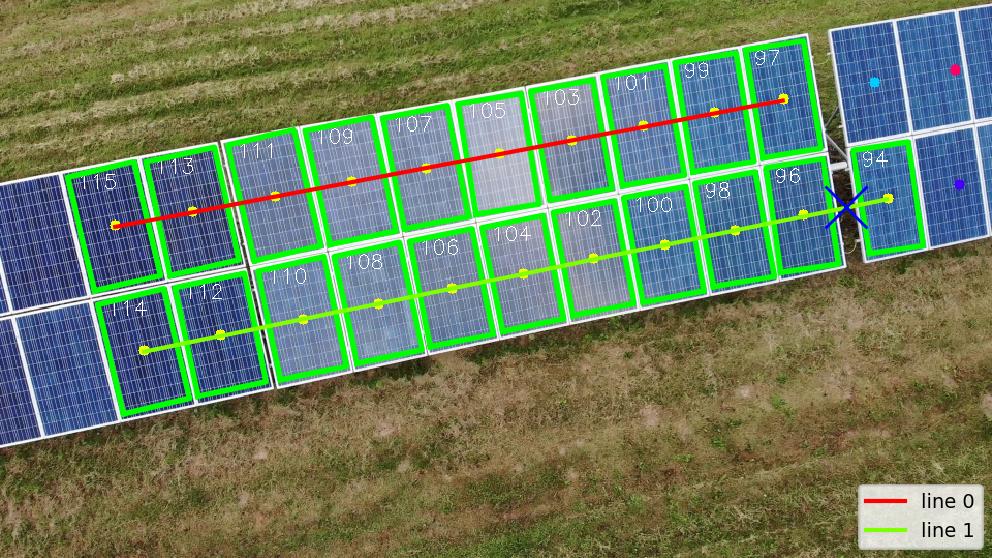}
  \caption{}
  \label{fig:cnn_structure_PPA2}
\end{subfigure}\hspace{2mm}%
\begin{subfigure}[t]{0.48\textwidth}
  \centering
  \includegraphics[width=1\textwidth]{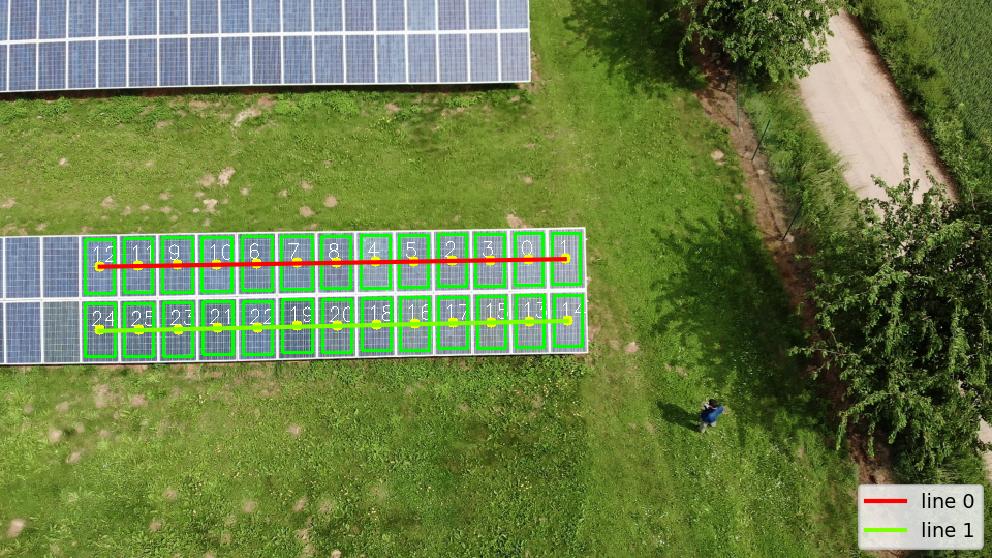}
  \caption{}
  \label{fig:cnn_structure_PPA}
\end{subfigure}

\vskip\floatsep%
\vspace{-0.1cm}

\begin{subfigure}[t]{0.48\textwidth}
  \centering
  \includegraphics[width=1\textwidth]{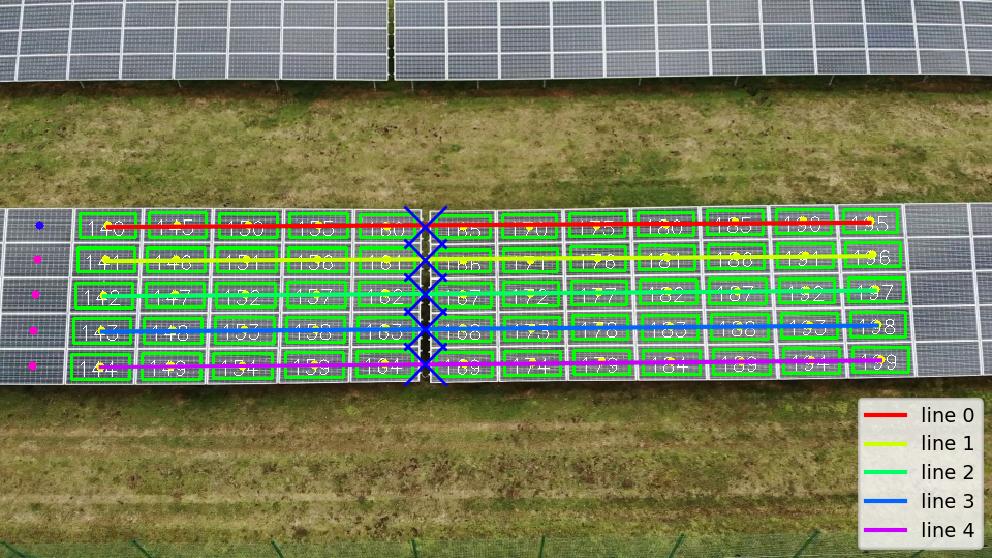}
  \caption{}
  \label{fig:cnn_structure_PPB}
\end{subfigure}

\caption{Semantic structures detected via the U-Net method and aligned using RANSAC.
Blue crosses mark detected bench gaps, closely described later in Section \ref{anchor_points}. Figures \textbf{(a)} and \textbf{(b)} show power plant A. Figure \textbf{(c)} shows an image from power plant B.}
\label{fig:semantic_structure_cnn}
\end{figure}

Following this, we identify modules belonging to different rows in the PV installation layout (Figure \ref{fig:semantic_structure_cnn}). 
The RANSAC algorithm is used to fit the center points of individual bounding boxes onto lines. In the algorithm, the maximum residual threshold value is set to 20\% of the representative bounding box height.  This can also eliminate possible outliers and false positive bounding box detections. Additionally, we conduct a parallelism check on the lines. Detections from individual rows are sorted by the position of their center points, forming a sequence that follows the direction of the flight path. Potential gaps caused by missing detections are identified by verifying the spacing between individual modules in each row and are considered while assigning the row sequence positions.

The final output of the CNN-based module detection methods is a semantic structure defined by a set of bounding boxes and their associated row and row sequence IDs.

\subsubsection{Edge-based Semantic Structure Detection}
\label{edge_based}

This method utilizes properties of typical PV power plant installations: distinctive frames of individual PV modules and perpendicular structures that typically consist of multiple aligned rows of PV modules. 

We employ the Canny edge detector to extract edges in the image and use the Hough transform to identify straight lines in the PV installation. The lines defined by module edges are mutually perpendicular, this geometric constraint is used to filter out redundant detections.

\begin{figure}[thpb]
\centering
\begin{subfigure}[t]{0.45\textwidth}
  \centering
  \includegraphics[width=1\textwidth]{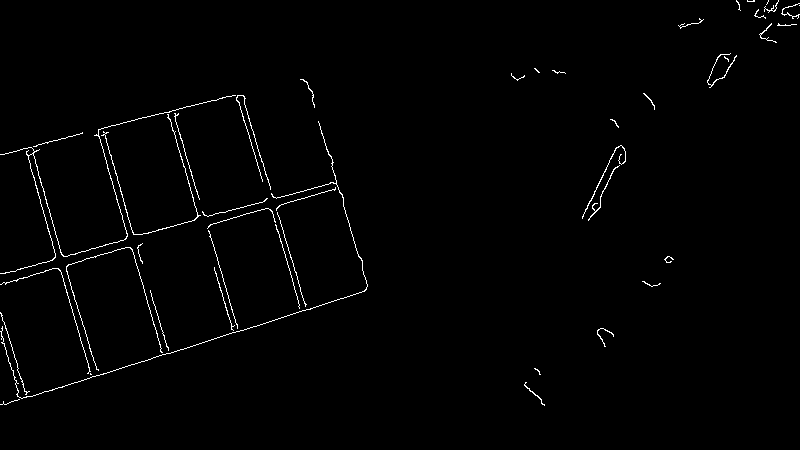}
  \caption{}
  \label{fig:edge_det0_edge}
\end{subfigure}\hspace{10mm}%
\begin{subfigure}[t]{0.45\textwidth}
  \centering
  \includegraphics[width=1\textwidth]{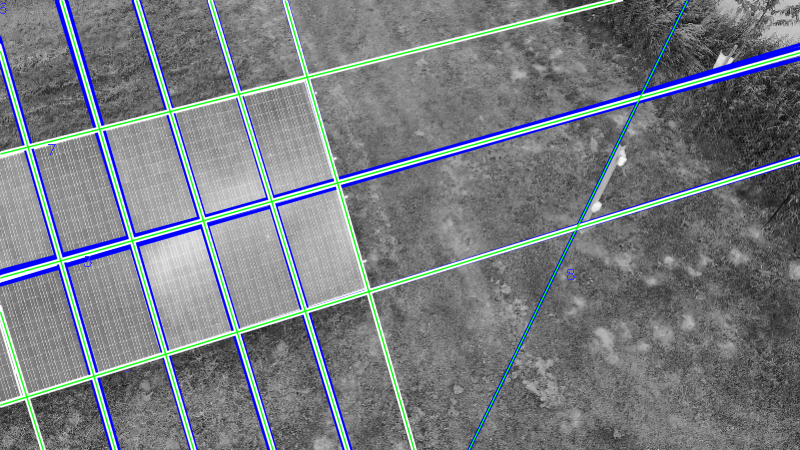}
  \caption{}
  \label{fig:edge_det1_lines}
\end{subfigure}

\vskip\floatsep%
\vspace{-0.1cm}

\begin{subfigure}[t]{0.45\textwidth}
  \centering
  \includegraphics[width=1\textwidth]{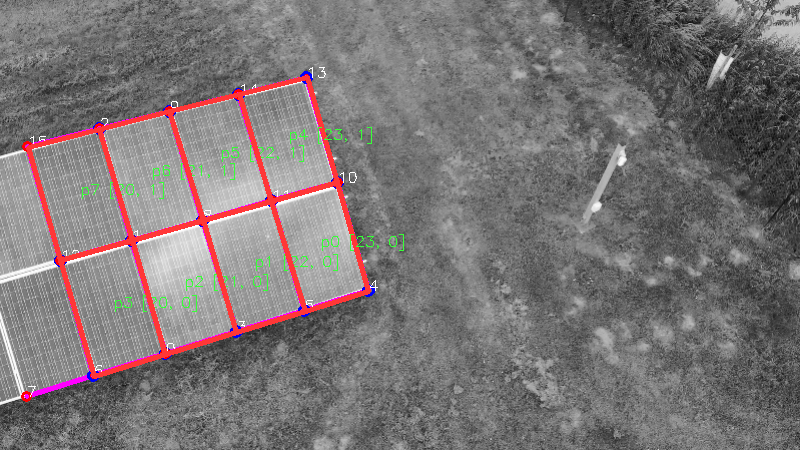}
  \caption{}
  \label{fig:edge_det2_final}
\end{subfigure}

\caption{Edge-based segmentation \textbf{(a)} Edge detection. \textbf{(b)} Line detection and filtering. \textbf{(c)} Final graph structure.}
\label{fig:edge_detection}
\end{figure}

The filtered lines are converted to an undirected graph $G (V, E)$ where the intersections of the lines form the vertices $V$ of the graph (Figure \ref{fig:edge_detection}). Finally, we identify sub-graphs that represent individual PV modules and use their adjacency to infer their logical coordinates in columns and rows.

The method builds upon the authors' previous work \citep{Kosnar2019}. The method was refined to reflect the specific properties of typical PV installations. In contrast to the previous implementation, the detected structures  (rows of PV panels) are long and narrow; therefore, line detection is divided into more steps, each one for a different direction. Moreover, multiple structures (rows) can be present in the image, so additional pre-segmentation is used. Detection of gaps is also not presented in the original approach. 
The detailed implementation of the method is described in \ref{a_edge_based}, where we provide technical details and the theory behind the method. We also introduce an optional pre-segmentation phase exploiting the U-Net segmentation model introduced in the previous sections.

The final output of the edge-based method is the vertices $V$ of the graph structure $G (V, E)$ and their logical coordinates. These represent the corners of PV modules that are used for camera localization.

\subsection{Anchor Point Initialization and Module Tracking}
\label{pv_tracking}

The detected PV modules are associated with the power plant model. The first association is performed using a set of visually recognizable navigation points. Afterwards, multi-object tracking is used to associate the detections between individual frames in the image sequence.

\subsubsection{Detection of Visual Anchor Points for Navigation}
\label{anchor_points}

We define two types of navigation points: bench ends and gaps between individual benches. The bench end and bench gap detection process is universal and is used with both edge-based and CNN-based detections.

The term "bench end" refers to both the start and the end of each row. During bench end identification, the in-image adjacency of the detected modules is used to infer the logical coordinates (columns and rows) of all detected modules. These are then associated with the power plant model.

\begin{figure}[h!]
    \centering
    \includegraphics[width=0.7\textwidth]{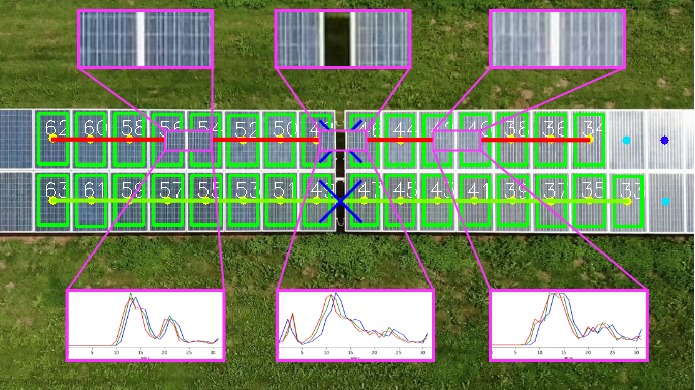}
    \caption{Illustration of extracted image patches with bench gap detection candidates and their respective intensity histograms. Blue crosses mark detected bench gaps.}
    \label{fig:spacing_detection}
\end{figure}

In this work, the localization pipeline is initialized solely from the bench ends. However, in case of defects at known positions, it can be beneficial to initiate inspection at an alternative location. Visually recognizable bench gaps, which can serve as an alternative to bench ends, typically exist between individual PV installation benches (Figure \ref{fig:spacing_detection}). 
We introduce and evaluate a new approach to their detection. 

Bench gaps are identified in two steps. First, we compute the distances between individual module detections in a row. Larger deviations in the module distance indicate the presence of gap candidates. We identify the median distance $d_{med}$ and set an acceptance deviation threshold of 10\% \mbox{($th = 0.1$)}. These are used to identify gap candidates based on the distance between neighboring modules $i$ and $j$:


\begin{equation}
\label{eq:gap_dist_med}
\begin{split}
    (1 + th) \cdot d_{med} < d_{ij} < 2(1 - th) \cdot d_{med}
\end{split}
\end{equation}

In the second step, we extract image patches that contain transitions between individual modules and verify the credibility of the gap candidates using histogram similarity (Figure \ref{fig:spacing_detection}). 
Bench gaps are identified by the presence of significantly darker transition patches.
This method was chosen because it does not require any training data. In practice, alternative classification techniques could also be employed.

\subsubsection{Module Tracking in an Image Sequence}
\label{module_tracking}

The PV module detections are tracked in the image sequence using optical flow. The detections are represented as a set of feature image points. The optical flow computed with each new image gives us the expected shift of these feature points in the image coordinates. New detections are associated with the previous image using the distance $d_{\Delta}$ between the predicted and new positions. The detections are associated when $d_{\Delta} < \alpha \frac{min(w,h)}{2}$, where $w$ and $h$ are the median values of the dimensions of the currently detected PV modules and $\alpha$ is an arbitrarily selected value (e.g. $\alpha = 0.8$), keeping the allowed distance below half of the smaller median module dimension.

A new unique track ID is assigned to all non-matched modules. 
To account for temporarily undetected modules, the motion of feature points from a longer history is calculated in consecutive frames.


\subsection{UAV Localization via Semantic Structures}
\label{uav_localization}

The developed navigation system works with a known model of the power plant that contains the global positions, orientations, and dimensions of all PV modules. The presented localization method uses the association of the detected semantic structures with the model to compute the relative position of the camera. 

\subsubsection{Relative Camera Localization}
\label{rel_cam_loc}

In the previous section, we presented three PV module detection approaches with two distinct output representations (Figure \ref{fig:semantic_structure0}). 
The output of the CNN-based approaches (Section \ref{CNN_based}) is a set of bounding box centers, denoting the centers of their respective PV modules. 
The output of the edge-based approach (Section \ref{edge_based}) is a graph structure created from the intersections of the detected lines. These represent the corners of their associated PV modules.

Since the positions and dimensions of the PV modules are known, both output representations can be expressed as a set of individual image points $\mathbf{p}^{\text{image}}$ associated with their respective real-world coordinates $\mathbf{p}^{\text{world}}$.

\begin{equation}
\label{eq:img_point_association}
     \{ (\mathbf{p}^{\text{image}}_n, \mathbf{p}^{\text{world}}_n) \}_{n=1}^N \quad \mathbf{p}^{\text{image}} \in R^2, \mathbf{p}^{\text{world}} \in R^3
\end{equation}

This general representation enables the use of different detection methods and reference points. 
The perspective projection model for cameras shown in equations \ref{eq:projection_model_simple} and \ref{eq:projection_model} describes the relation between the detected image point coordinates and their real-world coordinates. The equation describes the projection using the scaling factor $s$, the image coordinates $u$ and $v$, the intrinsic parameters of the camera defined by the focal lengths $f_x$ and $f_y$ and the coordinates of the principal point $c_x$ and $c_y$, the extrinsic parameters of the camera represented by the rotation matrix $R$ and the translation vector $t$, and finally the 3D coordinates of the point in the word coordinate system ($x,y,z$).

\begin{equation}
\label{eq:projection_model_simple}
 s\mathbf{p}^{\text{image}} = K[R|t]\mathbf{p}^{\text{world}} 
\end{equation}

\begin{equation}
\label{eq:projection_model}
s \begin{bmatrix} u \\ v \\ 1 \end{bmatrix} =
\begin{bmatrix} f_x & 0 & c_x \\ 0 & f_y & c_y \\ 0 & 0 & 1 \end{bmatrix}
 \begin{bmatrix} R & t \\ 0 & 1 \end{bmatrix} 
 \begin{bmatrix} x \\ y \\ z \\ 1 \end{bmatrix}
\end{equation}

The relative position of the camera can be obtained by solving the Per\-spec\-tive-n-Point (PnP) problem. Several common methods that can be used to solve it are available. 
Since our evaluation showed comparable performance across the tested PnP solvers, we utilize the EPnP algorithm presented in \citep{epnp2009}. 
The algorithm calculates the transformation $R$, $t$ that denotes the position of the origin of the world coordinate system (origin of the environment model) in the camera frame of reference. The position of the camera in the world coordinates is calculated using the inverse transformation. 

\begin{equation}
\label{eq:camera_position}
C = R^T(-t)
\end{equation}

The algorithm output also includes the reprojection error ($\varepsilon_r$), a measure of the disparity between the projected 2D points and their corresponding observed 2D points in the image. We rely on the reprojection error to evaluate the quality of the pose estimate.

\subsubsection{Continuous Localization} 

The relative localization method described in the previous section provides camera pose estimates from individual images. However, these are prone to noise and occasional errors. To enhance the accuracy and robustness of these estimates, a simplified Kalman filter is employed as a state estimator, integrating PnP-based pose estimations, GNSS data, and the UAV motion model.

The Kalman filter is used to produce a state vector $X_k$ with the camera position, velocities, and orientation at time $k$.

\begin{equation}
\label{eq:state_vector}
X_k = \begin{bmatrix} x , y , z , v_x , v_y , v_z, o_x , o_y , o_z \end{bmatrix}^{T}_k
\end{equation}

The prediction step utilizes the motion model of the UAV, incorporating the elapsed time $\Delta t$ through the state transition matrix $F(\Delta t)$, and the optional control input $U_k$:

\begin{equation}
\label{eq:prediction}
\hat{X}_k = F(\Delta t) \cdot \hat{X}_{k-1} + B(\Delta t) \cdot U_k
\end{equation}

Here, $\hat{X}_k$ is the predicted state vector and $B(\Delta t)$ is the control input matrix. 

The update step refines the prediction using actual measurements $Z_k$:

\begin{equation}
\label{eq:measurement_vector}
Z_k = \begin{bmatrix} x , y , z , v_x , v_y , v_z, o_x , o_y , o_z \end{bmatrix}^{T}_k
\end{equation}

The $Z_k$ vector is composed using a combination of the PnP pose estimation and the GNSS data. The results of the camera pose estimation method described in the previous section are used as the position and orientation measurement values. This provides direct feedback on the camera position relative to the detected structures. However, because these are prone to noise, they are not suitable for estimating velocity. The velocity measurements are derived from GNSS data, since these are relatively stable and suitable for velocity inference. 

The update step is defined as follows:

\begin{equation}
\label{eq:update}
\hat{X}_k = \hat{X}_{k|k-1} + K(w_{\text{PnP}}, w_{\text{vel}})_k \cdot (Z_k - H \cdot \hat{X}_{k|k-1})
\end{equation}

Here, $H$ is the measurement matrix that maps the state space to the measurement space. Since the measurement vector $Z_k$ takes the same form as the state vector $X_k$, $H$ is an identity matrix.

The diagonal Kalman gain matrix $K_k$ represents the weights given to the measurements. We adjust the Kalman gain with confidence values $w_{\text{PnP}}$ and $w_{\text{vel}}$ used on the main diagonal of the matrix. 

We used a fixed confidence value ($w_{\text{vel}} = 1$) for velocity measurements ($v_x , v_y$, $v_z$). 
However, we adjust the weight of PnP position and orientation measurements using the reprojection error $\varepsilon_r$ and an adjustable sensitivity factor $\sigma$. Additionally, we directly discard measurements with a reprojection error greater than a defined reprojection threshold $th_r$ and measurements deviating from the predicted state vector position by a margin larger than distance threshold $th_d$. We also limit the maximum value of $w_{\text{PnP}}$.

\begin{equation}
\label{eq:kalman_weight}
w_{\text{PnP}} = \begin{cases}
    0 & \text{if } \varepsilon_r > th_r \text{  or  } \varepsilon_d > th_d\\
    \min(\frac{th_r}{\varepsilon_r}\sigma, 2\sigma) 
\end{cases}
\end{equation}

Here, $\varepsilon_d $ marks the deviation of the current measurement $Z_k$ from the last state estimate $\hat{X}_{k-1}$ in the $x, y,$ and $z$ axes.

\section{Results} 
\label{results}

This section presents results from the experimental demonstration and evaluation of the proposed visual localization pipeline. Section \ref{experimental_setup} describes the experimental setup and the testing and training datasets. The evaluation of the presented PV module detection methods is given in Section \ref{pv_detection_results}. 
The localization results are presented in Section \ref{localization_results}. 
The computation times are given in Section \ref{comp_times}. Finally, Section \ref{results_and_discussion} concludes with a discussion of the achieved results.

\subsection{Experimental Setup}
\label{experimental_setup}

The localization system was tested using the DJI Matrice 300 RTK drone equipped with the DJI Zenmuse H20T camera capable of capturing 30 frames per second (fps). 
The testing images were captured during a manual inspection flight using a video stream with the camera in wide mode, resulting in a sequence of RGB images with 1920 x 1080 px resolution. 
GNSS positions from the UAV's onboard computer were acquired without the use of a real-time kinematic (RTK) positioning unit. 
The SciPy \citep{scipy2020}, OpenCV \citep{opencv_library}, and scikit-learn \citep{scikit} libraries were used in the implementation of the system.

\subsubsection{Testing Dataset}
\label{test_datasets}

The evaluation dataset was acquired from two power plants, which differ in the layout of the PV modules. Modules in power plant A are positioned vertically in two rows per bench, while modules in power plant B are positioned horizontally in five rows per bench. Power plant A also contains a mixture of modules with a lighter and darker appearance (see Figure \ref{fig:cnn_structure_PPA2}). The images were captured in fine weather conditions; however, some exhibit slight motion blur. 

The first dataset for the power plant A (PPA1) was captured in close proximity during a manual flight over a row of 3 benches, with a total of 154 PV modules (Figure \ref{fig:cnn_structure_PPA2}). It consists of 327 image frames, with a total of 7678 possible PV module detection instances and 184 bench gap detection instances. 
The second dataset from power plant A (PPA2) was captured from a larger distance (Figure \ref{fig:cnn_structure_PPA}). The PPA2 dataset contains 10 benches and 462 PV modules. It consists of 1083 image frames and contains 43706 PV module detection instances and 1450 bench gaps instances. The power plant B test dataset (PPB) contains two benches with a total of 350 PV modules and consists of 378 image frames, with a total of 19940 PV module detection instances and 300 bench gap instances (Figure \ref{fig:cnn_structure_PPB}). 
Details for each testing dataset are summarized in Table \ref{tbl_test_dataset}.

\begin{table}[h!]
\renewcommand{\arraystretch}{1.3}
\caption{Testing dataset parameters.}
\vspace{-2mm}
\label{tbl_test_dataset}
\scriptsize
\begin{center}
\begin{tabular}{|l|l||c|c|c|}
\hline
\multicolumn{2}{|l||}{Dataset} & PPA1 & PPA2 & PPB  \\ \hline \hline
\multirow{3}{*}{\makecell{Power plant \\ segment \\ parameters}} & N. of PV module rows & 2 & 2 & 5 \\ \cline{2-5}
 & N. of benches & 3 & 10 & 2 \\ \cline{2-5}
 & N. of PV modules & 154 & 462 & 350 \\ \hline \hline
\multirow{6}{*}{\makecell{Dataset \\ parameters}}  & Image frames & 327 & 1083 & 378\\ \cline{2-5}
 & Possible PV module detections & 7378 & 43706 & 19940 \\ \cline{2-5} 
& Possible bench gap detections & 184 & 1450 & 300 \\ \cline{2-5}
& Image capture distance from modules & 12 m & 23 m & 14 m \\ \cline{2-5}
& Maximum flight speed & 0.8 m/s & 3.8 m/s & 1.8 m/s \\ \cline{2-5}
& Tested fps values & 3 & 15 & 6 \\ \hline
\end{tabular}
\end{center}
\end{table}

The approximate distances between the camera and the inspected PV modules for the PPA1, PPA2, and PPB datasets were 12, 23, and 14 meters, respectively. The image processing frequencies are adjusted to reflect the maximum speed of the UAV during each flight. We set the minimum requirement to one frame for each 0.33 meters (one third of the width of a PV module). The PPA1 flight had a maximum speed of 0.8 m/s. We set the processing frequency to 3 fps. The PPA2 flight had a maximum speed of 3.8 m/s and the frequency was set to 15 fps, taking every second frame from the camera stream. The PPB flight had a maximum speed of 1.8 m/s and the frequency was set to 6 fps.

A known power plant model is needed to associate the detected structures with
their respective real-world coordinates. We evaluated the proposed localization methods using two types of power plant model, the high-altitude (\textit{H-Alt}) model and the structure-from-motion (\textit{SfM}) model. The \textit{H-Alt} model was created using publicly available high-altitude aerial images \citep{mapyCZ} and topographic maps \citep{SRTM2014}. It served as a valid use case scenario, as this is one of the common methods used for power plant model acquisition. The \textit{SfM} model was generated directly from the power plant A and B datasets using the OpenSfM library \citep{opensfm} for structure-from-motion reconstruction. This model is suitable for assessing the accuracy of the proposed localization methods, since OpenSfM provides both the environment model and the camera positions optimized over all input images. 

In practice, the georeferenced PV module positions are a part of a compact database file that describes the power plant structure, wiring, details of individual PV modules, and various other information. 
For reference, power plant A spans an area of 275 x 140 meters, contains a total of 4974 PV modules and the size of the full database file is 600kB.

\subsubsection{CNN Training}
\label{train_datasets}

CNN-based object detection methods generally require a large amount of training data. We created a custom dataset with more than 3,000 PV module annotations, which can be used to train both the YOLO and \mbox{U-Net} models.
Half of the dataset targets the intended use case scenario and contains images captured during manual aerial PV power plant inspection. The second half of the training dataset comprises publicly available images with different PV installations, which are included to diversify the training data. Training images include only images of power plant A and publicly available images. Power plant B was not included in the training dataset, in order to test the generalizability of the segmentation methods. 
Both the training dataset and the trained models have been made publicly available at \url{https://imr.ciirc.cvut.cz/Datasets/PVsegmentation}.

Image annotation was performed using the Grid Annotation Tool \citep{annotationTool}. The tool enables annotation through the use of intersecting lines. This speeds up the annotation process, since there is no need to draw bounding boxes for each object individually. It also leads to cleaner and more aligned edges. 
We perform additional erosion on the resulting masked cells to separate individual instances more clearly.  
The training of the U-Net model for PV plant presegmentation (supporting the edge-based method) was performed on labels without the additional erosion of masks.

The annotated dataset was split into training (80\%) and validation (20\%) partitions. The setup of the two CNN-based methods is as follows.

We utilized the smallest YOLOv8n-seg model pre-trained on the COCO dataset \citep{coco2014}. The model was re-trained for the detection of PV modules for 200 epochs with a learning rate of 0.01. The training input was the original RGB images and contours of individual segmented objects. 

The U-Net model was trained for 50 epochs with a learning rate of $1\times10^{-5}$.
The U-Net training uses the original images and their respective segmentation masks.

\subsubsection{Settings}

The PV module detection methods are used as follows. 
The YOLO instance segmentation framework is used in its default configuration. The input images are scaled to a maximum dimension of 640 px. The detection threshold value for YOLO was set to $th=0.8$. 
The input dimensions for U-Net segmentation can be freely adjusted. We used input images resized to 480 x 270 px. The segmentation mask threshold value was set to $th=0.8$. 
The edge-based method uses images scaled to a maximum dimension of 800 px. The additional U-Net presegmentation is used with images resized to 400 x 225 px.

In the localization pipeline, we filter unreliable PnP position estimates using a threshold based on the median reprojection error values of all current PnP estimates ($th_r = 2 \times \text{med}(\varepsilon_r)$). This is a characteristic of the system and the optimal value of $th_r$ should be estimated and fixed during the initial setup of the system for a new power plant. 
The Kalman filter was used without control input $U_k$, since it was not available in the log data. The sensitivity factor for the Kalman gain was set to $\sigma =0.16$. The distance threshold $th_d$ was set to 10 meters.

\subsection{Photovoltaic Module Detection}
\label{pv_detection_results}

The performance of the three presented methods for visual segmentation of PV modules was evaluated on the PPA1, PPA2 and PPB testing datasets. 
The test images from PPA2 and PPB contain segments with neighboring bench rows. However, these are successfully filtered as a result of the data association performed in combination with module tracking and initialization through visual anchor points.

\begin{table}[h!]

\renewcommand{\arraystretch}{1.3}
\caption{Success rates for detection of PV modules and bench gaps.}
\vspace{-2mm}
\label{tbl_detection_success}
\scriptsize
\begin{center}
\begin{tabular}{|l||c|c|c|c|c|c|c|c|}
\hline
     & \multicolumn{2}{c|}{PPA1} & \multicolumn{2}{c|}{PPA2} & \multicolumn{2}{c|}{PPA2 - narrow} & \multicolumn{2}{c|}{PPB}  \\ \hline 
    & Modules & Gaps & Modules & Gaps & Modules & Gaps & Modules & Gaps  \\ \hline \hline
Edge & 99.7\% & 64.7\% & 74.3\% & 60.0\% & 84.5\% & 74.4\% &  96.1\%   & 68.0\%  \\ \hline
U-Net & 98.0\% & 83.7\%  & 89.5\% & 84.0\% & 97.2\% &  94.1\% & \textbf{100.0\%}   & \textbf{97.0\%}  \\ \hline
YOLO  & \textbf{100.0\%} & \textbf{92.4\%} & \textbf{93.5\%} & 8\textbf{4.8\%} & \textbf{98.3\%} & \textbf{97.1\%} &  85.8\% & 66.3\%  \\ \hline
\end{tabular}
\end{center}
\end{table}

\begin{figure}[thpb]
\centering
\begin{subfigure}[t]{.45\textwidth}
  \centering
  \includegraphics[width=1\textwidth]{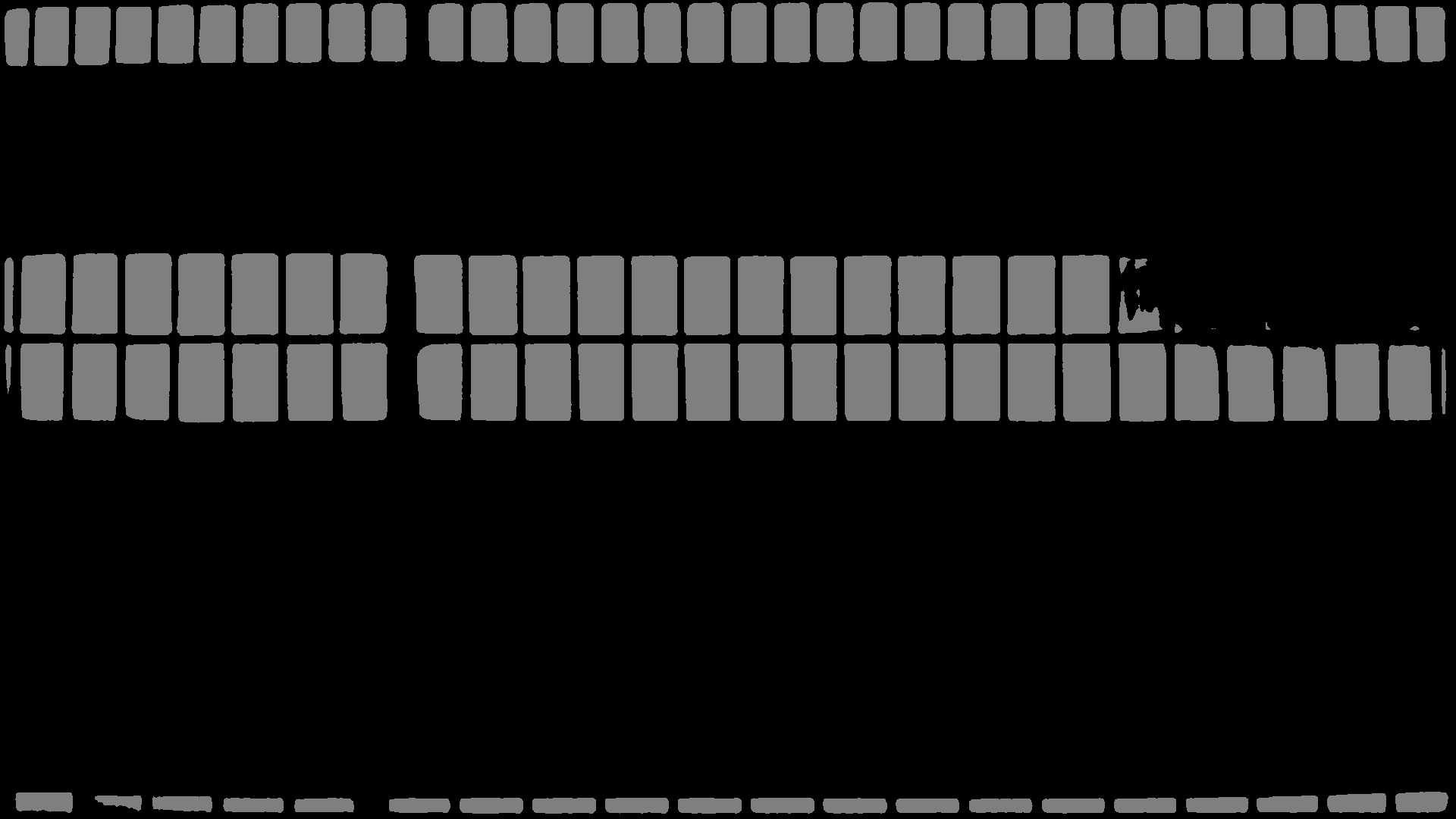}
  \caption{}
  \label{fig:wrong_mask}
\end{subfigure}\hspace{3mm}%
\begin{subfigure}[t]{.45\textwidth}
  \centering
  \includegraphics[width=1\textwidth]{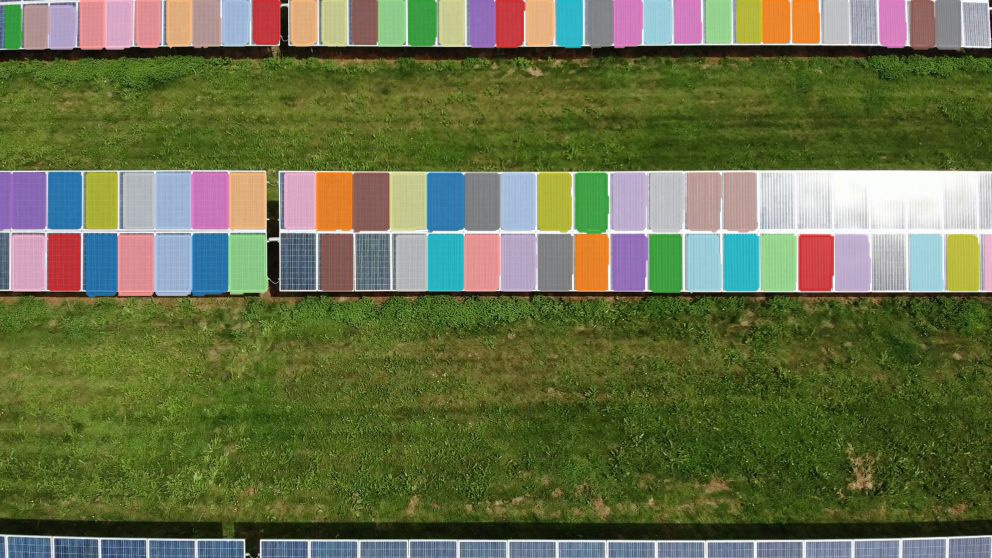}
  \caption{}
  \label{fig:wrong_yolo}
\end{subfigure}

\caption{\textbf{(a)} U-Net: Incomplete segmentation mask caused by the sun reflection. \textbf{(b)} YOLO: Systematic failure caused by sun reflection and several incidental missed PV module detections.}
\label{fig:sun_reflection}
\end{figure}

The detection success rates for individual segmentation methods are presented in Table \ref{tbl_detection_success}. A systematic error caused by sun reflection was present in the PPA2 dataset (Figure \ref{fig:sun_reflection}). We offer additional detection results for the dataset with a narrowed field of view, showing the success rates without this interference. This is done by cropping the image by 20\% on each side. 

In power plant A, both YOLO and U-Net performed better than the edge-based method. However, the YOLO detection success rate for power plant B is noticeably lower than that of the other two methods. This drawback can be attributed to the underrepresentation of photovoltaic installations with horizontally arranged modules in training data. From the results, we can infer that the generalization of this method to various power plant layouts is limited. However, this can be mitigated by providing additional training data.

The performance of the edge-based method decreased in the second half of the PPA2 dataset, which contained darker PV modules. This can be attributed to the fixed parameter tuning, which is not re-adjusted when switching between different module types. Overall, the method successfully detected a valid semantic structure only in 99.1\% of the PPA2 images.

\subsection{UAV Localization}
\label{localization_results}

This section details the performance of the localization pipeline, focusing on the results of the PnP and Kalman pose estimations. 
The experiments were carried out using two types of power plant models: the high-altitude (H-Alt) model and the structure-from-motion (SfM) model. 

\subsubsection{UAV Localization - \textit{H-Alt} Model}
\label{halt_results}

The PnP estimates give a position relative to the power plant model. 
For reference, we also included the GNSS position from the UAV's onboard computer. 
However, the model is not synchronized with the GNSS data from the UAV. 
Thus, if the position of the model is inaccurate, the output of the localization method will diverge from the absolute frame of reference (word coordinates). A similar situation may arise if the GNSS data from the UAV exhibit faults (e.g., shift or drift). 

The presented navigation pipeline was designed to correct for such inaccuracies. The vision-based camera-to-PV module localization allows us to maintain an optimal position and distance from the detected objects.

\begin{figure}[thpb]
\centering
\begin{subfigure}[t]{.47\textwidth}
  \centering
  \includegraphics[width=1\textwidth]{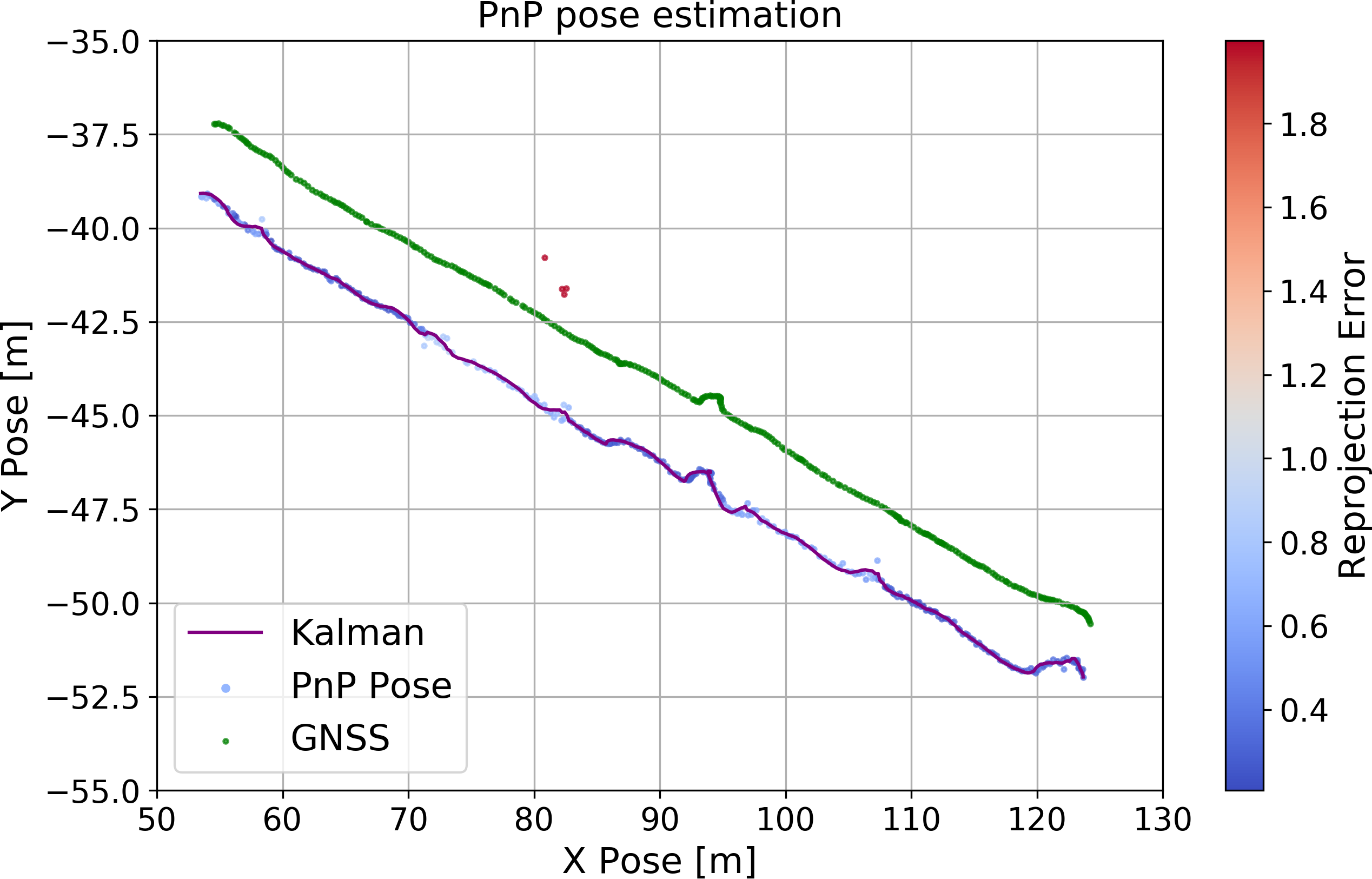}
  \caption{}
  \label{fig:PPA2edgeY}
\end{subfigure}\hspace{3mm}%
\begin{subfigure}[t]{.47\textwidth}
  \centering
  \includegraphics[width=1\textwidth]{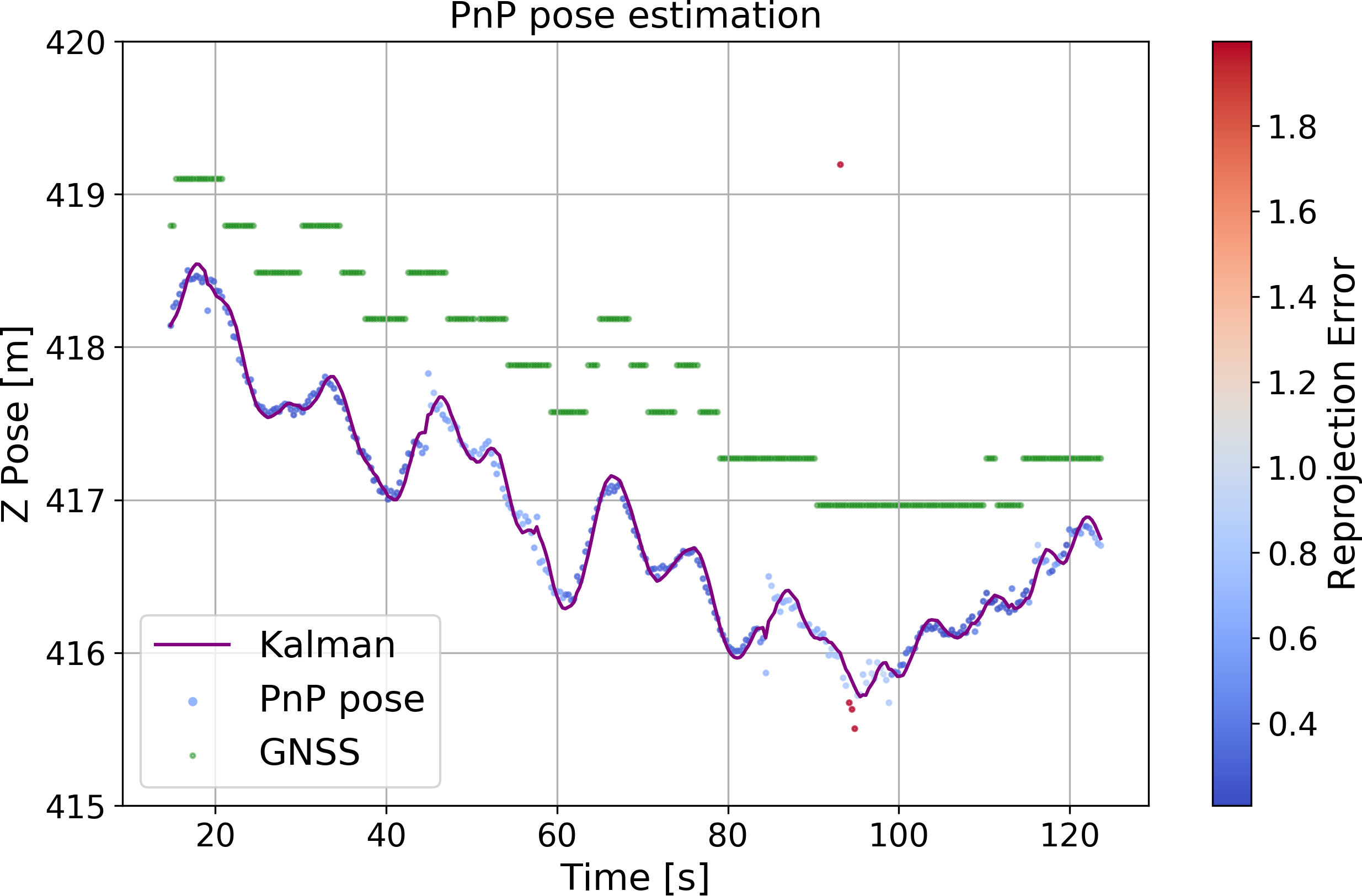}
  \caption{}
  \label{fig:PPA2edgeZ}
\end{subfigure}


\begin{subfigure}[t]{.47\textwidth}
  \centering
  \includegraphics[width=1\textwidth]{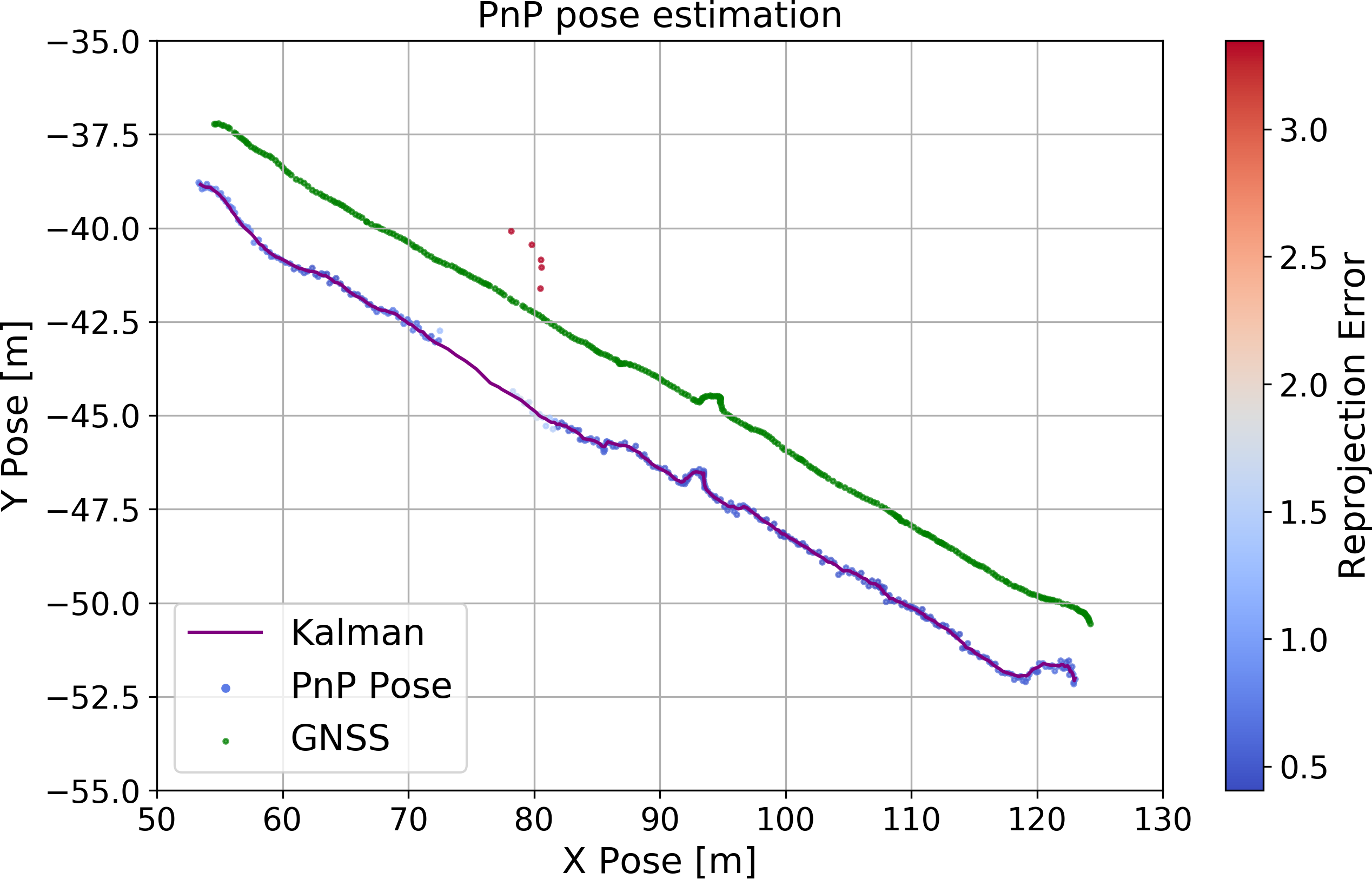}
  \caption{}
  \label{fig:PPA2unetX}
\end{subfigure}\hspace{3mm}%
\begin{subfigure}[t]{.47\textwidth}
  \centering
  \includegraphics[width=1\textwidth]{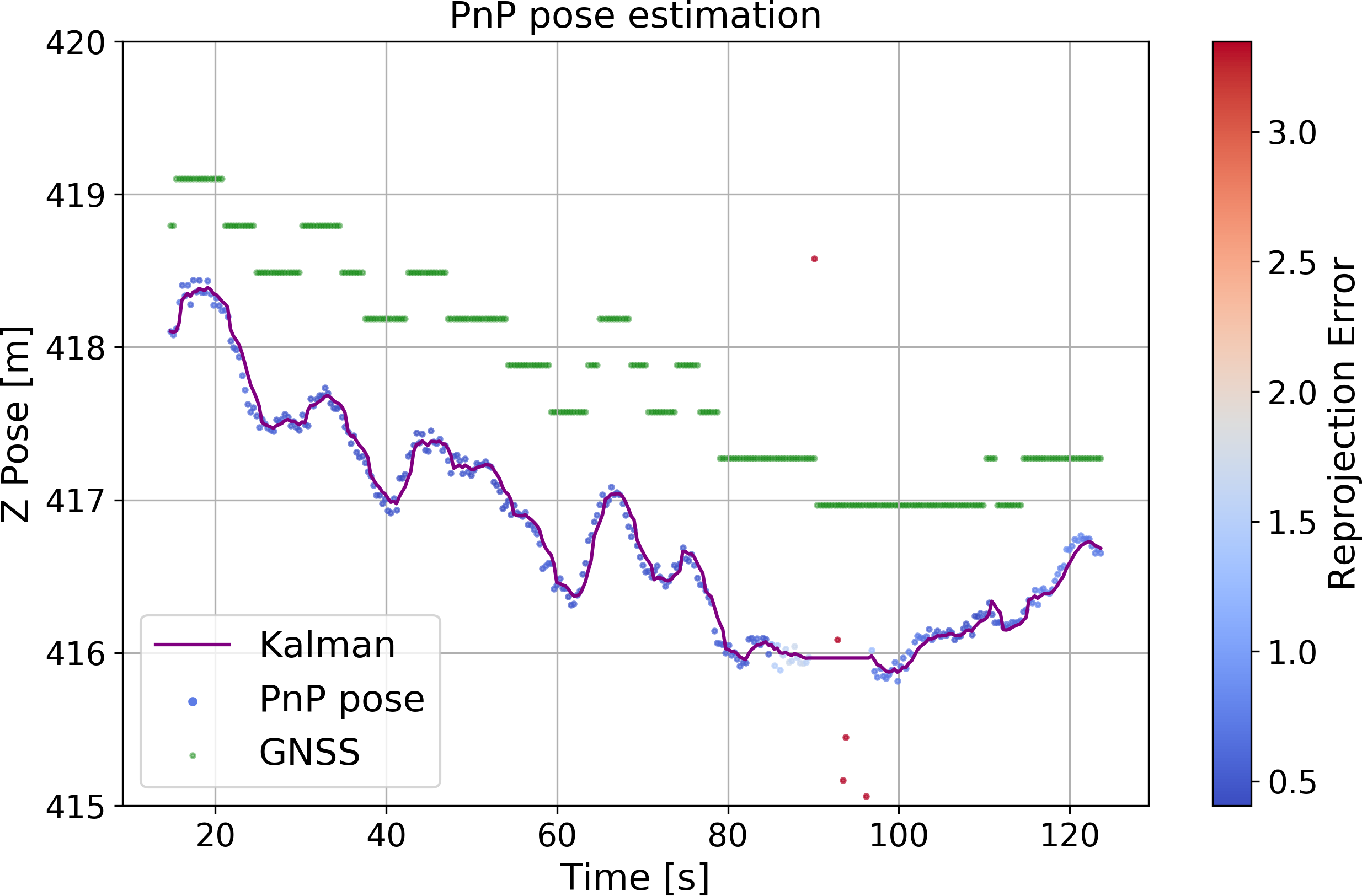}
  \caption{}
  \label{fig:PPA2unetZ}
\end{subfigure}


\begin{subfigure}[t]{.47\textwidth}
  \centering
  \includegraphics[width=1\textwidth]{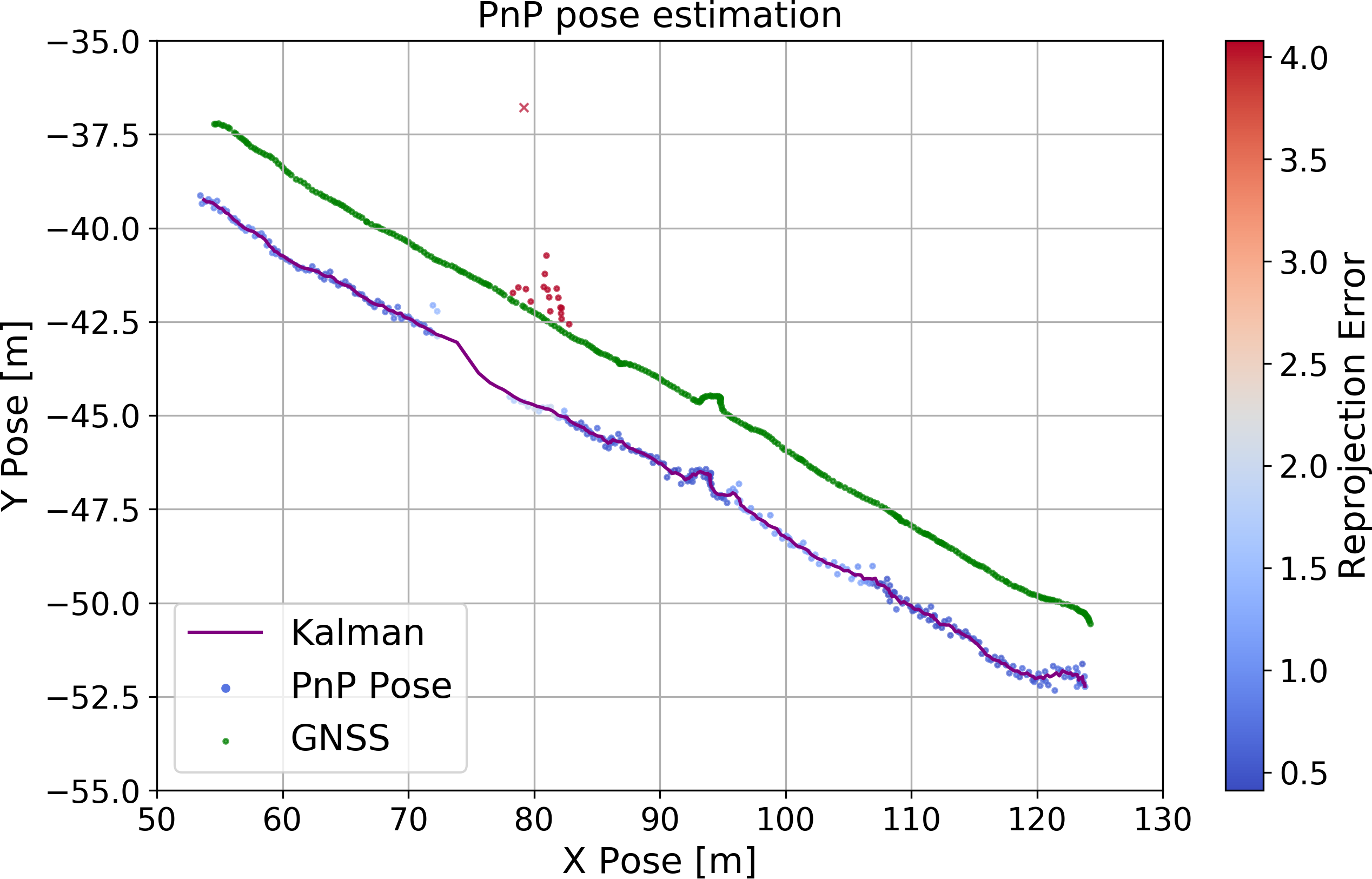}
  \caption{}
  \label{fig:PPA2yoloX}
\end{subfigure}\hspace{3mm}%
\begin{subfigure}[t]{.47\textwidth}
  \centering
  \includegraphics[width=1\textwidth]{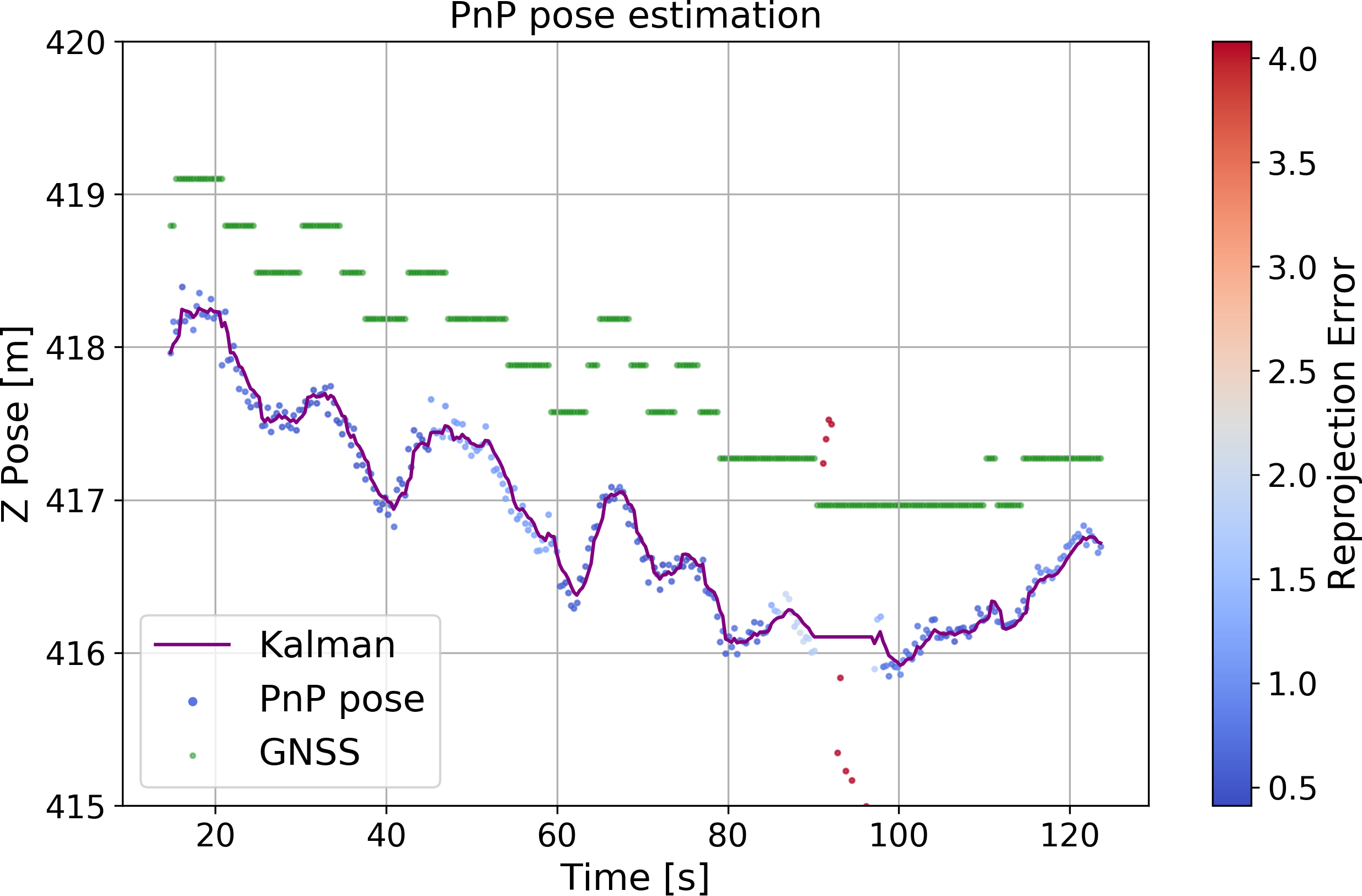}
  \caption{}
  \label{fig:PPA2yoloZ}
\end{subfigure}


\caption{The PnP output for PPA1 using the \textit{H-Alt} model. The reprojection error is illustrated using a color map. Deviations exceeding the distance threshold $th_d$ are indicated by crosses. Figures \textbf{(a)} and \textbf{(b)} present results of the edge-based method. Figures \textbf{(c)} and \textbf{(d)} present results for U-Net. Figures \textbf{(e)} and \textbf{(f)} present results for YOLO.}
\label{fig:axes_repr_HAlt_PPA2_new}
\end{figure}

Detailed results of pose estimation on the PPA1 dataset are presented in Figure \ref{fig:axes_repr_HAlt_PPA2_new}. This is an example figure which provides a detailed insight, containing both the Kalman and the PnP pose estimation results. In addition, individual PnP estimates are colored according to their reprojection error. 
We proceed by presenting the remaining results without individual PnP estimations; however, detailed figures for the remaining datasets can be found in \ref{apdx_flights_halt}.

There is a noticeable shift between the GNSS positions and the PnP position estimates. This disparity arises due to the shift between the absolute position of the \textit{H-Alt} model and the GNSS data from the UAV. 
However, it can be seen that the perceived trajectory closely resembles the GNSS data, and we are able to compensate for the position shift using the presented localization method.

\begin{figure}[thpb]
\centering

\begin{subfigure}[t]{.47\textwidth}
  \centering
  \includegraphics[width=1\textwidth]{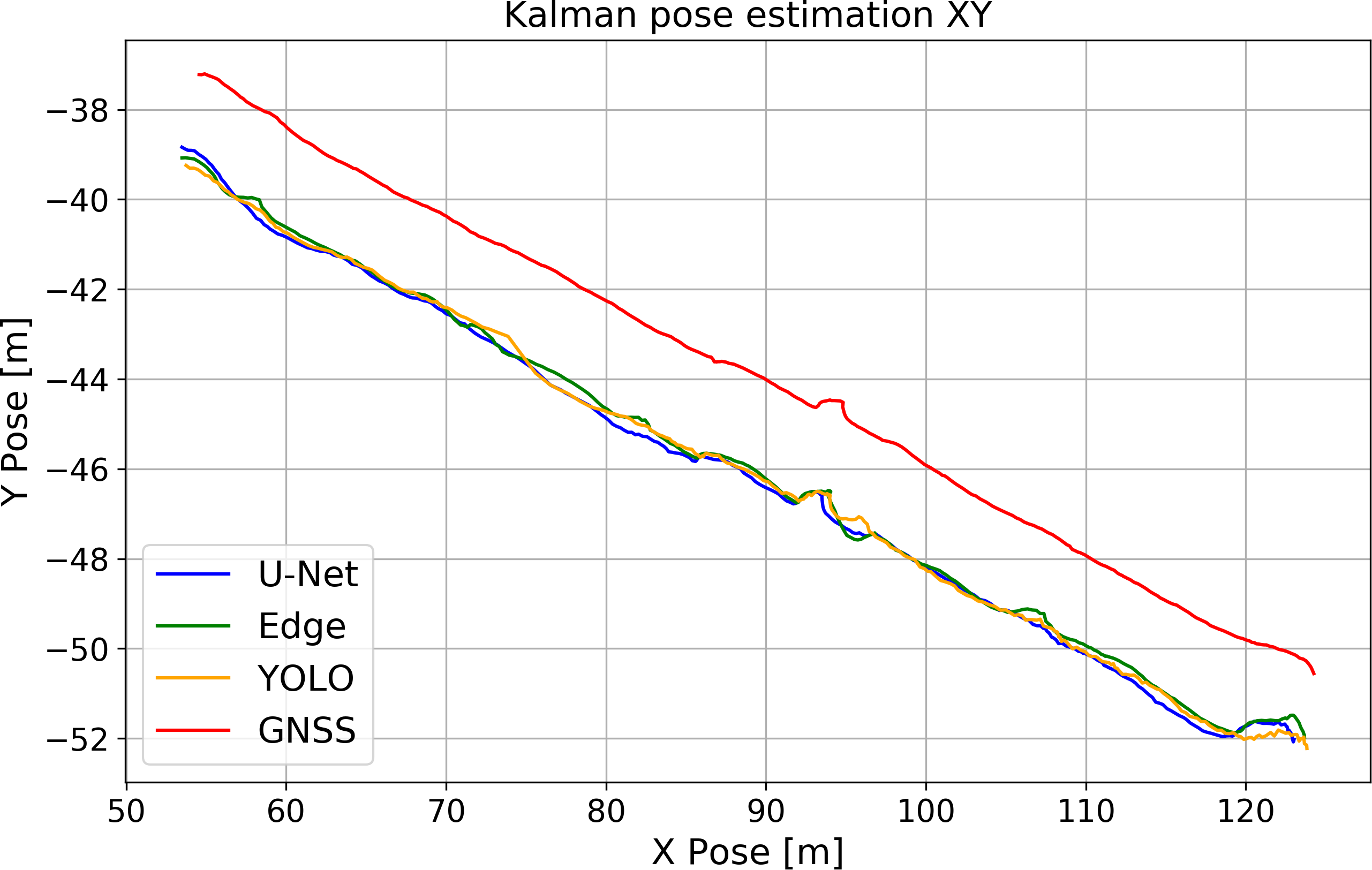}
  \caption{}
  \label{fig:kalmans_A2_XY}
\end{subfigure}\hspace{3mm}%
\begin{subfigure}[t]{.47\textwidth}
  \centering
  \includegraphics[width=1\textwidth]{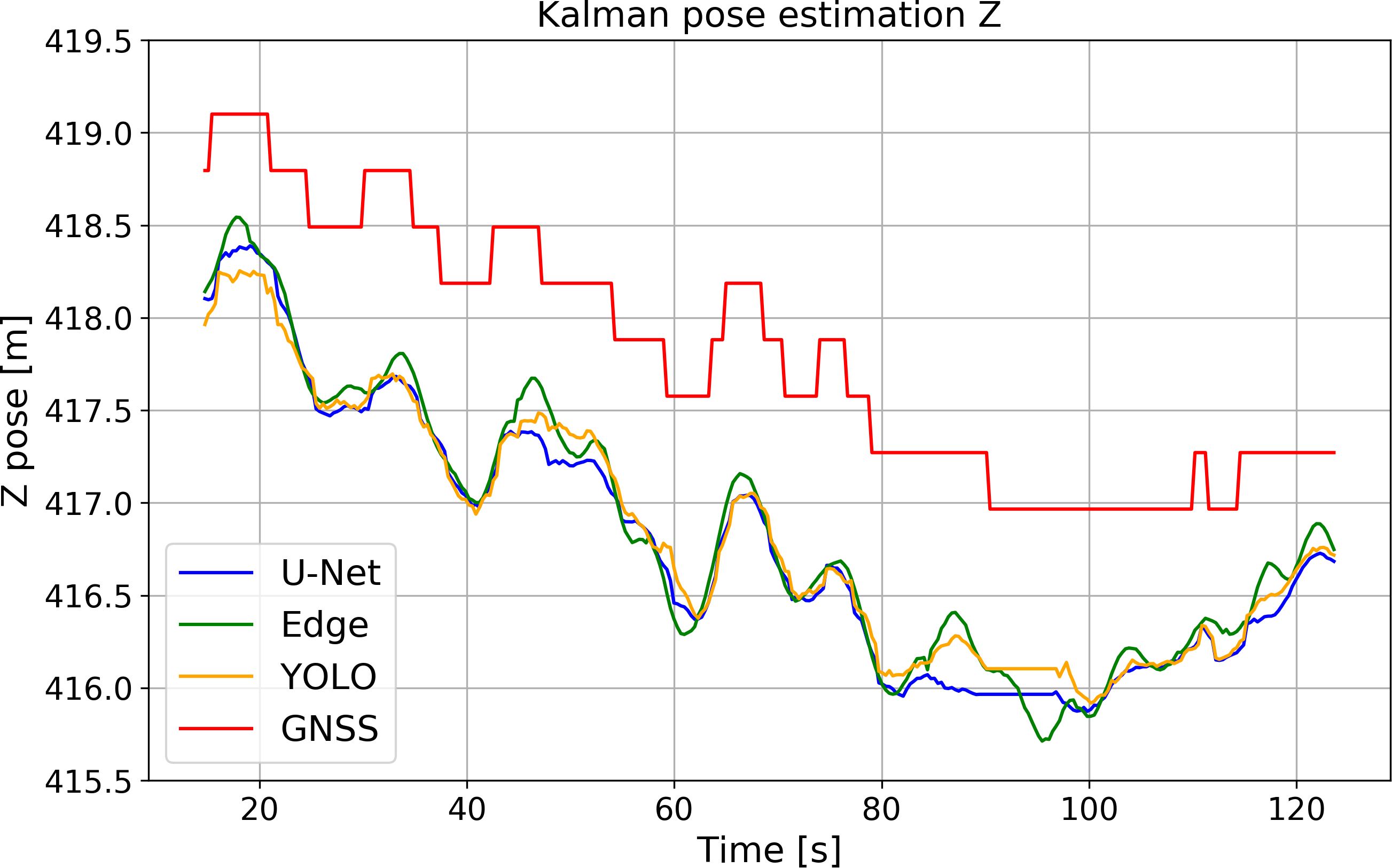}
  \caption{}
  \label{fig:kalmans_A2_Z}
\end{subfigure}

\begin{subfigure}[t]{.47\textwidth}
  \centering
  \includegraphics[width=1\textwidth]{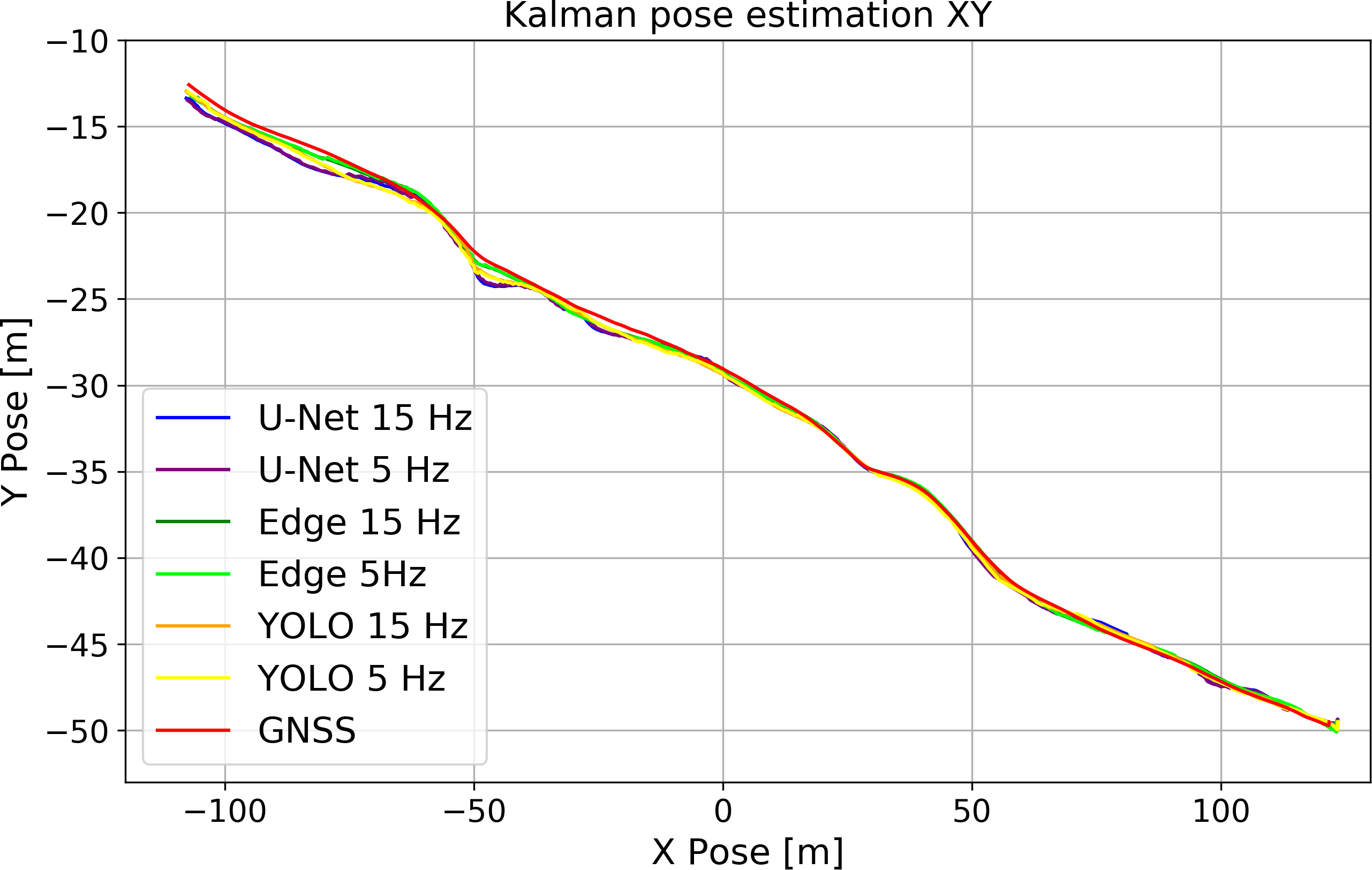}
  \caption{}
  \label{fig:kalmans_A_XY}
\end{subfigure}\hspace{3mm}%
\begin{subfigure}[t]{.47\textwidth}
  \centering
  \includegraphics[width=1\textwidth]{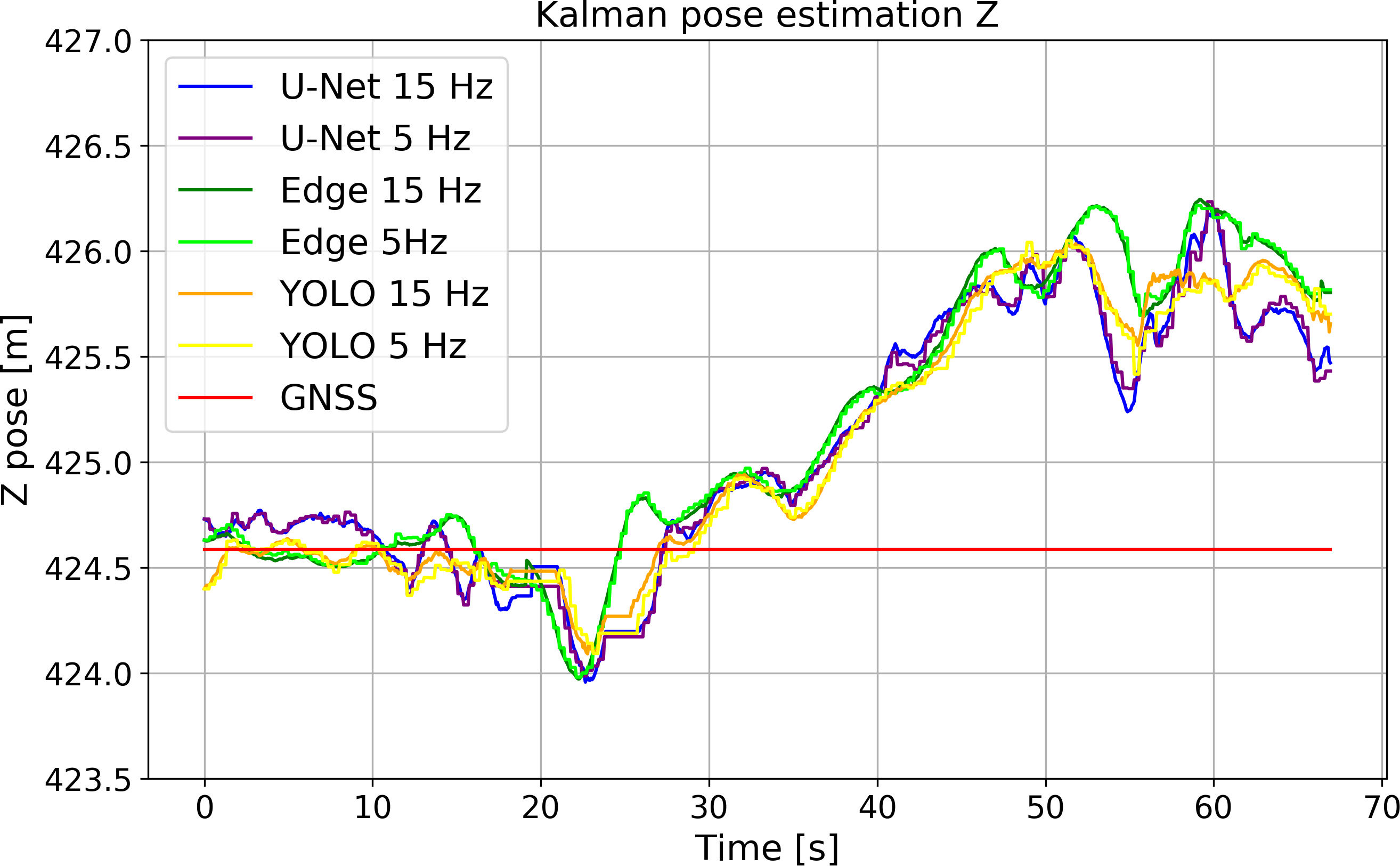}
  \caption{}
  \label{fig:kalmans_A_Z}
\end{subfigure}

\begin{subfigure}[t]{.47\textwidth}
  \centering
  \includegraphics[width=1\textwidth]{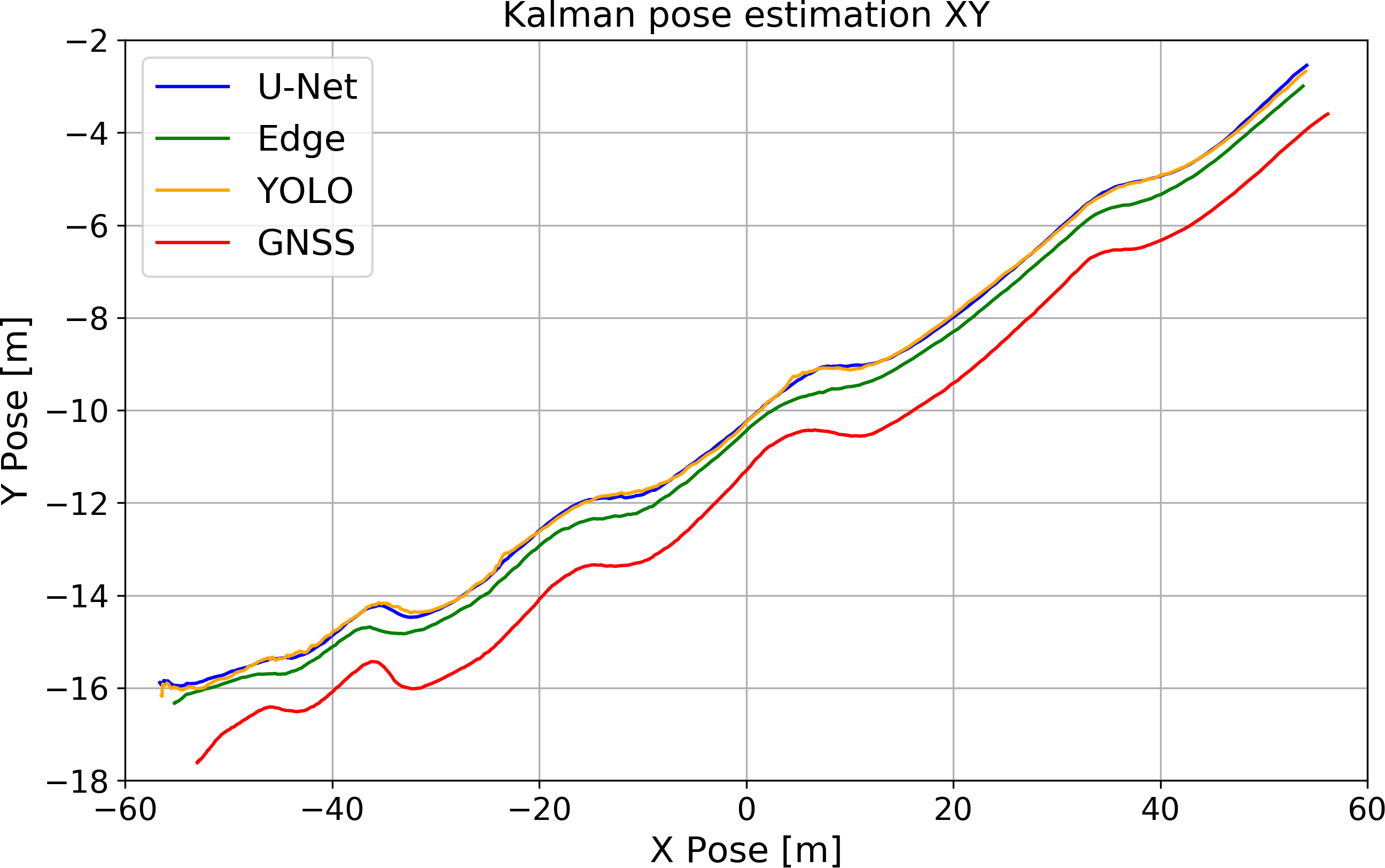}
  \caption{}
  \label{fig:kalmans_B_XY}
\end{subfigure}\hspace{3mm}%
\begin{subfigure}[t]{.47\textwidth}
  \centering
  \includegraphics[width=1\textwidth]{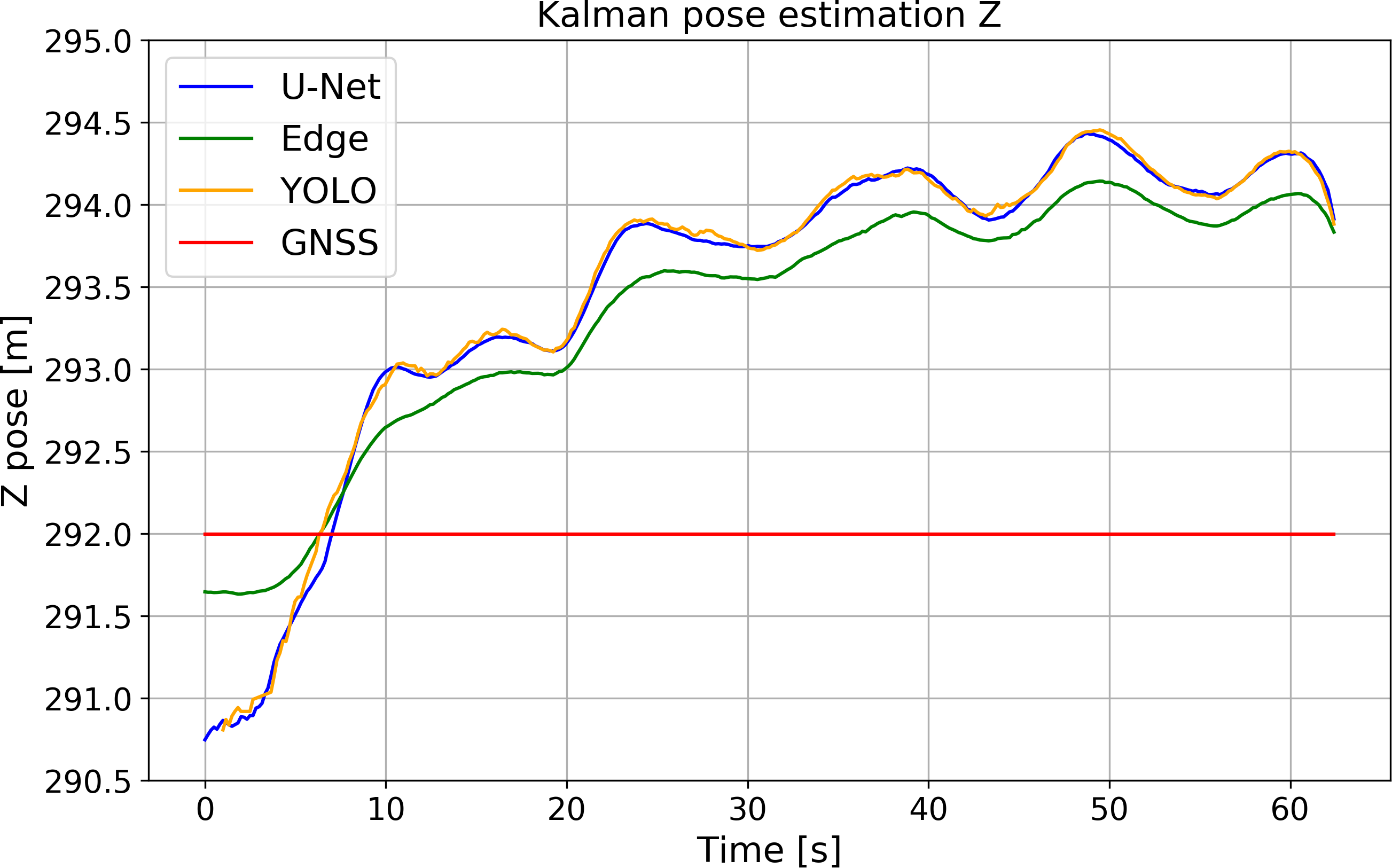}
  \caption{}
  \label{fig:kalmans_B_Z}
\end{subfigure}

\caption{\textit{H-Alt} model. Kalman pose estimation for individual methods, together with the GNSS position values. Figures (a) and (b) show results for PPA1. Figures (b) and (c) show results for PPA2. Figures (d) and (e) show results for PPB.}
\label{fig:halt_kalmans}
\end{figure}

A comparison of the resulting Kalman pose estimates on all datasets is given in Figure \ref{fig:halt_kalmans}. 
An additional pass for the PPA2 flight with the stream limited to 5 fps was performed to simulate possible conditions with slower computational speeds. (Figures \ref{fig:kalmans_A_XY} and \ref{fig:kalmans_A_Z}).

The desired altitudes during the PPA2 and PPB flights were set to a fixed value. We can see that the Z-axis position given by the GNSS is constant in Figures \ref{fig:kalmans_A_Z} and \ref{fig:kalmans_B_Z}. This is in contrast to the estimated positions that diverge from this value. 
This is caused by two factors. First, the internal UAV navigation module fails to accurately reflect changes in the real altitude of the UAV. 
Second, our localization method estimates the position relative to the power plant model, which may not coincide with its true location, allowing the method to adjust by localizing directly with respect to the detected structures.




\begin{table}[h!]
\renewcommand{\arraystretch}{1.3}
\caption{Pose estimation results - \textit{H-Alt} model. 
}
\vspace{-2mm}
\label{tbl_HAlt_pnp_success}
\scriptsize
\begin{center}
\begin{tabular}{|l||c|c|c|c|c|c|}
\hline
        & Method & valid PnP &  median $\varepsilon_r$ [px] & $\varepsilon_r < th_r$ & $\varepsilon_d < th_d$ & mean $\varepsilon_d$ [m] \\ \hline \hline
PPA1    & Edge &  100\%  & 0.33 & 98.5\% & 98.5\% & \textbf{0.20} \\ \cline{2-7}
        & U-Net & 100\%  & 0.55 & 93.6\% & 93.6\% & 0.37\\ \cline{2-7} 
        & YOLO &  100\%  & 0.68 & 93.9\% & 93.9\% & 0.39\\ \hline \hline
PPA2    & Edge &  99.1\%  & 0.49 & 98.9\% & 98.9\% & \textbf{1.03} \\ \cline{2-7} 
        & U-Net & 100\%  & 0.41  & 93.4\% & 93.4\% & 1.16 \\ \cline{2-7} 
        & YOLO &  100\%  & 0.43 & 93.5\% & 93.5\% & 1.10\\ \hline \hline
PPB     & Edge &  100\% & 1.45 & 100\% & 100\% & \textbf{0.29}\\ \cline{2-7} 
        & U-Net &  100\%  & 1.64 & 100\% & 100\% & 0.35 \\ \cline{2-7} 
        & YOLO &  98.4\%  & 1.74 & 92.6\% & 92.6\% & 0.38\\ \hline \hline
\end{tabular}
\end{center}
\end{table}

The results of the pose estimation are given in Table \ref{tbl_HAlt_pnp_success}. The valid PnP output percentage indicates the number of measurements that produced a pose estimation result. The following percentages indicate the number of valid PnP measurements after filtering by the reprojection and distance thresholds. The presented mean $\varepsilon_d$ values are deviations from the Kalman filter pose estimation. Since there are no ground truth positions available for the \textit{H-Alt} model, the $\varepsilon_d$ values serve mainly as an indication of the stability of the pose estimates.

PnP pose estimates from the PPA1 and PPB datasets exhibit significantly higher precision compared to PPA2. We can deduce that this is due to the detected PV structures covering significantly larger areas of the image than in the PPA2 images. Furthermore, power plant B contains five rows in each bench, providing a significantly higher number of reference points for the EPnP algorithm.

The results show that the localization method achieved sufficient precision and robustness in all power plants. The method can effectively compensate for the inaccuracies of the power plant model and the offset and drift of the GNSS. Thus, ensuring precise camera positioning relative to the inspected PV modules through visual feedback.

\subsubsection{Localization Results - \textit{SfM} Model}
\label{sfm_results}

Due to the characteristics of the structure-from-motion process, the created SfM model cannot be accurately matched to the GNSS data. Thus, the velocity estimates are based purely on the Kalman pose estimates. This negatively impacts the results, since the Kalman filter is not able to react to abrupt changes in direction properly. In contrast, it appears that the precision of the power plant model has improved, as the PnP pose estimates exhibit lower variance.

\begin{figure}[thpb]
\centering
\begin{subfigure}[t]{.47\textwidth}
  \centering
  \includegraphics[width=1\textwidth]{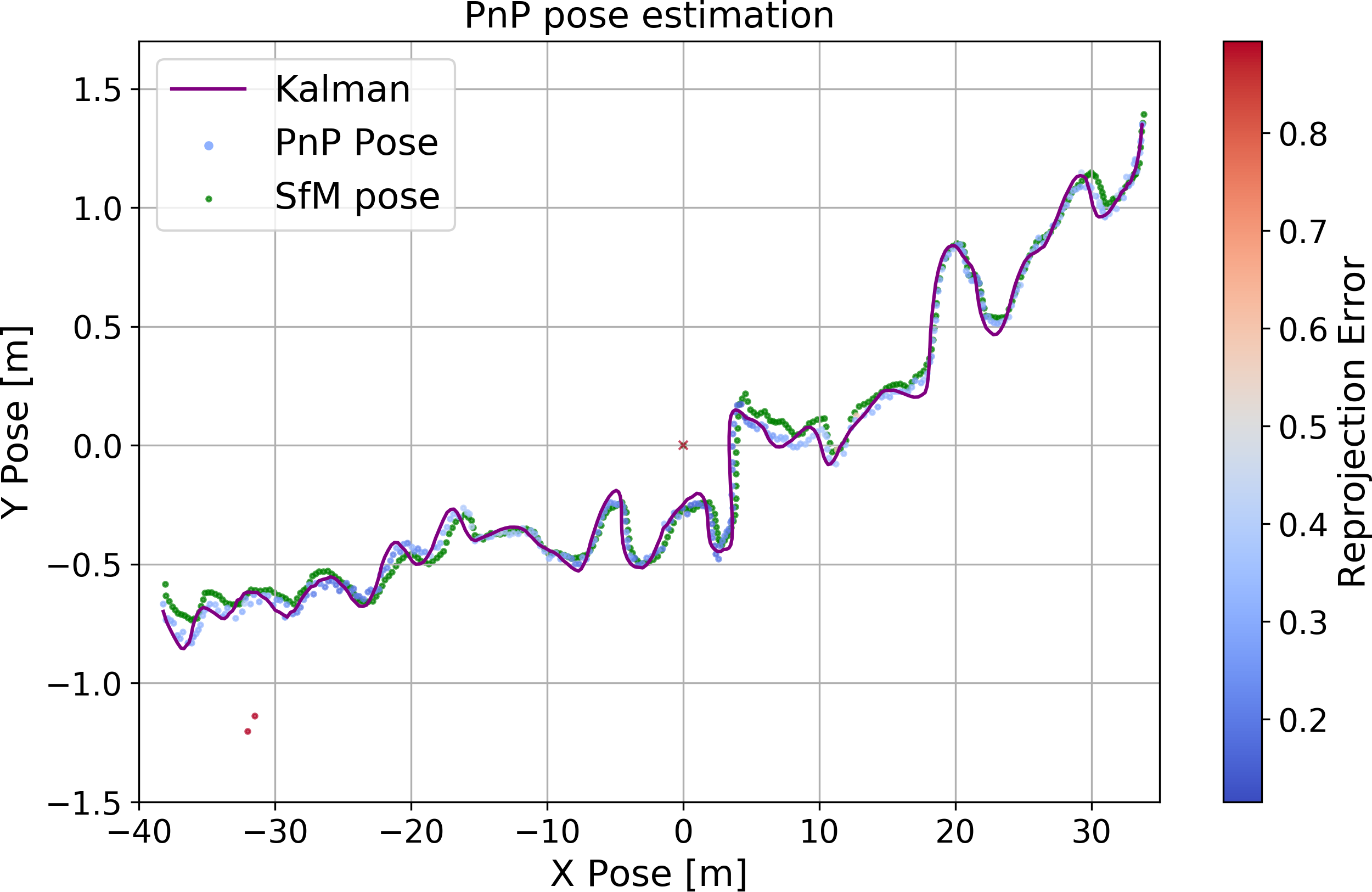}
  \caption{}
  \label{fig:SedgeXY_PPA2}
\end{subfigure}\hspace{3mm}%
\begin{subfigure}[t]{.47\textwidth}
  \centering
  \includegraphics[width=1\textwidth]{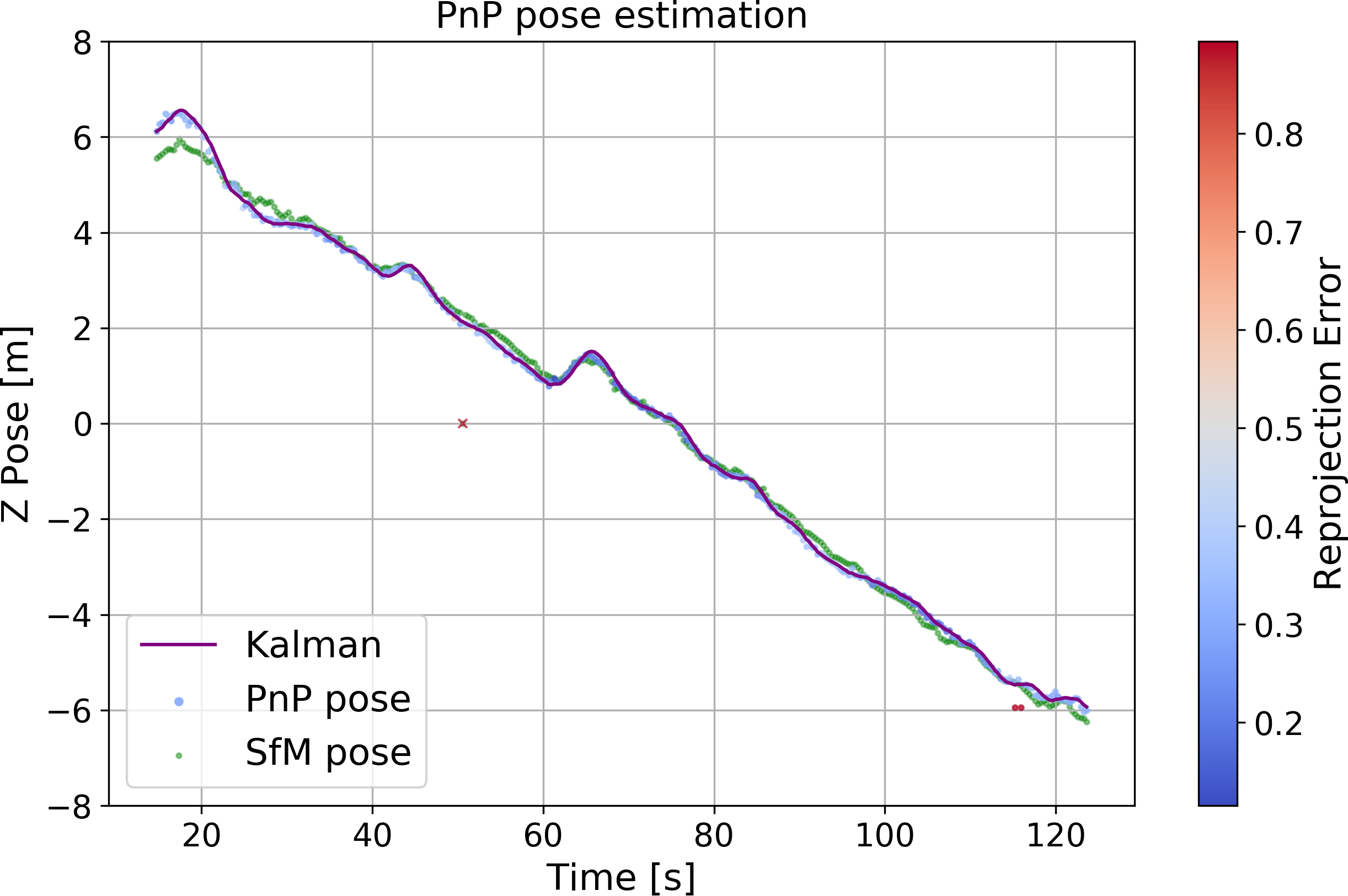}
  \caption{}
  \label{fig:SedgeZ_PPA2}
\end{subfigure}


\begin{subfigure}[t]{.47\textwidth}
  \centering
  \includegraphics[width=1\textwidth]{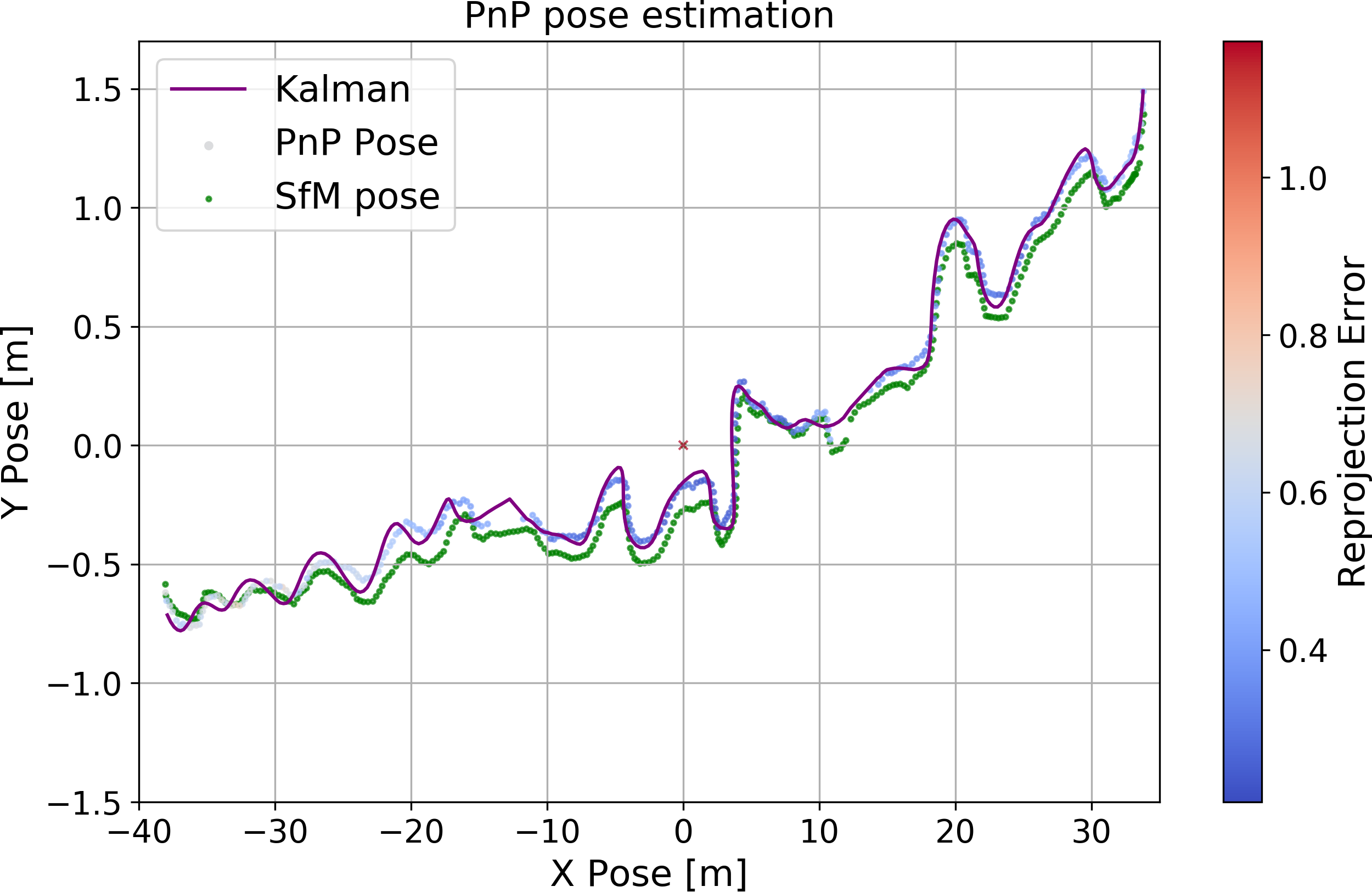}
  \caption{}
  \label{fig:SunetXY_PPA2}
\end{subfigure}\hspace{3mm}%
\begin{subfigure}[t]{.47\textwidth}
  \centering
  \includegraphics[width=1\textwidth]{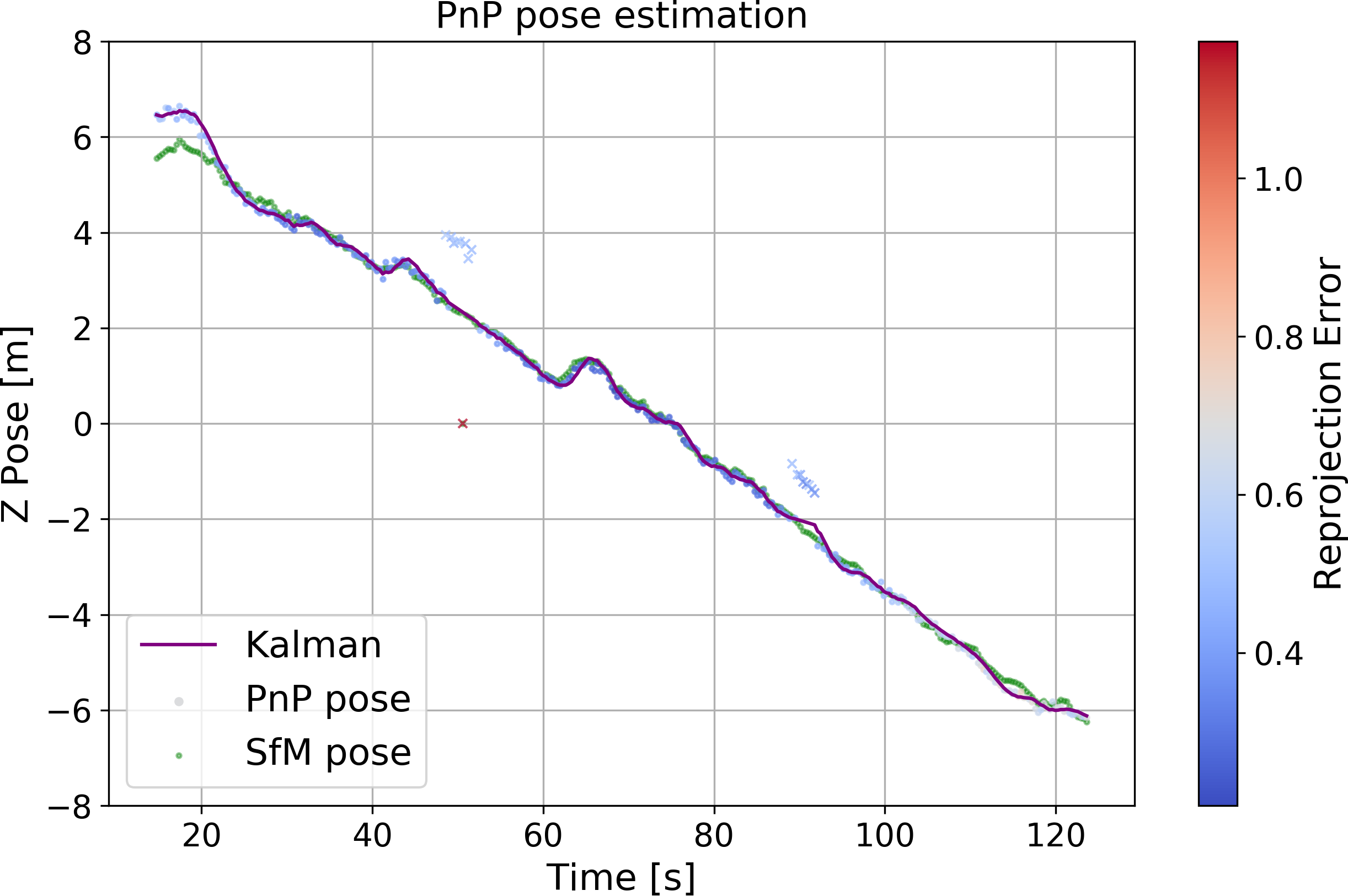}
  \caption{}
  \label{fig:SunetZ_PPA2}
\end{subfigure}


\begin{subfigure}[t]{.47\textwidth}
  \centering
  \includegraphics[width=1\textwidth]{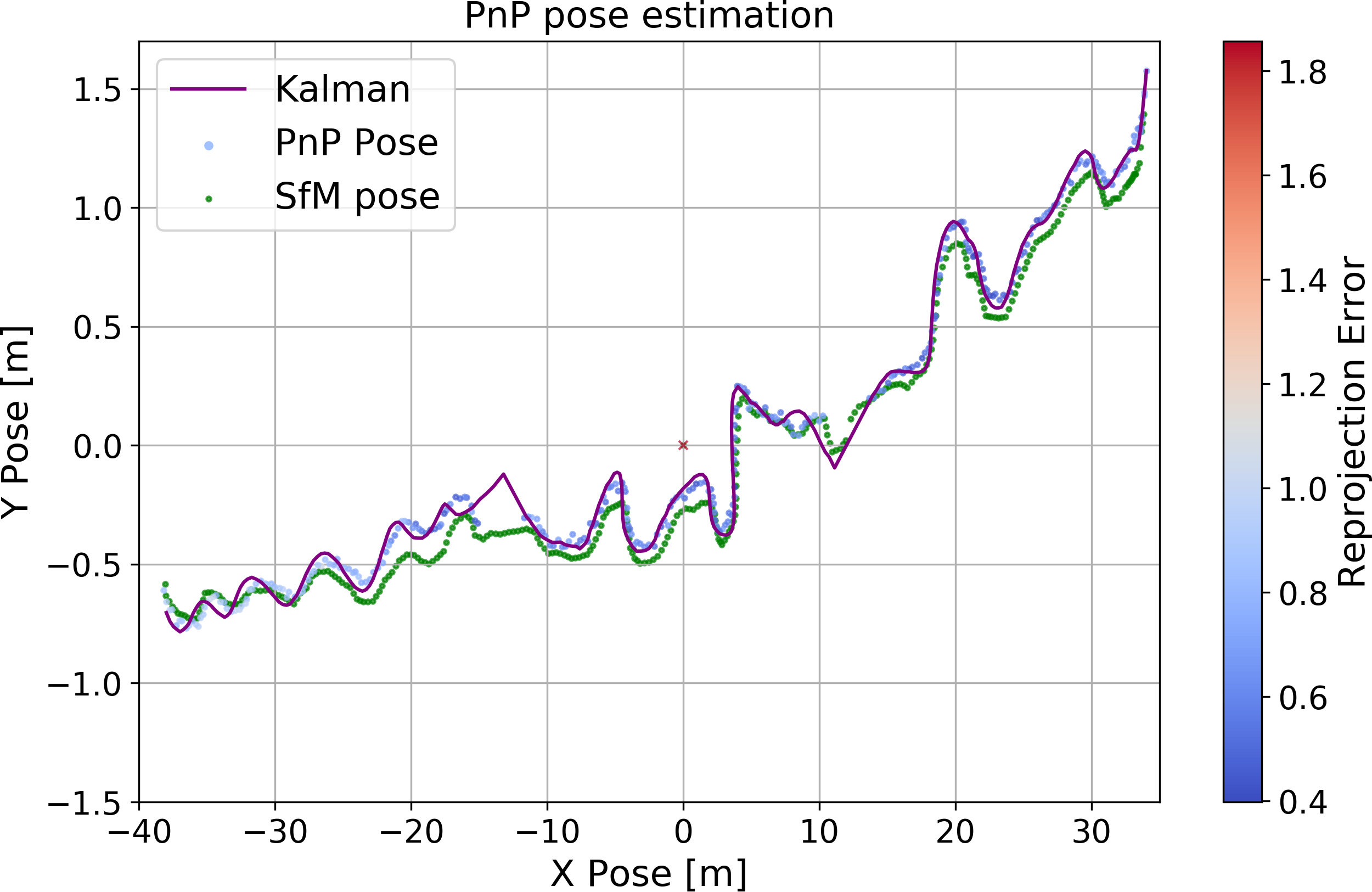}
  \caption{}
  \label{fig:SyoloXY_PPA2}
\end{subfigure}\hspace{3mm}%
\begin{subfigure}[t]{.47\textwidth}
  \centering
  \includegraphics[width=1\textwidth]{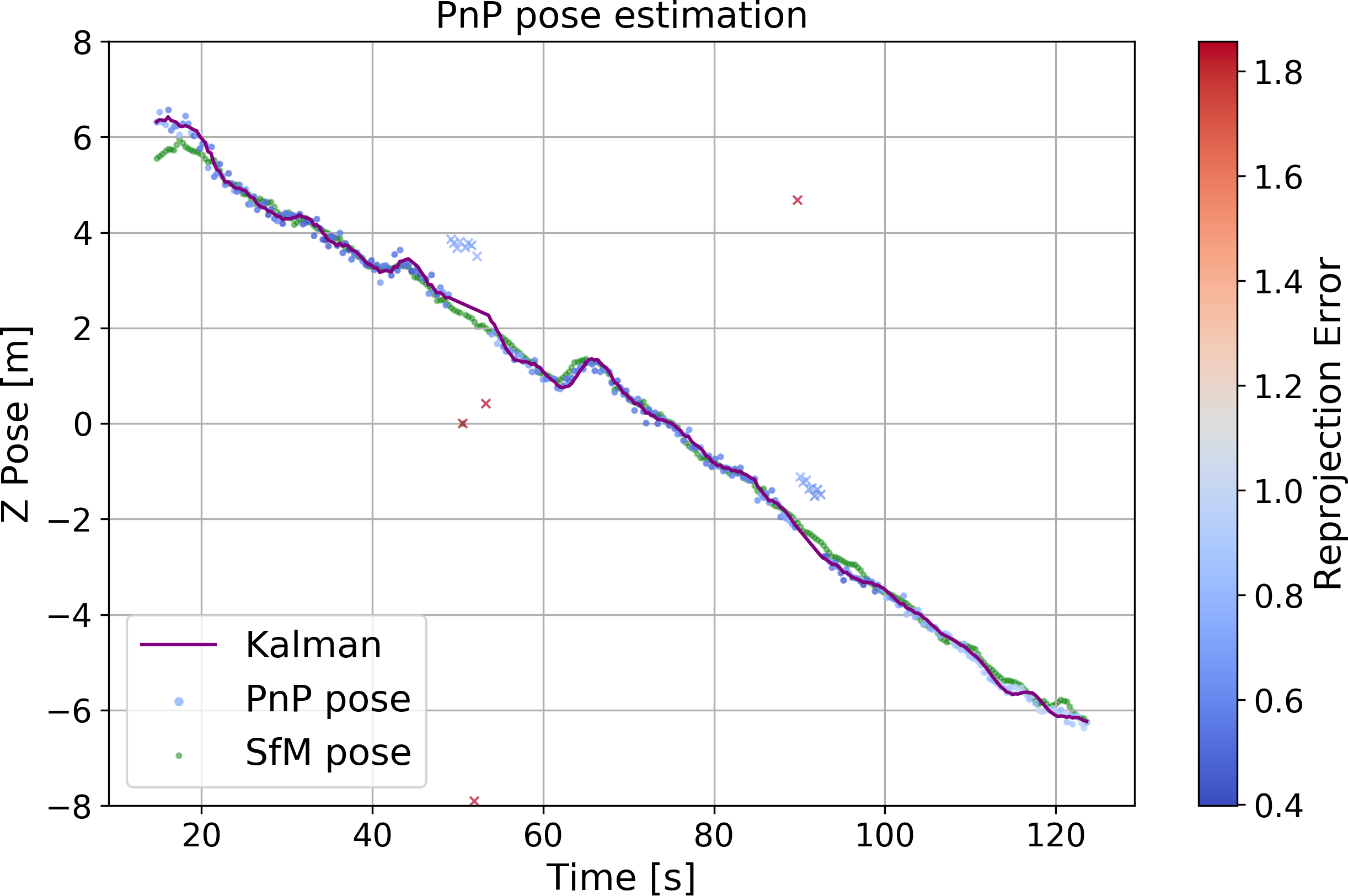}
  \caption{}
  \label{fig:SyoloZ_PPA2}
\end{subfigure}

\caption{PPA1, \textit{SfM} model. The reprojection error is  illustrated using a color map. Figures \textbf{(a)} and \textbf{(b)} present results of the edge-based method. Figures \textbf{(c)} and \textbf{(d)} present results for U-Net. Figures \textbf{(e)} and \textbf{(f)} present results for YOLO.}
\label{fig:Saxes_repr_SfM_PPA2_new}
\end{figure}

Detailed results of pose estimation on the PPA1 dataset are presented in Figure \ref{fig:Saxes_repr_SfM_PPA2_new}. Detailed figures for PPA2 and PPB can be found in \mbox{\ref{apdx_flights_sfm}}.

\begin{figure}[thpb]
\centering

\begin{subfigure}[t]{.47\textwidth}
  \centering
  \includegraphics[width=1\textwidth]{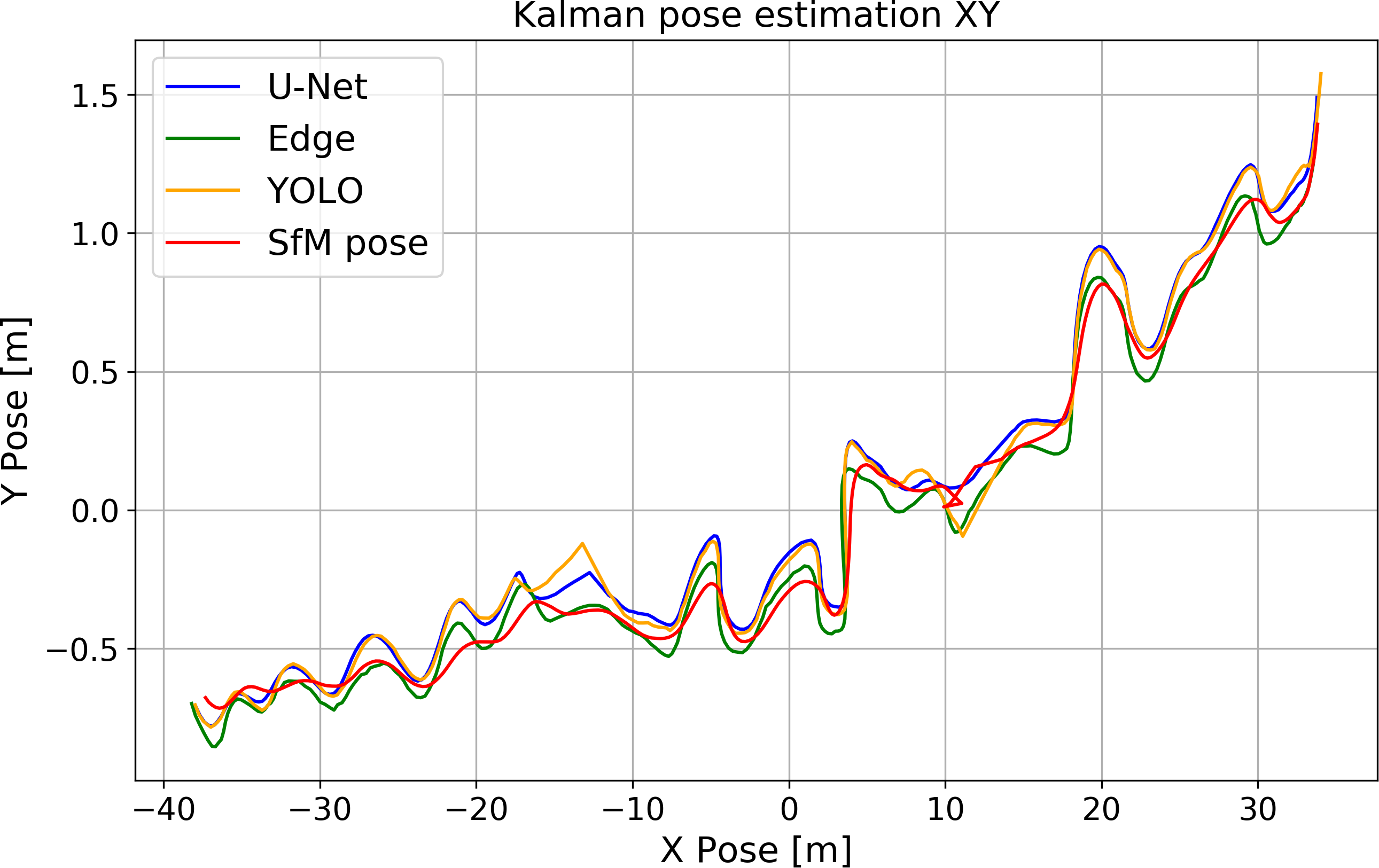}
  \caption{}
  \label{fig:SPPA2_kalmans_A_XY}
\end{subfigure}\hspace{3mm}%
\begin{subfigure}[t]{.47\textwidth}
  \centering
  \includegraphics[width=1\textwidth]{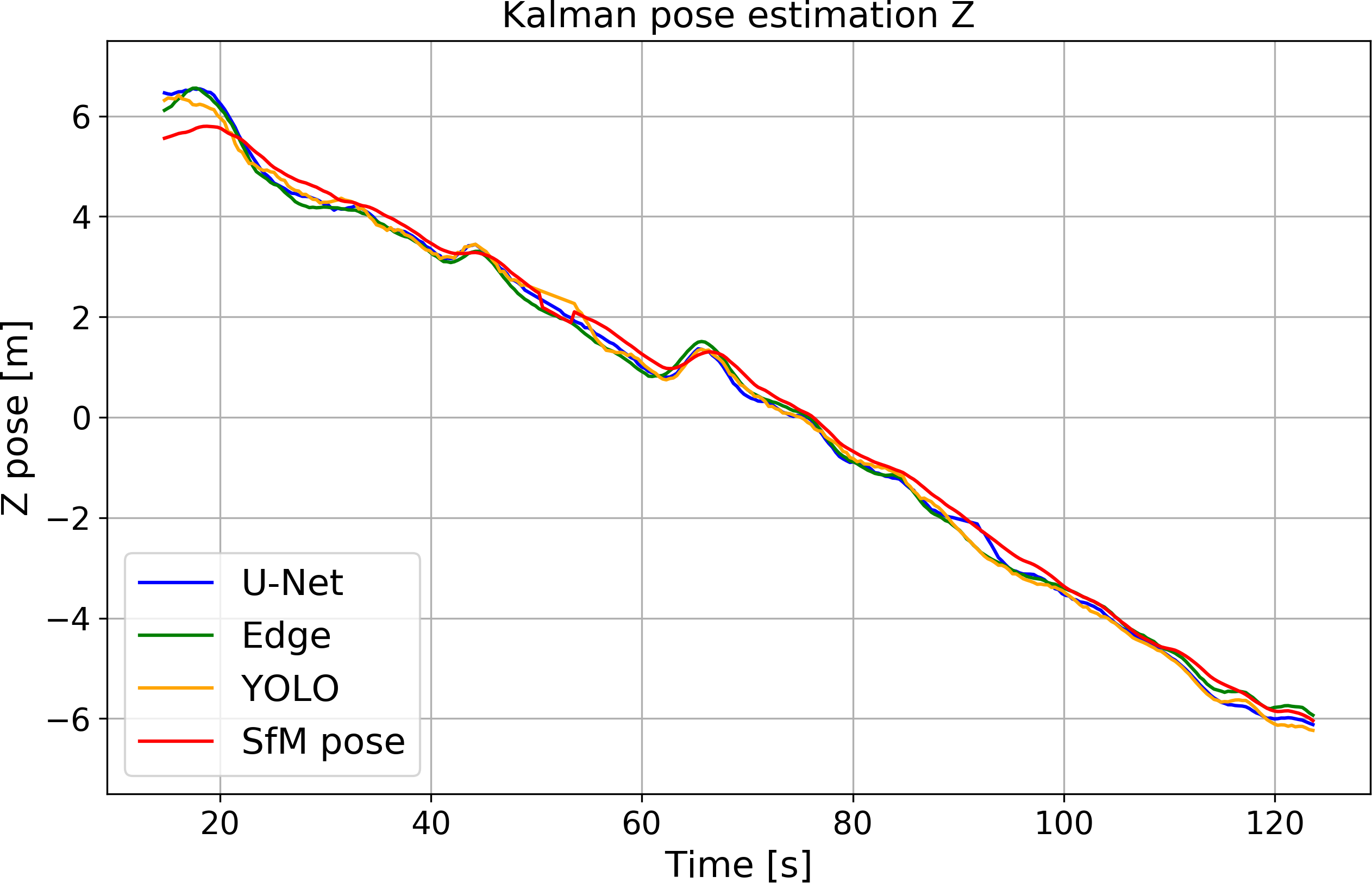}
  \caption{}
  \label{fig:SPPA2_kalmans_A_Z}
\end{subfigure}

\begin{subfigure}[t]{.47\textwidth}
  \centering
  \includegraphics[width=1\textwidth]{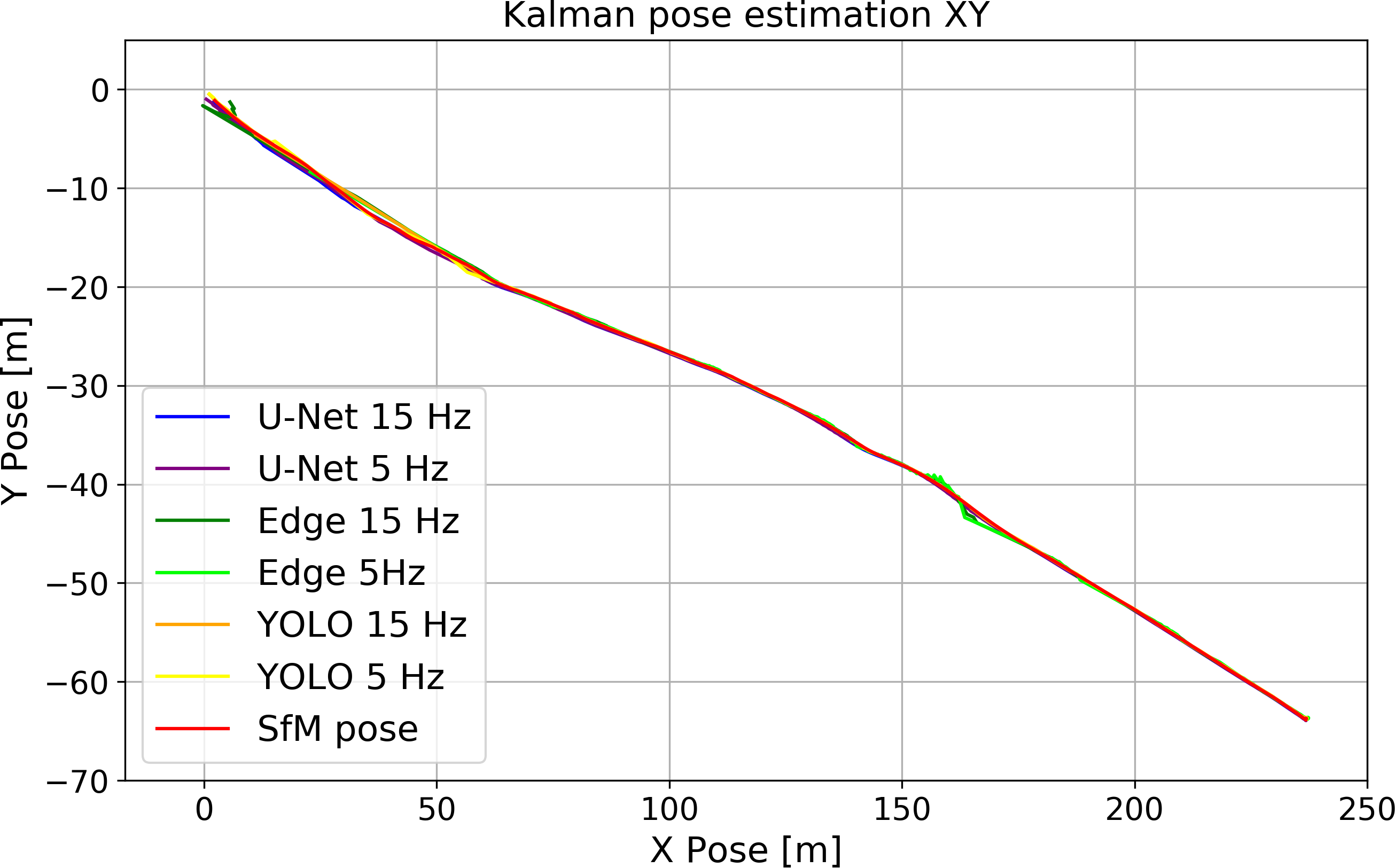}
  \caption{}
  \label{fig:SPPA_kalmans_A_XY}
\end{subfigure}\hspace{3mm}%
\begin{subfigure}[t]{.47\textwidth}
  \centering
  \includegraphics[width=1\textwidth]{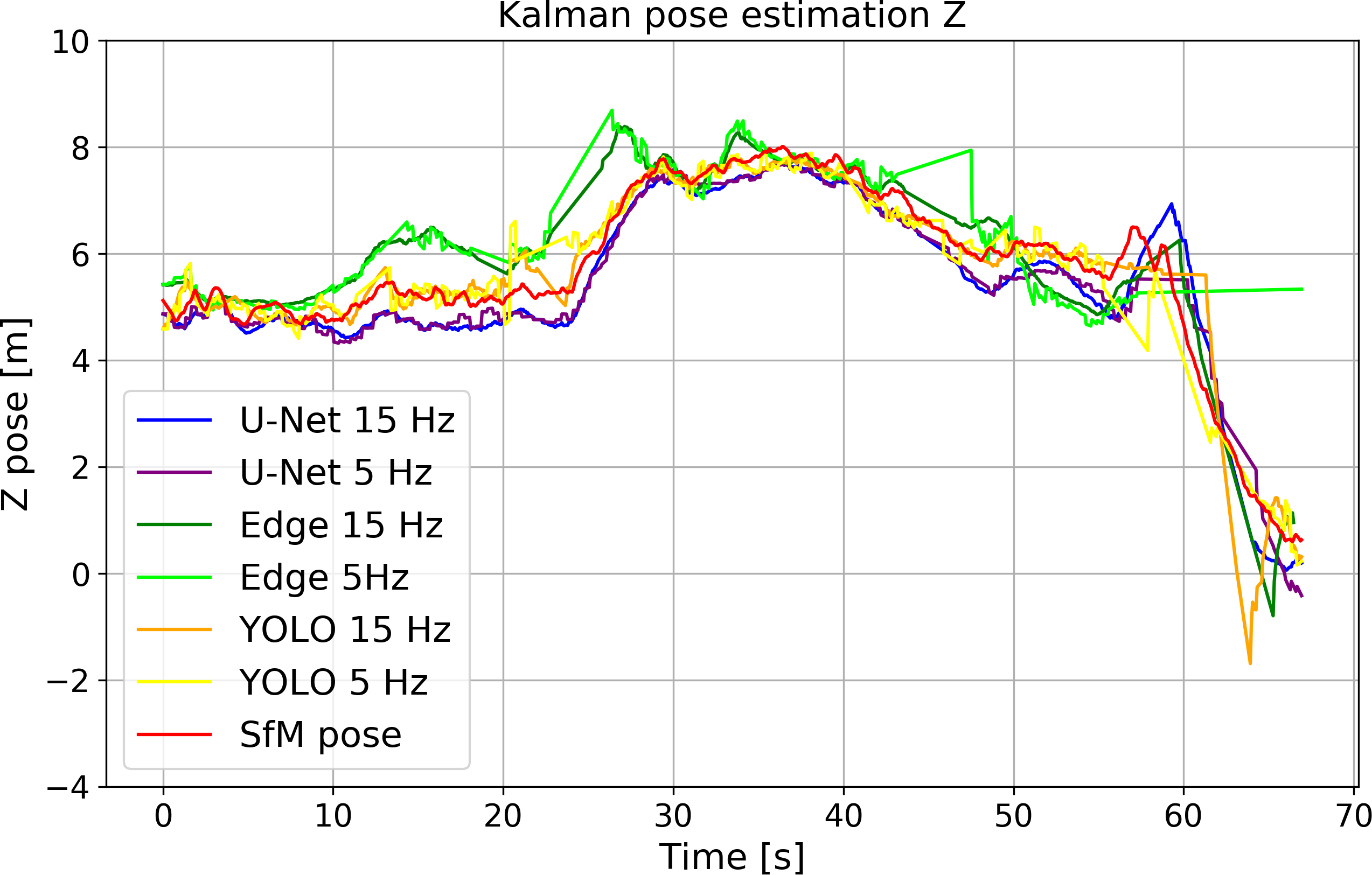}
  \caption{}
  \label{fig:SPPA_kalmans_A_Z}
\end{subfigure}

\begin{subfigure}[t]{.47\textwidth}
  \centering
  \includegraphics[width=1\textwidth]{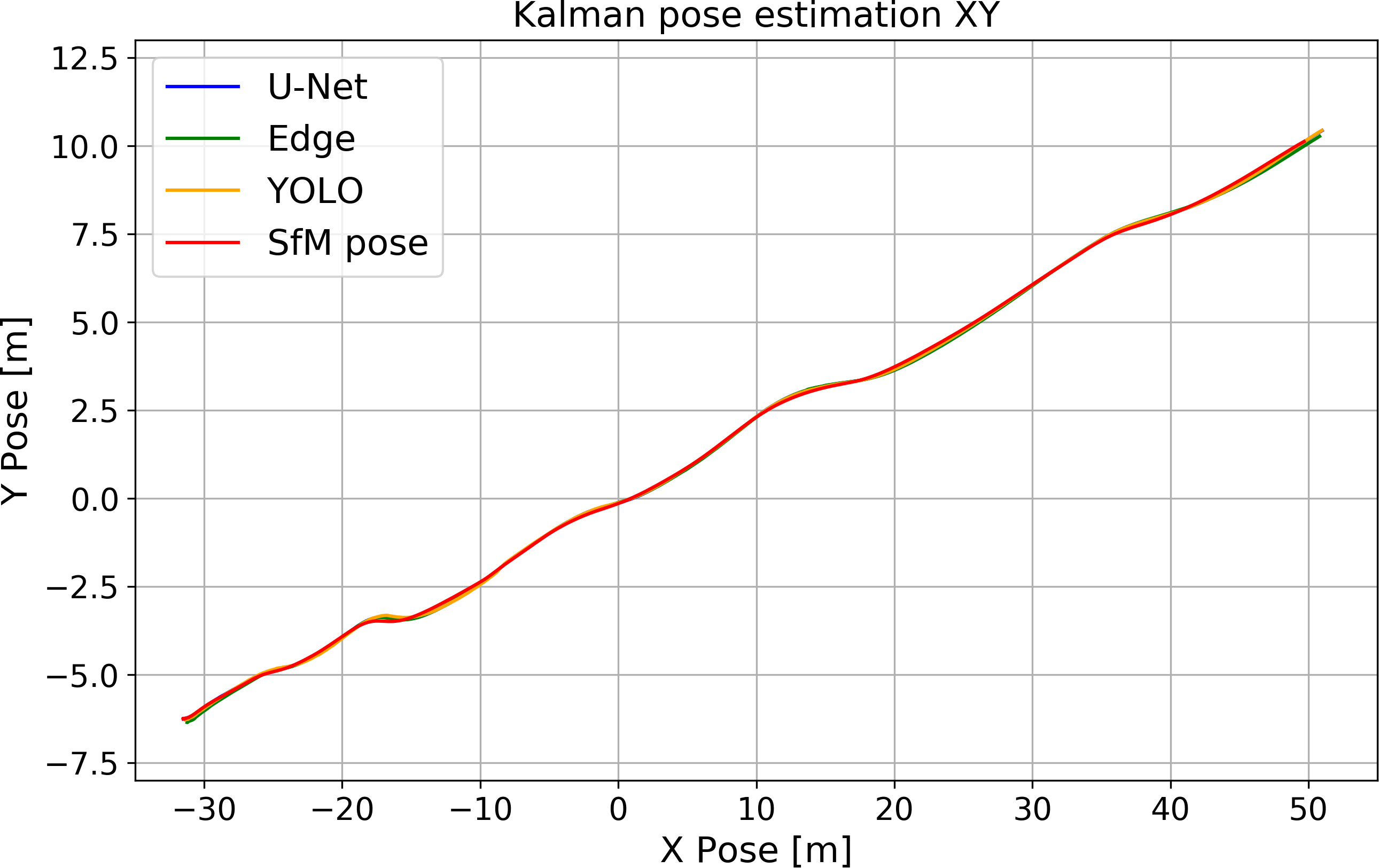}
  \caption{}
  \label{fig:Skalmans_B_XY}
\end{subfigure}\hspace{3mm}%
\begin{subfigure}[t]{.47\textwidth}
  \centering
  \includegraphics[width=1\textwidth]{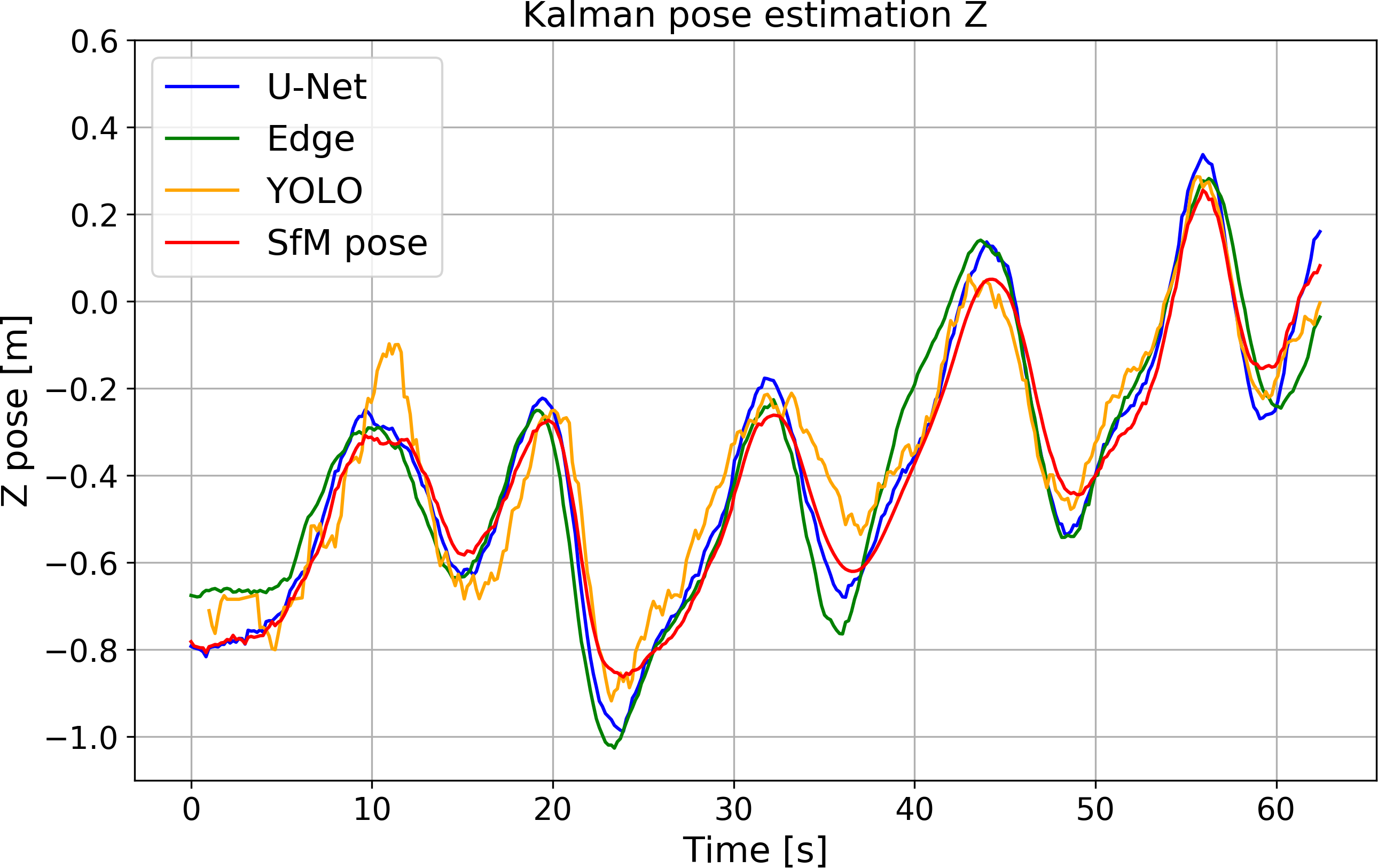}
  \caption{}
  \label{fig:Skalmans_B_Z}
\end{subfigure}

\caption{\textit{SfM} model. Kalman pose estimation for individual methods, together with the SfM ground truth position values. Figures (a) and (b) show results for PPA1. Figures (b) and (c) show results for PPA2. Figures (d) and (e) show results for PPB.}
\label{fig:sfm_kalmans}
\end{figure}

A comparison of the resulting Kalman pose estimates for all datasets is given in Figure \ref{fig:sfm_kalmans}. The SfM ground truth positions are included as a reference. The results include an additional pass for the PPA2 flight with the stream limited to 5 fps.

\begin{table}[h!]
\renewcommand{\arraystretch}{1.3}
\caption{Pose estimation results - \textit{SfM} model type.} 
\vspace{-2mm}
\label{tbl_SfM_pnp_success}
\scriptsize
\begin{center}
\begin{tabular}{|l||c|c|c|c|c|c|c|c|}
\hline
           & Method & PnP & median& $\varepsilon_r < th_r$ & $\varepsilon_d < th_d$ & mean $err$ & mean $err$  & $P_{90} \text{ } err$\\ 
           & & output &  $\varepsilon_r$  [px]&  & & PnP [m] & Kalman [m] & Kalman [m] \\ \hline \hline
PPA1        & Edge & 100\% & 0.298 & 99.1\% & 99.1\% & 1.01 & 0.89 & 1.15\\ \cline{2-9}
            & U-Net & 100\% & 0.391 & 99.1\% & 94.2\% & \textbf{0.90} & 0.77 & \textbf{1.01}\\ \cline{2-9}
            & YOLO & 100\% & 0.619 & 97.9\% & 92.9\% & \textbf{0.90} & \textbf{0.76} & 1.08\\ \hline \hline
PPA2        & Edge & 99.1\% & 0.517 & 84.4\% & 74.6\% & 1.19 & 1.29 & 1.72 \\ \cline{2-9}
            & U-Net & 100.0\% & 0.197  & 92.5\%  & 87.0\% & 1.16 & 0.95 & \textbf{1.22} \\ \cline{2-9}
            & YOLO & 100\% & 0.232 & 83.0\% & 75.0\% & \textbf{0.99} & \textbf{0.88} & 1.35\\ \hline \hline
PPB         & Edge  & 100.0\% &  0.248 & 97.1\% & 97.1\% & 1.11 & 0.85 & 1.27\\ \cline{2-9}                       
            & U-Net & 100.0\% & 0.131  & 100.0\% & 100.0\% & \textbf{0.89} & 0.68 & \textbf{1.19} \\ \cline{2-9}
            & YOLO & 98.4\% & 0.311 & 90.2\% & 90.2\% & 0.95 & \textbf{0.65} & 1.20\\ \hline 
\end{tabular}
\end{center}
\end{table}

\begin{figure}[h!]
    \centering
\begin{subfigure}[t]{1\textwidth}
  \centering
  \includegraphics[width=1\textwidth]{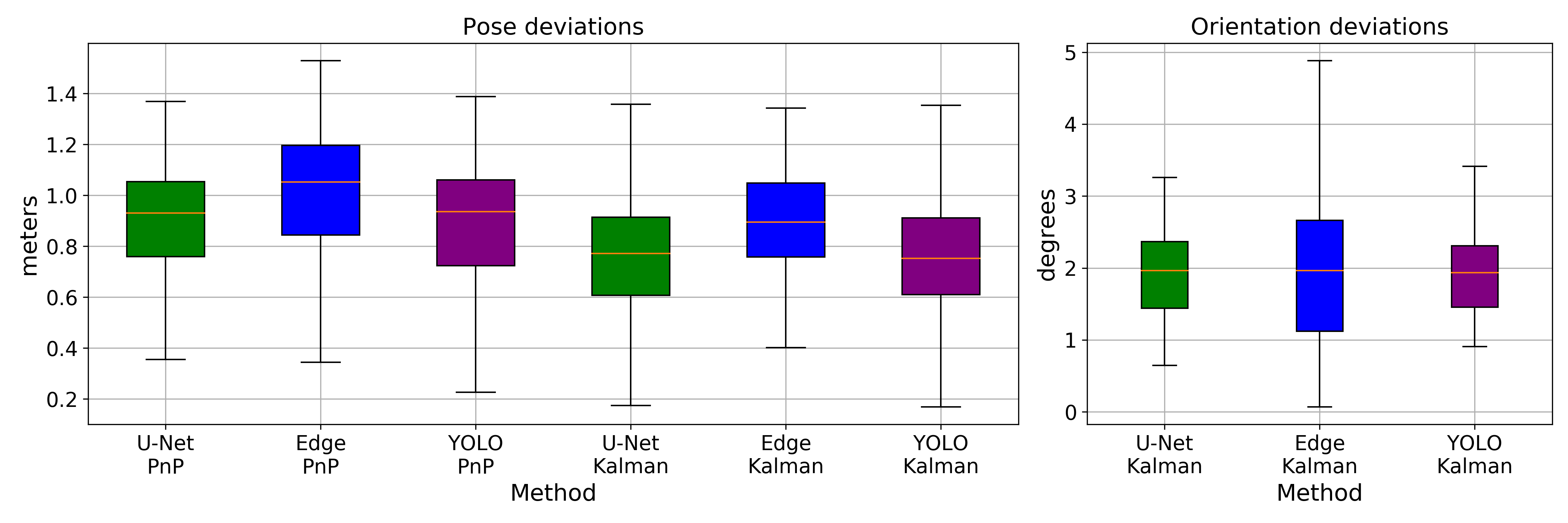}
  \caption{}
  \label{fig:DevPPA2}
\end{subfigure}

\begin{subfigure}[t]{1\textwidth}
  \centering
  \includegraphics[width=1\textwidth]{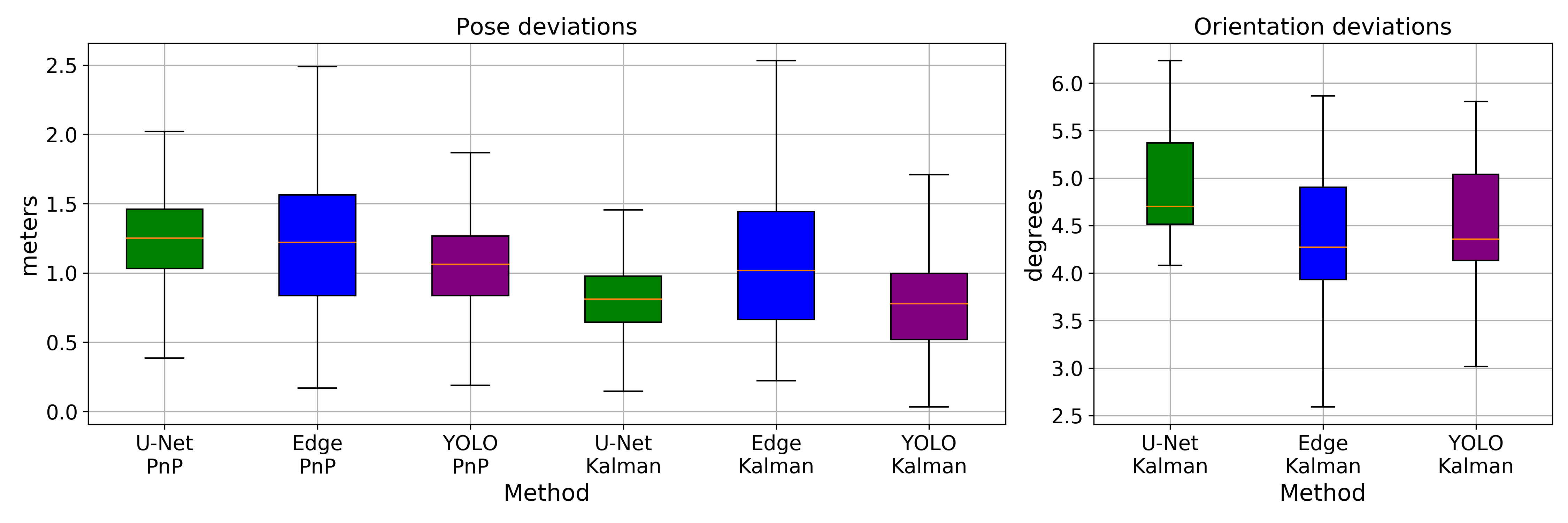}
  \caption{}
  \label{fig:DevPPA1}
\end{subfigure}

\begin{subfigure}[t]{1\textwidth}
  \centering
  \includegraphics[width=1\textwidth]{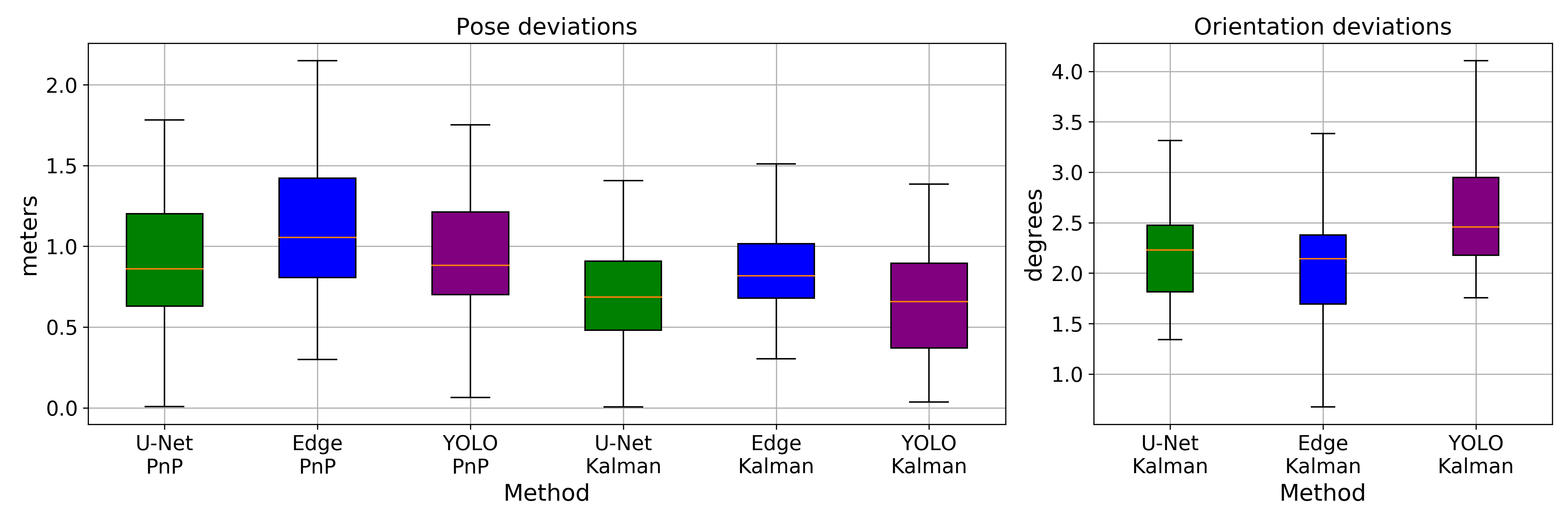}
  \caption{}
  \label{fig:DevPPB}
\end{subfigure}\hspace{3mm}%

    \caption{Five point statistics of position errors for Kalman pose estimates and valid PnP measurement subsets. The presented orientation deviations are calculated from the Kalman orientation estimates using the angular distance. \textbf{(a)} PPA1. \textbf{(b)} PPA2. \textbf{(c)} PPB.}
    \label{fig:deviation_boxplots}
\end{figure}

The pose estimation results are given in Table \ref{tbl_SfM_pnp_success}. The optimized camera positions estimated using the OpenSfM library \citep{opensfm} are used as ground truth to evaluate the accuracy of the presented localization methods. 
\mbox{Figure \ref{fig:deviation_boxplots}} presents five point statistics of position deviations for all datasets. The angular distance is used as a metric to measure the difference between the estimated and ground-truth orientation in a three-dimensional space.

The results show that the proposed visual localization method can track the UAV trajectory even without GNSS information. The localization accuracy is sufficient for the intended inspection pipeline. The orientation error is relatively minor and can be further mitigated by using a gimbal to align the detected structures with the vertical center of the image.

\subsection{Computation Times}
\label{comp_times}

\begin{table}[h!]

\renewcommand{\arraystretch}{1.3}
\caption{Average computation times [ms]. Note that the CNN segmentation step is optional for the edge-based method.}
\vspace{-2mm}
\label{tbl_comp_times}
\scriptsize
\begin{center}
\begin{tabular}{|l||c|c||c|c||c|c|}
\hline
     & \multicolumn{2}{c||}{Edge} & \multicolumn{2}{c||}{U-Net} & \multicolumn{2}{c|}{YOLOv8n-seg} \\ \hline 
    & CPU & GPU & CPU & GPU & CPU & GPU \\ \hline \hline
CNN segmentaion & 24 & 16 & 39 & 25 & 112 & 36 \\ \hline
Traditional CV  &  \multicolumn{2}{c||}{34}  &  \multicolumn{2}{c||}{5} & \multicolumn{2}{c|}{1} \\ \hline
Tracking & \multicolumn{6}{c|}{3} \\ \hline
PnP & \multicolumn{6}{c|}{5}  \\ \hline \hline
Total & 66 & 58 & \textbf{52} & \textbf{38} & 121 & 45 \\ \hline
\end{tabular}
\end{center}
\end{table}

The computation times for individual steps in the localization pipeline are given in Table \ref{tbl_comp_times}. The times were measured on a computer with an Intel Core i7-7700 @ 3.60 GHz processor, 64 GB RAM, and an Nvidia GeForce 2080Ti GPU. The times given for the tracking and PnP steps are independent of the used segmentation method. We also offer a comparison of CNN computation times for both GPU and CPU, which may be crucial, depending on available hardware.

\subsection{Results and Discussion}
\label{results_and_discussion}

One of the main contributions of the presented work is the ability to directly utilize PV module detections for UAV localization on the level of individual modules. Thus, providing direct feedback on their position relative to the UAV. Having an object-oriented navigation method gives us the possibility of obtaining high-resolution pictures of the PV modules. That is, we can navigate the drone so that a desired PV module spans a significant portion of the captured image. If an anomaly detection procedure is run simultaneously with the navigation pipeline, a detailed image of all potentially faulty PV modules can be acquired autonomously. Similarly, the system can be used to monitor the deterioration of known faulty modules during regular inspection flights.

The localization pipeline was successfully verified using three flight scenarios on two different power plants. All segmentation methods achieved a satisfactory success rate in PV module detection and provided sufficient precision for the inspection task while enabling real-time application. The combination of the PnP method with the Kalman filter enabled robust localization, which was tested using different image processing frequencies. All presented segmentation methods support direct image patch extraction for anomaly detection. Additionally, the localization pipeline allows for seamless alternation between these methods, enhancing its versatility and practicality.

\subsubsection{Power Plant Models}

The proposed method relies on a compact georeferenced model of the power plant. In contrast, traditional environment models for visual navigation typically require considerable memory and substantial effort to construct and are prone to appearance changes caused by illumination or weather variations.

The methods were verified using two types of power plant models. The \textit{H-Alt} model has generated from publicly available high-altitude aerial images and topographic maps. This represents a realistic use-case scenario in which the creation of the model is independent of the inspection flight. 
However, since the GNSS data from the UAV flight are not synchronized with the model, the resulting flight can diverge from the desired trajectory. A similar situation may arise if the power plant model is inaccurate. Such situations can negatively affect the inspection procedure. 
The proposed localization method successfully compensates for these inaccuracies by providing position estimates relative to the inspected PV modules. 

The \textit{SfM} model was created from images acquired during the experiment flights using structure-from-motion. This model enabled us to evaluate the precision of the proposed methods using optimized camera position estimates from structure-from-motion as ground truth. However, it must be emphasized that using such a model is not a realistic use-case scenario, since its creation is highly time-consuming and unfeasible for larger power plants.

\subsubsection{Segmentation Methods}



We have presented three different PV module segmentation approaches which were verified and evaluated independently within the localization pipeline.

The U-Net segmentation method was shown to be the most robust in terms of the success rate in detecting PV modules and bench gaps in the test images. 
Its computation complexity is also lower than for the other segmentation methods.

The main advantage of YOLO is its straightforward usability. Given a sufficient amount of training data, it should be applicable to a wide variety of objects. 
The results for YOLO for power plant A are comparable to U-Net. However, the method suffered from a decrease in performance for power plant B. We attributed this to the underrepresentation of photovoltaic installations with horizontally arranged modules in the training data. It also relies on GPU for fast computation.

The results of the edge-based approach are satisfactory for PPA1 and PPB. However, it experienced a decline in detection success within the PPA2 dataset, when encountering a darker type of PV modules. This is likely due to the fixed parameter tuning, which is not optimized for varying module types.

Both CNN-based methods can be used with minimal parameter optimization and are more resistant to imprecisions. 
However, they require domain-specific training data, which makes their application to a different type of PV modules tedious. 
In comparison, the edge-based method relies only on parameter tuning, without the need for a new training dataset. It requires a precise camera calibration for image rectification and the individual parameters of the image processing steps must be properly optimized. Nevertheless, it is possible to transfer this method to varying types of PV modules, as long as they can be represented as a graph composed of intersections of straight lines.


The results presented in this paper focus on the performance of these segmentation methods within the proposed localization pipeline. Both the trained CNN segmentation models and the training dataset were made publicly available. Additionally, the same U-Net and YOLO models were employed in our other work on PV power plant mapping from higher altitudes (60-100 meters) \citep{kozakMappingPV}, where we verified their applicability at larger imaging distances.

\subsubsection{System Limitations and Inspection Scenario Design}

The experimental results enable us to identify several limitations in the model and the localization method, hinting at the need for a suitable flight scenario design. The results for the PPA2 dataset are noticeably worse than for the remaining two datasets. The high position deviation values ($\varepsilon_d$) for PPA2 when using the \textit{H-Alt} model type also indicate a possible imprecision in the model. Especially in comparison with the \textit{SfM} model type, for which the localization pipeline achieved higher precision. From this we can conclude that inconsistencies in the power plant model can detrimentally affect the PnP method. 

In comparison, localization achieved significantly better performance in the PPA1 and PPB datasets, even though the \textit{H-Alt} model creation process was identical. This indicates that a well-designed scenario can mitigate the influence of the imprecision of the model. The PPA1 flight was performed from closer proximity to the modules. The PPB PV installation contains 5 rows on each bench, providing a higher number of reference points for the EPnP algorithm. In both cases, the detected PV structures cover significantly larger areas of the image than in the PPA2 images. Future work could focus on improving navigation stability by incorporating multiple rows (benches) during the inspection process, as initial results suggest that this might enhance performance compared to using only a single row.

In this work, localization is performed relative to the repeating semantic structures in the row layouts commonly found in PV plants. However, in principle, the method can be applied to other non-row layouts, e.g., to rooftop installations.

\section{Conclusions}
\label{conclusions}

We present a novel approach to real-time UAV localization for autonomous inspection of PV power plants. The method directly utilizes PV module detections for UAV localization, enabling autonomous navigation and optimal camera positioning during data acquisition. Segmented image patches with individual modules are associated with the power plant model and can simultaneously be used for defectoscopy.

We have proposed two methods for visual segmentation of PV modules and utilized one additional state-of-the-art method. We have compared the three methods in terms of their performance in relation to the proposed navigation pipeline. 
We also identified potential shortcomings regarding robustness to adversarial conditions and other limitations.

The experimental results demonstrated the effectiveness of the proposed method in accurately localizing UAVs in real-time during PV power plant inspections. It allows for optimal positioning of the UAV and has the potential to improve the efficiency of UAV inspection of PV power plants. 
The method can also be extended to other types of structures beyond PV modules, further expanding its potential applications.

With regard to practical application, we propose the use of visually identifiable navigation anchor points used for the initial association of the detected modules with the power plant model and introduce a method for their detection. 
The method operates with a compact georeferenced power plant model, avoiding the need for extensive appearance-driven environment representations typically used in vision-based localization.
We utilize two distinct approaches to power plant model generation and identify the limitations of the model and the inspection flight scenarios. 


The presented methods are being integrated into a complex system that will encompass a spatiotemporal model for the diagnostic and predictive maintenance of photovoltaic power plants. Future work will focus on the integration of visual and thermal defectoscopy directly into the inspection pipeline.

\section*{CRediT authorship contribution statement}

\textbf{Viktor Kozák:} Conceptualization, Methodology, Software, Validation, Data curation, Writing - Original Draft, Review \& Editing, Visualization. 
\textbf{Karel Košnar:} Conceptualization, Methodology, Software, Investigation, Writing - Review \& Editing. 
\textbf{Jan Chudoba:} Methodology, Software, Resources, Data curation. 
\textbf{Miroslav Kulich:} Supervision, Writing - Review \& Editing.
\textbf{Libor Přeučil:} Supervision, Funding acquisition.

\section*{Declaration of competing interest}
The authors declare that they have no known competing financial interests or personal relationships that could have appeared to influence the work reported in this paper.

\section*{Acknowledgement}

This work was co-funded by the European Union under the project ROBOPROX (reg. no. CZ.02.01.01/00/22\_008/0004590).

\section*{Declaration of generative AI and AI-assisted technologies in the writing process}

During the preparation of this work, the authors used Writefull and ChatGPT AI-based software tools for grammar checks and minor text refinements. After using these tools, the authors reviewed and edited the content as needed and take full responsibility for the content of the publication.



  \bibliographystyle{elsarticle-num} 
  \bibliography{bibliography_abbr}








\appendix

\section{Photovoltaic Module Detection in Thermal Images}
\label{apdx_thermal}

\begin{figure}[h!]
\centering
\begin{subfigure}[t]{.24\textwidth}
  \centering
  \includegraphics[width=1\textwidth]{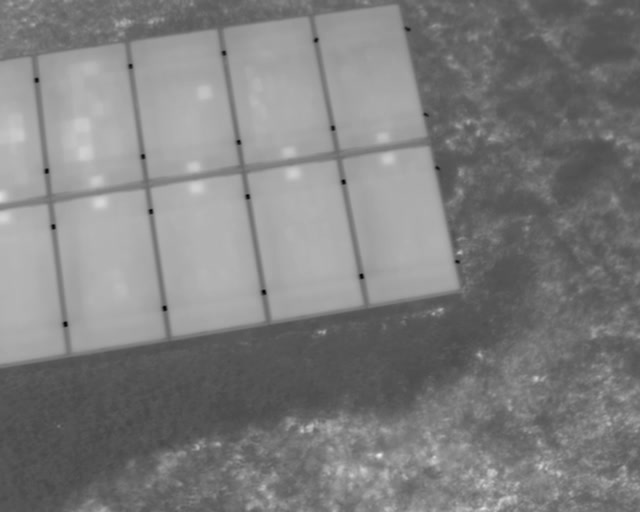}
  \caption{}
  \label{fig:t_orig}
\end{subfigure}\hspace{1mm}%
\begin{subfigure}[t]{.24\textwidth}
  \centering
  \includegraphics[width=1\textwidth]{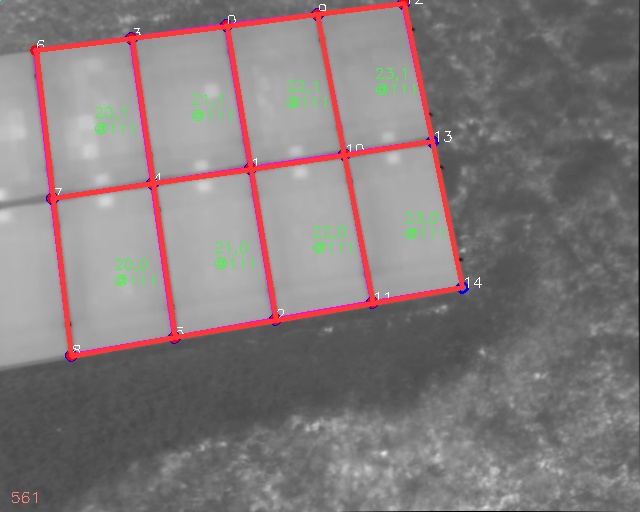}
  \caption{}
  \label{fig:t_unet}
\end{subfigure}\hspace{1mm}%
\begin{subfigure}[t]{.24\textwidth}
  \centering
  \includegraphics[width=1\textwidth]{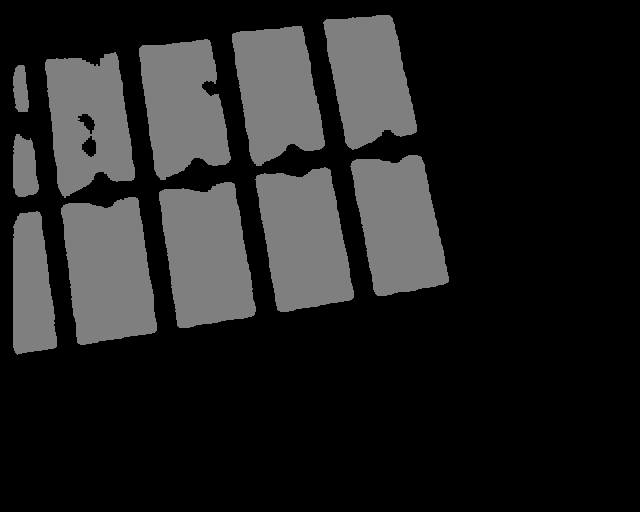}
  \caption{}
  \label{fig:t_orig2}
\end{subfigure}\hspace{1mm}%
\begin{subfigure}[t]{.24\textwidth}
  \centering
  \includegraphics[width=1\textwidth]{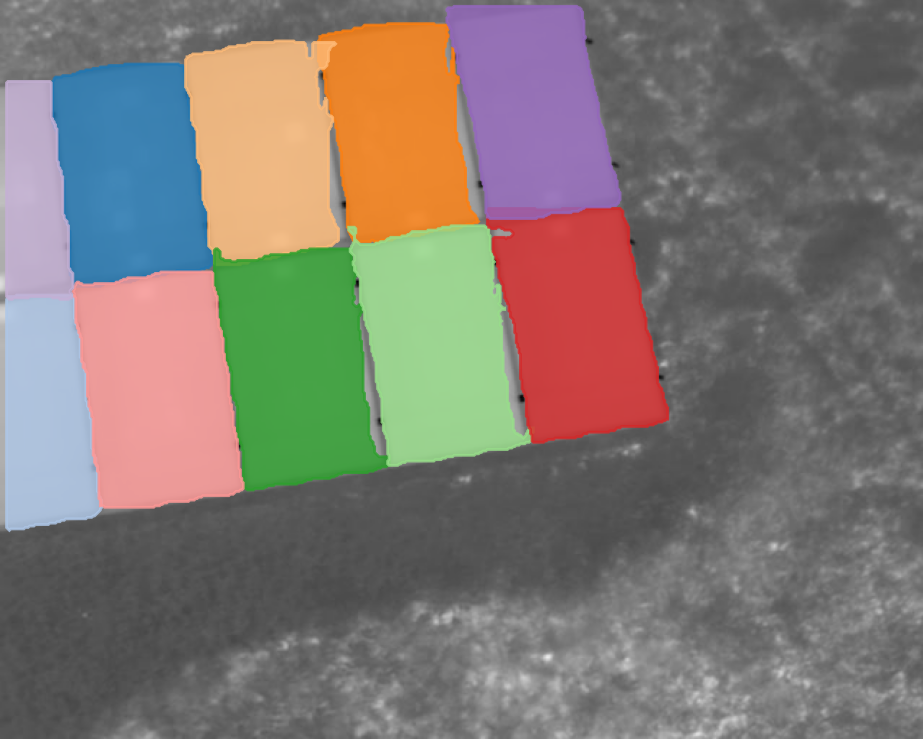}
  \caption{}
  \label{fig:t_unet2}
\end{subfigure}

\caption{The segmentation output for thermal data. \textbf{(a)} Original image. \textbf{(a)} Edge-based segmentation. \textbf{(c)} The U-Net segmentation mask. \textbf{(d)} YOLO masks.}
\label{fig:thermal_detections}
\end{figure}

This section presents a proof of concept for the use of the proposed segmentation methods in thermographic data (Figure \ref{fig:thermal_detections}). This covers PV module detection and segmentation, combined with module tracking and association with the power plant model. 
The CNN-based detection methods were retrained on a new dataset with thermal images. The parameters of the edge-based methods were manually adjusted to adapt to this new domain.

\begin{table}[h!]
\renewcommand{\arraystretch}{1.3}
\caption{Success rates for detection of PV modules and bench gaps on the thermal dataset.}
\vspace{-2mm}
\label{tbl_detection_success_thermo_row22}
\scriptsize
\begin{center}
\begin{tabular}{|l|c|c|c|}
\hline
        & Edge & U-Net & YOLO\\ \hline 
Modules   & 98.6\% & 99.1\% & 98.1\% \\ \hline
Gaps      & 86.3\% & 90.0\% &  68.1\% \\ \hline
\end{tabular}
\end{center}
\end{table}

The testing dataset with thermal images was acquired from power plant A during an autonomous flight using the proposed localization and inspection pipeline. It covers a row with 3 benches and 144 PV modules. It consists of 431 thermal images with a total of 4590 possible PV module detection instances and 110 bench gap detection instances. The results of the detection of PV modules in the thermal dataset can be seen in Table \ref{tbl_detection_success_thermo_row22}. All segmentation methods produced usable results and the achieved success rate is comparable to RGB data.

\section{Edge-based Semantic Structure Detection}
\label{a_edge_based}


The method utilizes perpendicular structures typically present in power plant installations. 
The frames of common PV modules are made from aluminum alloy that contrasts significantly with the black surface of PV cells. These form distinct edges that can be easily detected. 
We start by undistorting the image, ensuring the straightness of the lines. 
To make the detection more robust and consistent, the input image is blurred by Gaussian blur. 
The kernel size of the Gaussian blur is automatically adapted to achieve the same focus measure value over time. 
The focus measure is estimated as the variance of the intensity over the image. 
The kernel size is increased until the focus measure meets the requested value. The Canny edge detector is used to detect edges in the processed image.

Straight lines formed by the structure of PV modules are detected by the Hough detector. 
These lines are filtered according to their weight, which is related to the quality and length of the line. 
Two thresholds are chosen: for the longer horizontal lines and for the vertical lines. 
Multiple lines are typically detected for each physical edge of the PV modules.

Detected lines are further clustered using hierarchical clustering as the number of clusters is a priori unknown. 
The Hausdorff distance is used to measure the distances between two lines $L$ and $K$. 
We identify a set of border points $l_0, l_1, k_0$, and $k_1$ created by the intersections between each line and the image border. 
The Hausdorff distance is the largest Euclidean distance $d$ of one line's border point to the other line.

\begin{equation}
\label{eq:hausdorff_distance}
    H(L,K) =  \max(d(l_0,K),d(l_1,K), d(k_0,L), d(k_1,L)),
\end{equation}

The clustering is stopped when the inter-cluster distance reaches a given threshold.
Each cluster is represented by a single line in further steps.
This representing line is computed as the vector mean of all lines.

\begin{figure}[thpb]
\centering
\begin{subfigure}[t]{0.55\textwidth}
  \centering
  \includegraphics[width=1\textwidth]{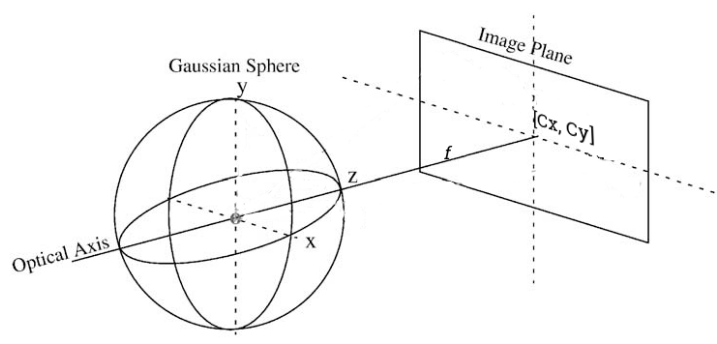}
  \caption{}
  \label{fig:sphere}
\end{subfigure}\hspace{1mm}%
\begin{subfigure}[t]{0.43\textwidth}
  \centering
  \includegraphics[width=1\textwidth]{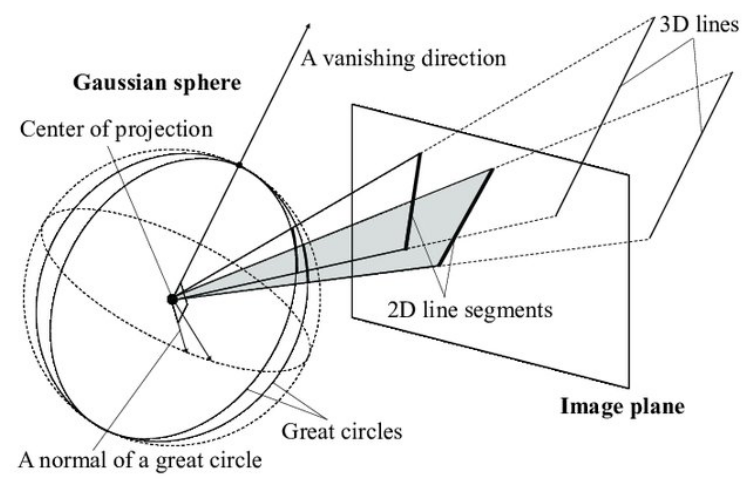}
  \caption{}
  \label{fig:lines_on_sphere}
\end{subfigure}

\caption{ \textbf{(a)} Placement of the image plane relative to the Gaussian sphere. \textbf{(b)} Projection of lines on the Gaussian sphere and vanishing point. The images originate from \citep{Kosnar2019}.}
\label{fig:gaussian_sphere}
\end{figure}

The perpendicular structure of PV modules is used to filter out lines that are not defined by module edges. 
The representation of lines as great circles on the Gaussian sphere allows us to detect parallel and perpendicular lines, even when distorted by perspective projection (Figure \ref{fig:gaussian_sphere}).

Lines that are parallel in the real world have a single intersection on an image plane - the vanishing point, which can be detected on the Gaussian sphere as an intersection of respective great circles. 
When there are two sets of parallel lines, the vectors from the origin of the Gaussian sphere to their vanishing points are perpendicular, if the lines are perpendicular in the real world. 
Lines that exhibit no signs of perpendicularity are discarded from further processing.

The filtered lines are converted to an undirected graph $G (V, E)$ where the intersections of the lines form the vertices $V$ of the graph. 
The edge $e(v_i^l, v_{i+1}^l) \in E$ connects two vertices $v_i^l, v_{i+1}^l$ if both vertices lie on the same line and no other vertex is on the line between them.

Graph isomorphism is used to detect sub-graphs that represent individual PV modules. 
The PV module's sub-graph has exactly four vertices and four edges, where edges form a simple cycle.  
Detected sub-graphs are further filtered according to the expected module size in image and aspect ratio. 
The adjacency of sub-graphs is used to infer the logical coordinates of PV modules in the columns and rows.


\subsection{CNN-based pre-segmentation}

\begin{figure}[thpb]
\centering
\begin{subfigure}[t]{0.31\textwidth}
  \centering
  \includegraphics[width=1\textwidth]{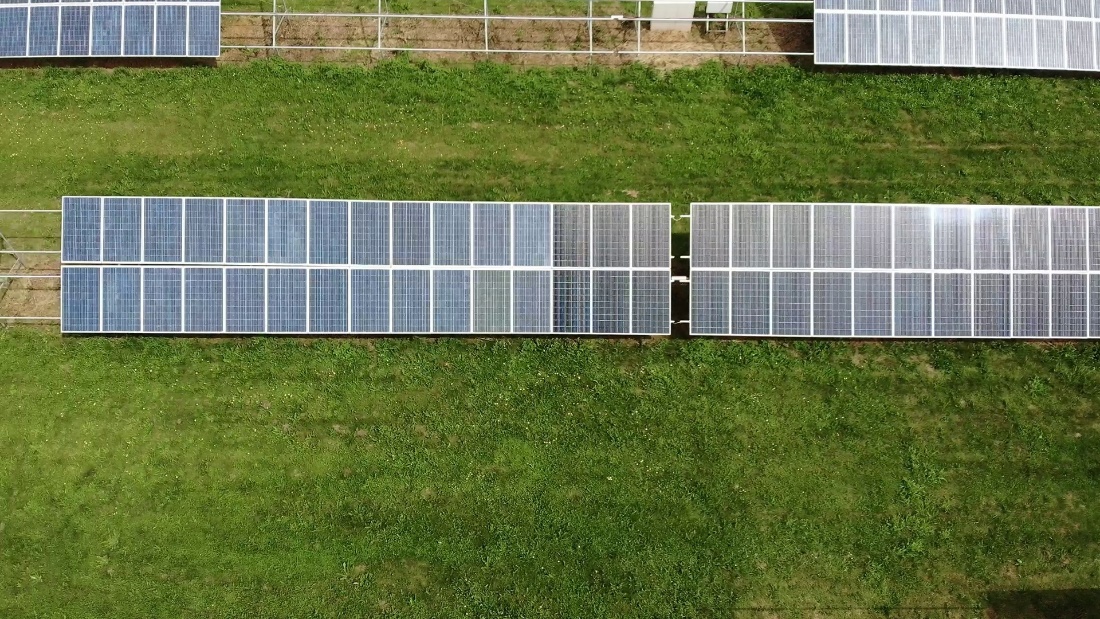}
  \caption{}
  \label{fig:pre_original}
\end{subfigure}\hspace{3mm}%
\begin{subfigure}[t]{0.31\textwidth}
  \centering
  \includegraphics[width=1\textwidth]{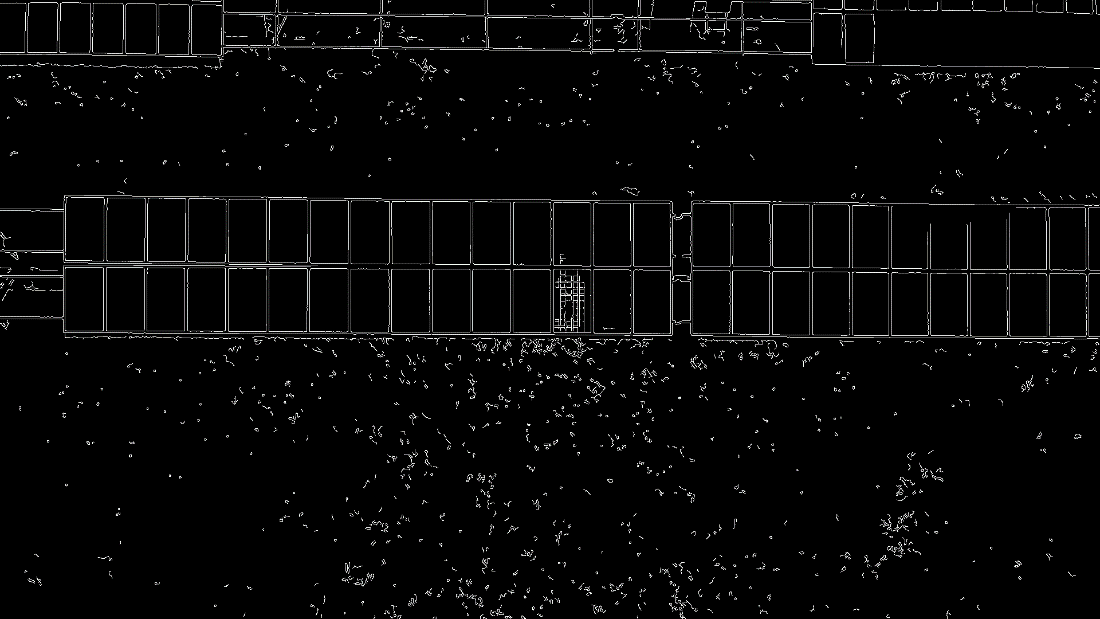}
  \caption{}
  \label{fig:pre_edges_full}
\end{subfigure}\hspace{3mm}%
\begin{subfigure}[t]{0.31\textwidth}
  \centering
  \includegraphics[width=1\textwidth]{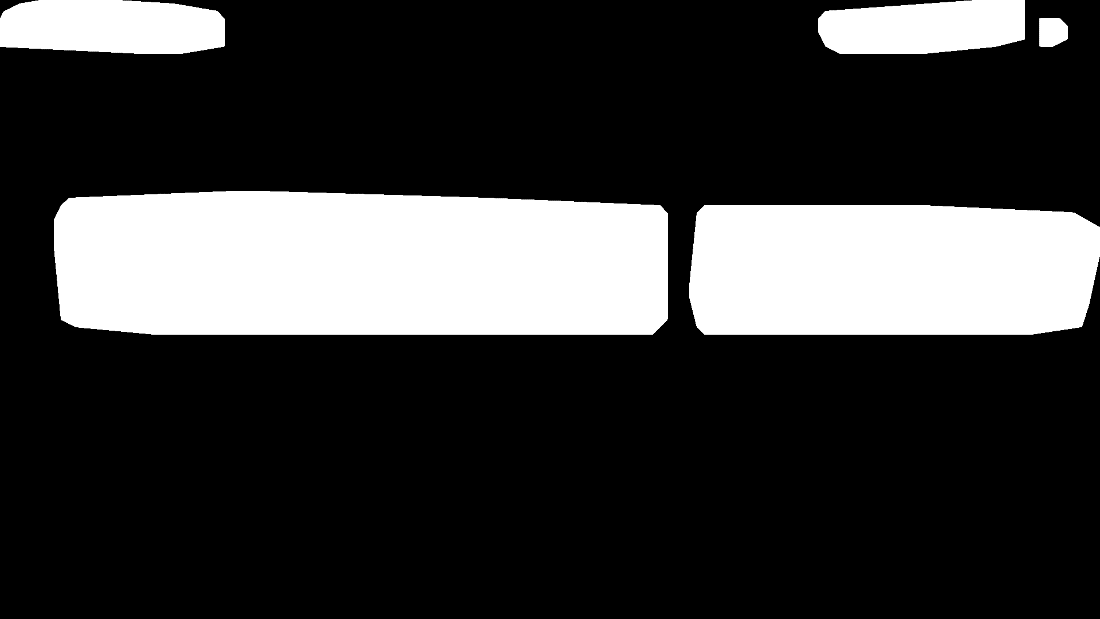}
  \caption{}
  \label{fig:presegmentation_unet}
\end{subfigure}

\vskip\floatsep%
\vspace{-0.1cm}

\begin{subfigure}[t]{0.31\textwidth}
  \centering
  \includegraphics[width=1\textwidth]{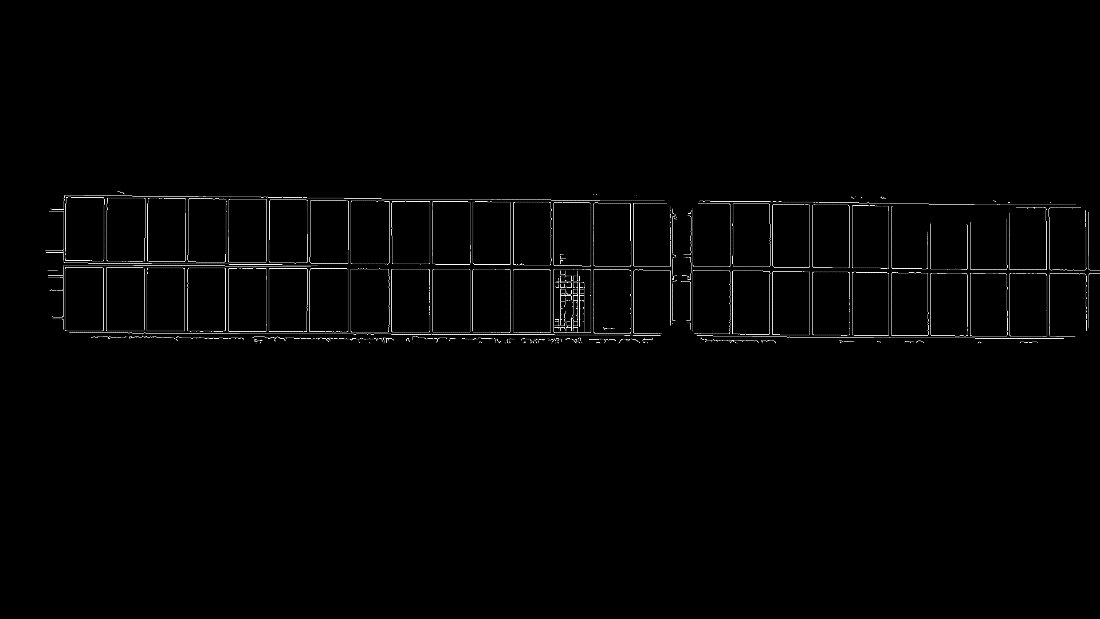}
  \caption{}
  \label{fig:pre_edges_segmented}
\end{subfigure}\hspace{3mm}%
\begin{subfigure}[t]{0.31\textwidth}
  \centering
  \includegraphics[width=1\textwidth]{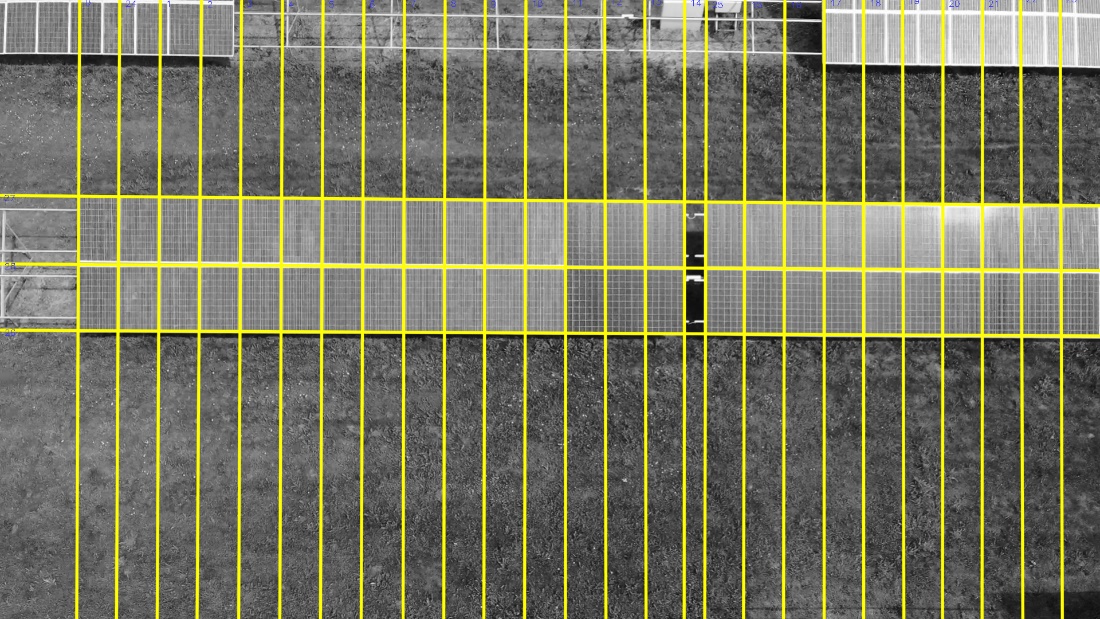}
  \caption{}
  \label{fig:pre_lines_result}
\end{subfigure}

\caption{U-Net based presegmentation. \textbf{(a)} Original image. \textbf{(b)} Edge detection result. \textbf{(c)} U-Net based presegmentation. 
\textbf{(d)} Edge detection result limited to the area of interest. \textbf{(e)} Resulting line detection.} 
\label{fig:presegmentation}
\end{figure}

To mask out neighboring rows present in the image, we introduce an additional presegmention phase. We use a simplified U-Net segmentation model to determine image areas containing PV installations (Figure \ref{fig:presegmentation}). The largest continuous segment is used as the starting point of the mask. We add to the mask all the segments that lie near the middle axis of the largest segment. We discard all detected edges outside of the mask. The CNN-based presegmention phase is optional; however, it can improve the robustness of the proposed edge-based method.

\section{Detailed UAV Localization Results}
\label{apdx_flights}

\subsection{H-Alt Model}
\label{apdx_flights_halt}

\begin{figure}[thpb]
\centering
\begin{subfigure}[t]{.41\textwidth}
  \centering
  \includegraphics[width=1\textwidth]{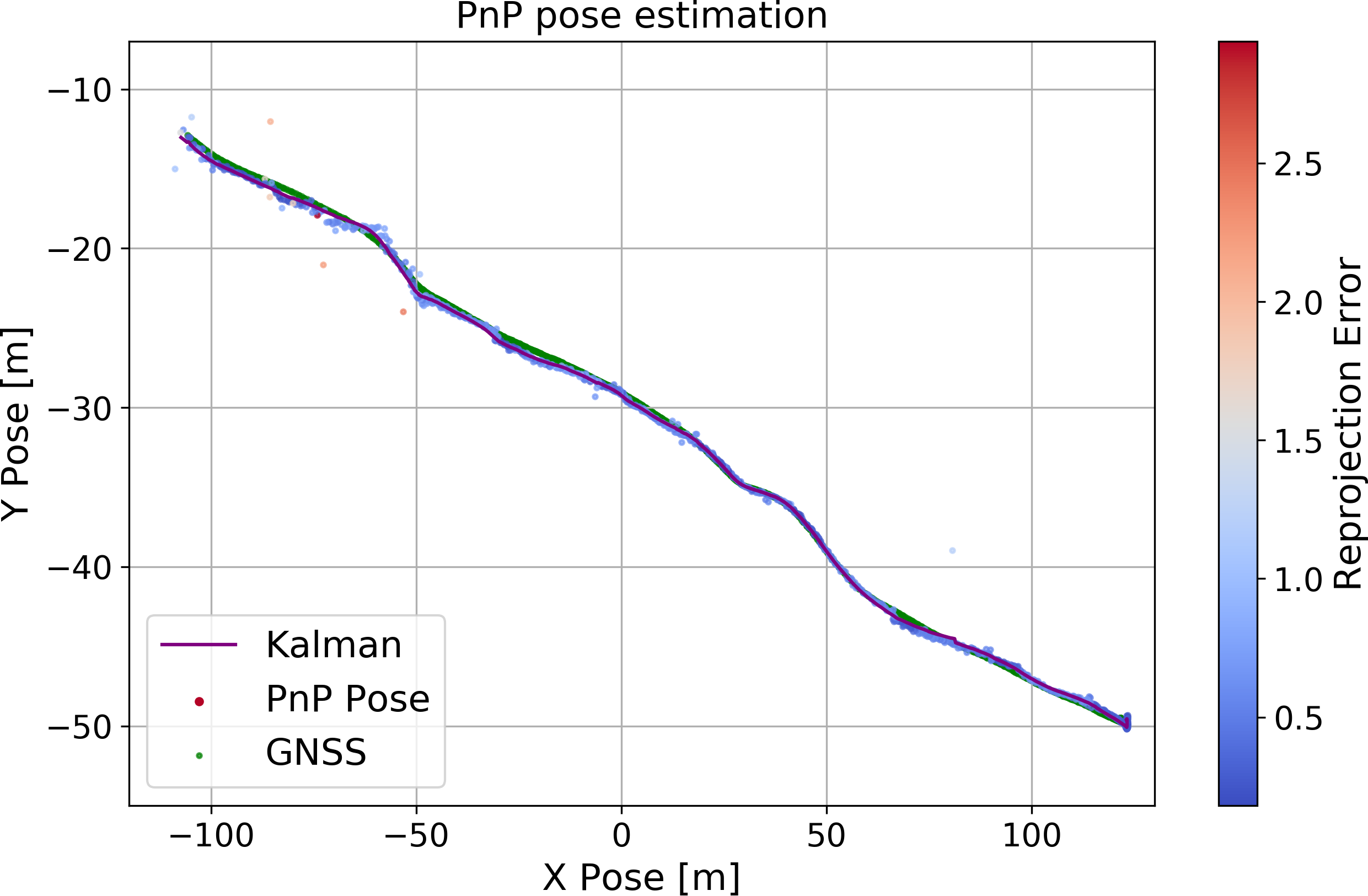}
  \caption{}
  \label{fig:PPAedgeY}
\end{subfigure}\hspace{3mm}%
\begin{subfigure}[t]{.41\textwidth}
  \centering
  \includegraphics[width=1\textwidth]{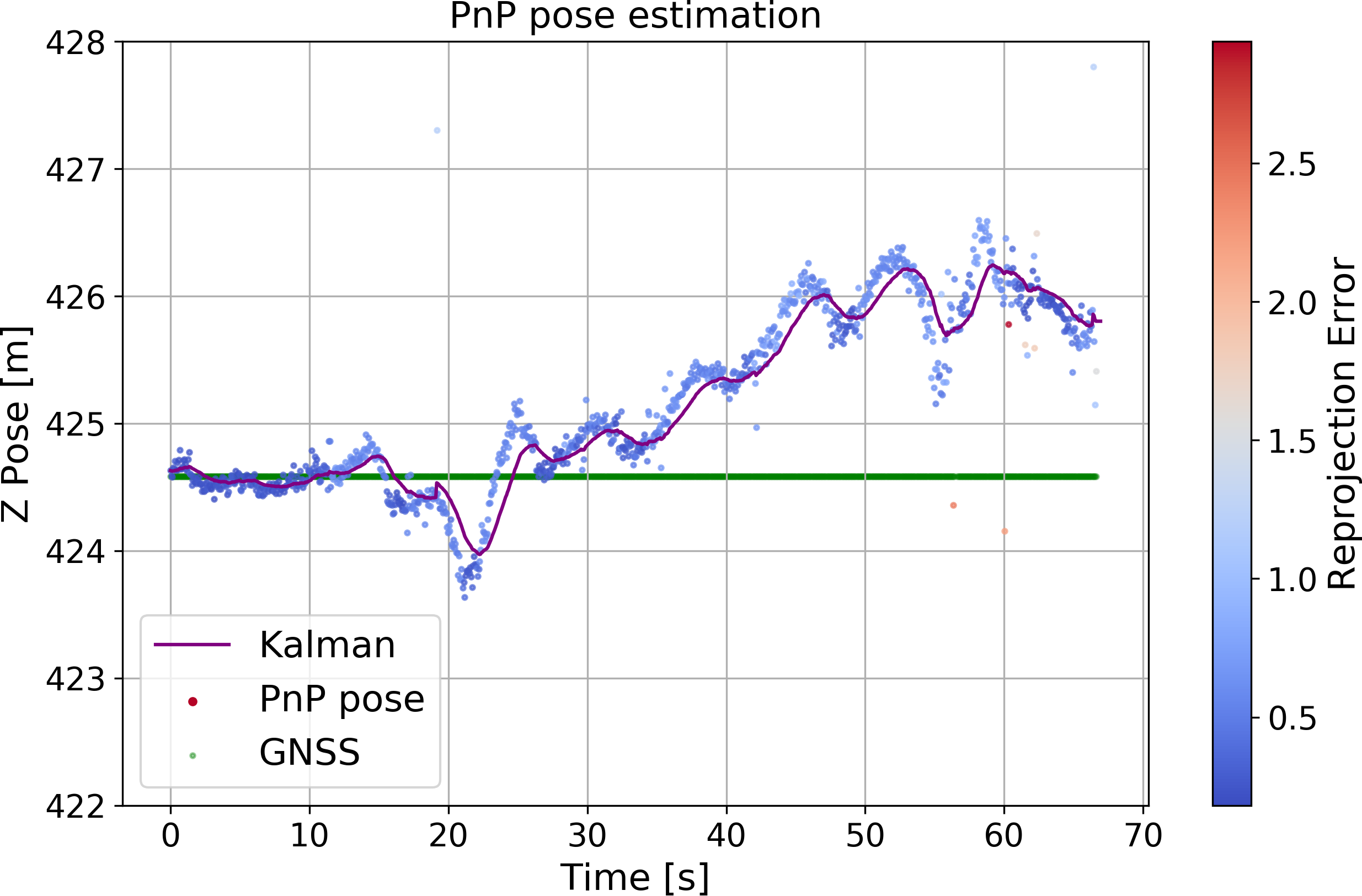}
  \caption{}
  \label{fig:PPAedgeZ}
\end{subfigure}


\begin{subfigure}[t]{.41\textwidth}
  \centering
  \includegraphics[width=1\textwidth]{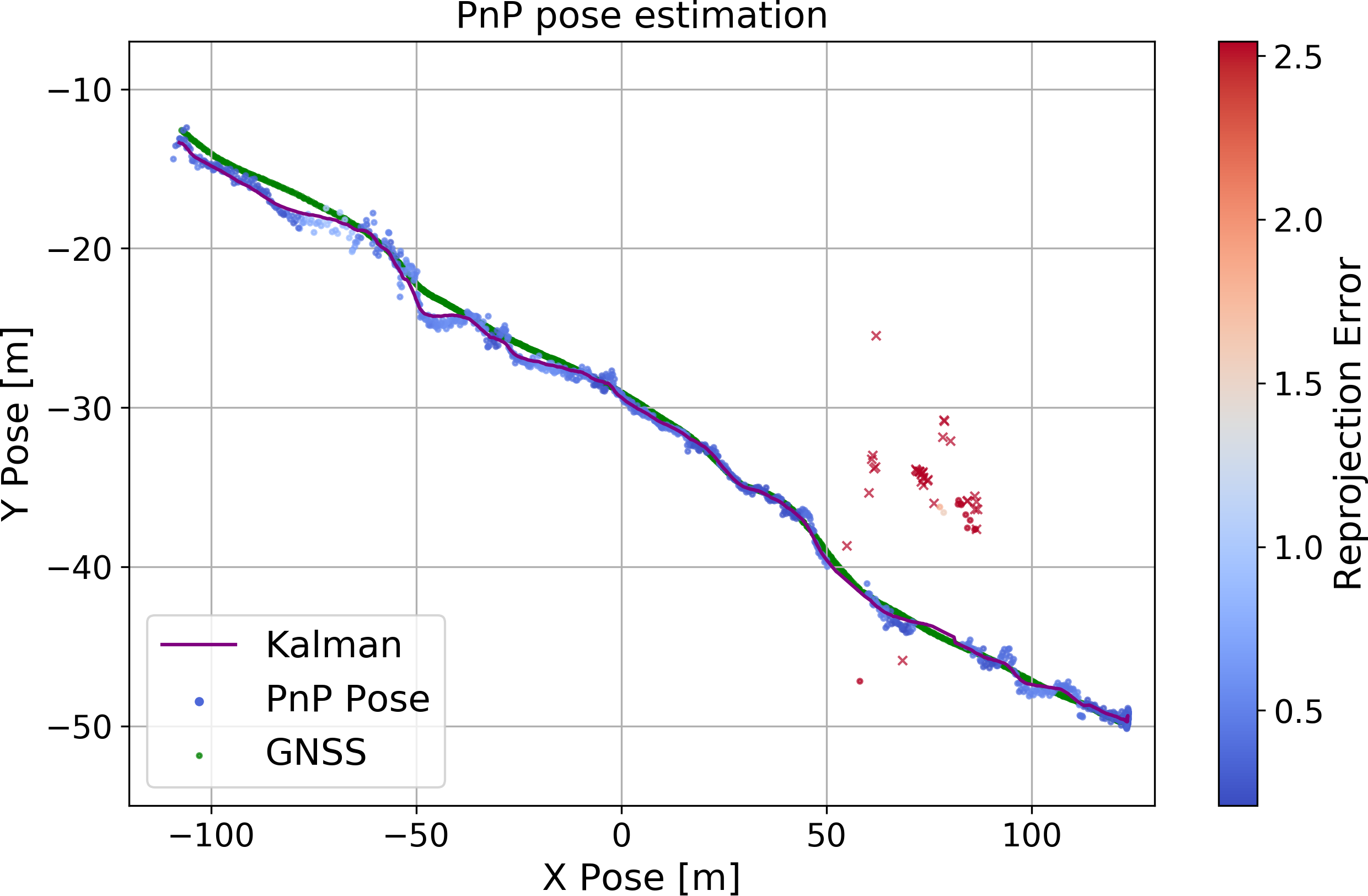}
  \caption{}
  \label{fig:PPAunetX}
\end{subfigure}\hspace{3mm}%
\begin{subfigure}[t]{.41\textwidth}
  \centering
  \includegraphics[width=1\textwidth]{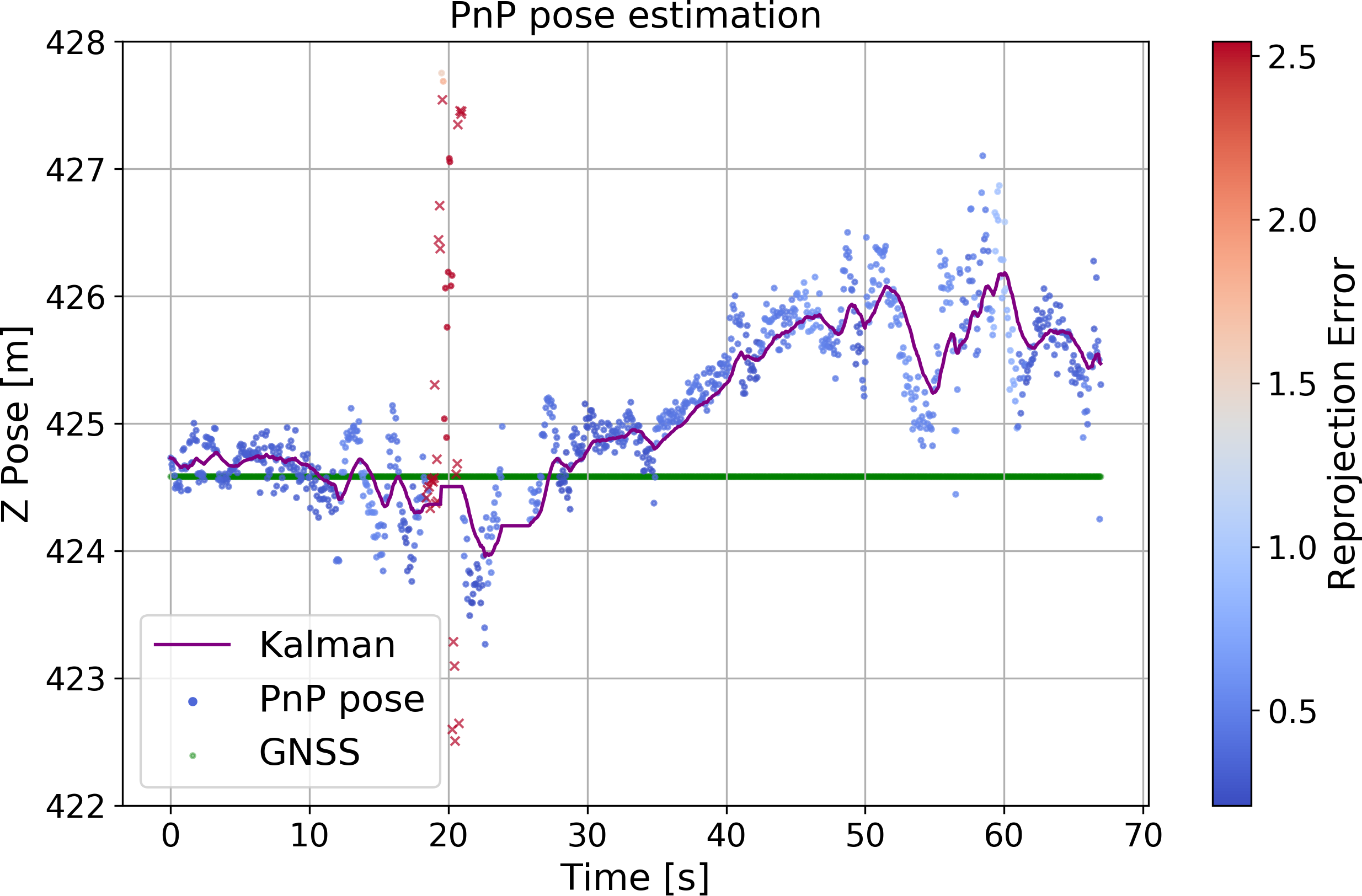}
  \caption{}
  \label{fig:PPAunetZ}
\end{subfigure}


\begin{subfigure}[t]{.41\textwidth}
  \centering
  \includegraphics[width=1\textwidth]{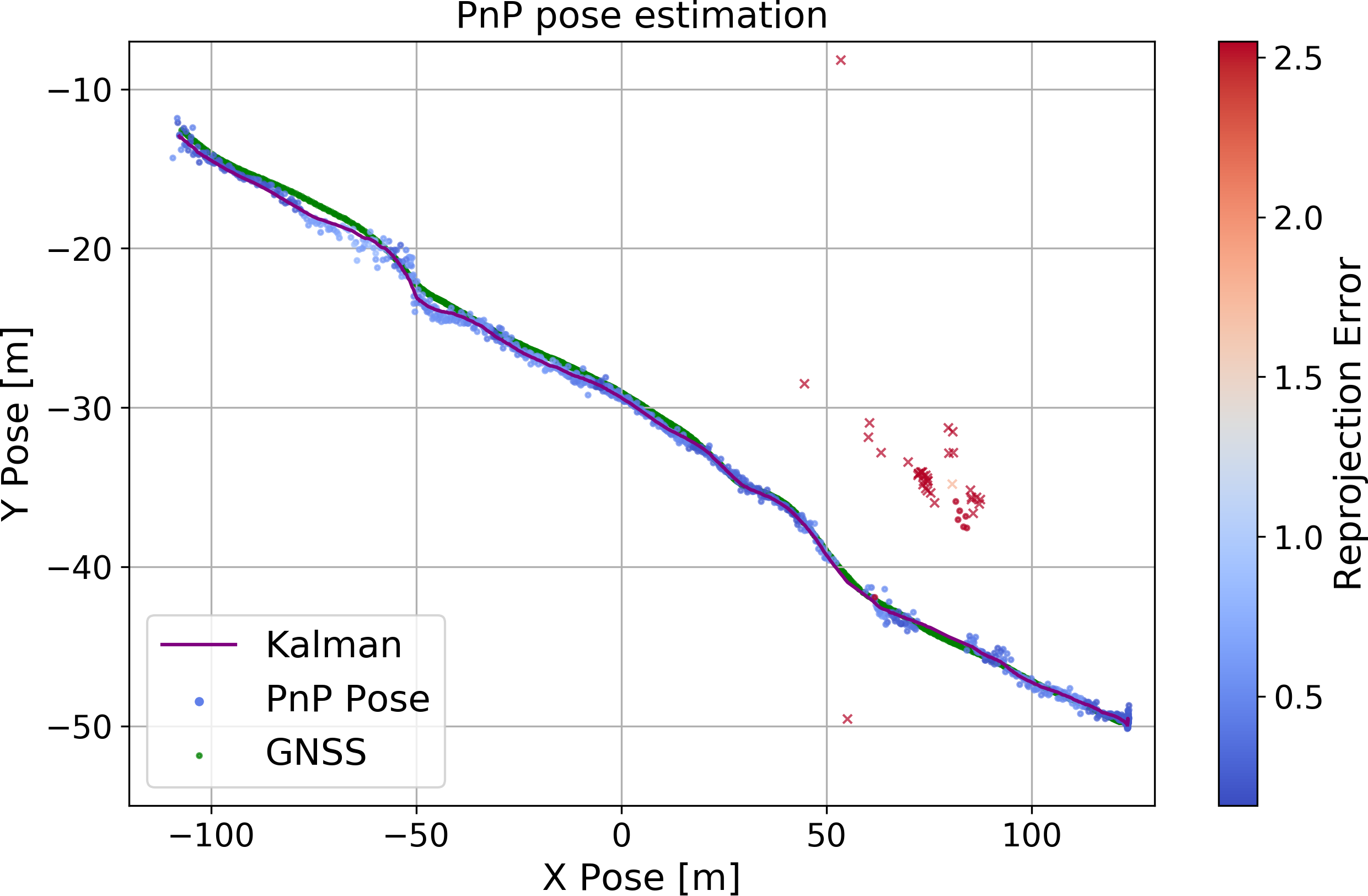}
  \caption{}
  \label{fig:PPAyoloX}
\end{subfigure}\hspace{3mm}%
\begin{subfigure}[t]{.41\textwidth}
  \centering
  \includegraphics[width=1\textwidth]{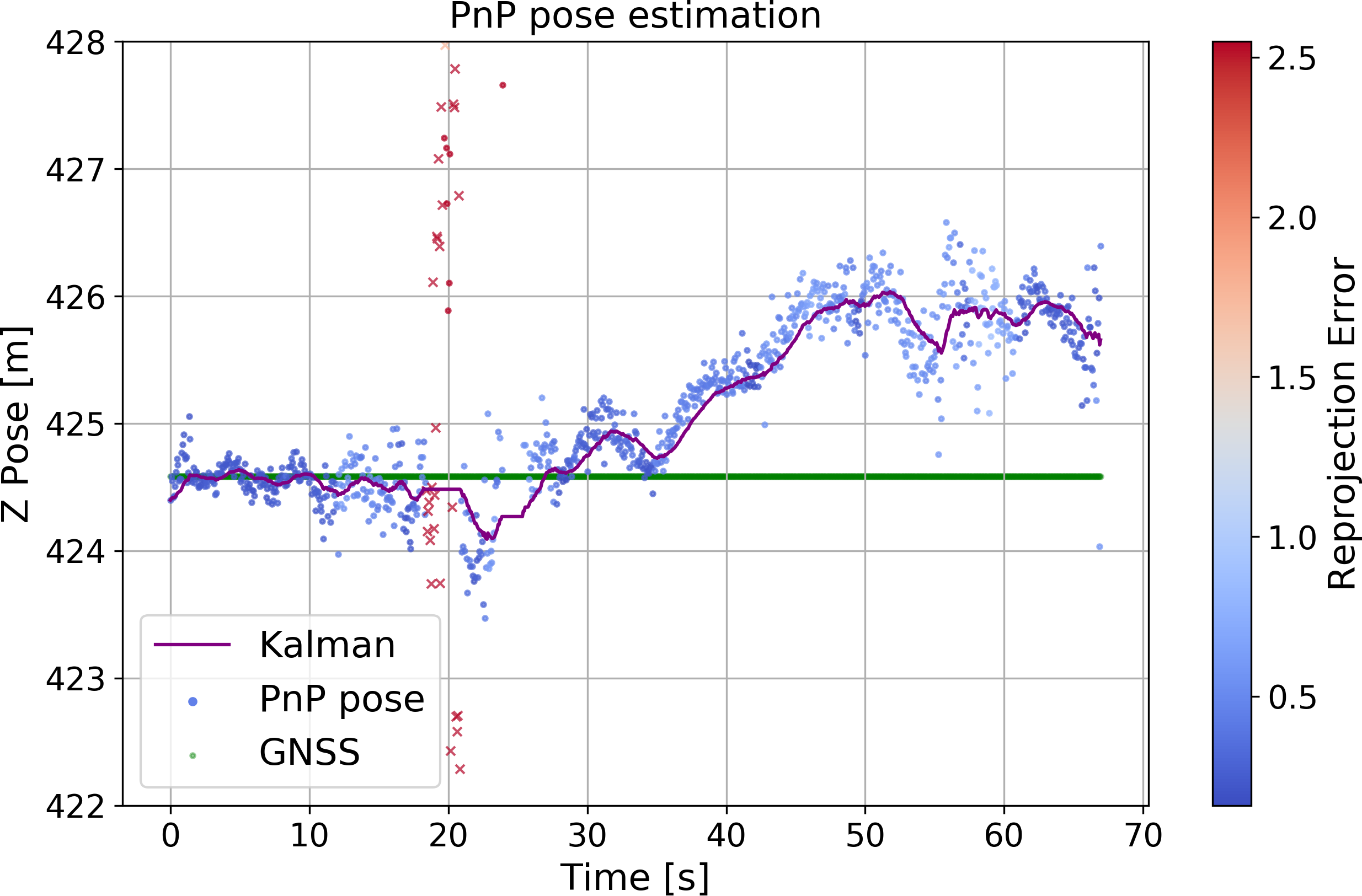}
  \caption{}
  \label{fig:PPAyoloZ}
\end{subfigure}

\caption{The PnP output for PPA2 using the \textit{H-Alt} model. The reprojection error is illustrated using a color map. Deviations exceeding the distance threshold $th_d$ are indicated by crosses. Figures \textbf{(a)} and \textbf{(b)} present results of the edge-based method. Figures \textbf{(c)} and \textbf{(d)} present results for U-Net. Figures \textbf{(e)} and \textbf{(f)} present results for YOLO.}
\label{fig:axes_repr_HAlt_PPA}
\end{figure}

\begin{figure}[thpb]
\centering
\begin{subfigure}[t]{.41\textwidth}
  \centering
  \includegraphics[width=1\textwidth]{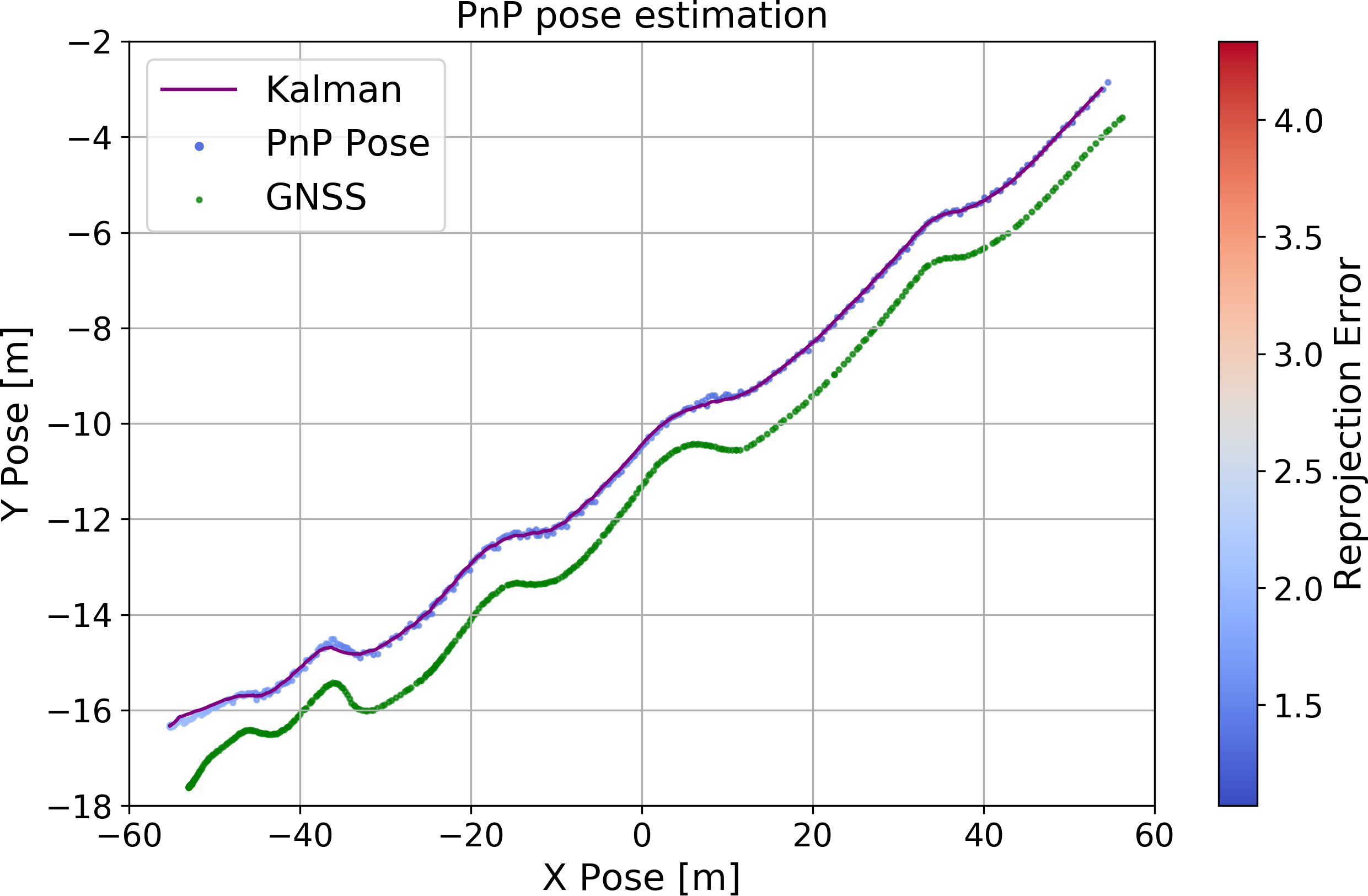}
  \caption{}
  \label{fig:PPBedgeY}
\end{subfigure}\hspace{3mm}%
\begin{subfigure}[t]{.41\textwidth}
  \centering
  \includegraphics[width=1\textwidth]{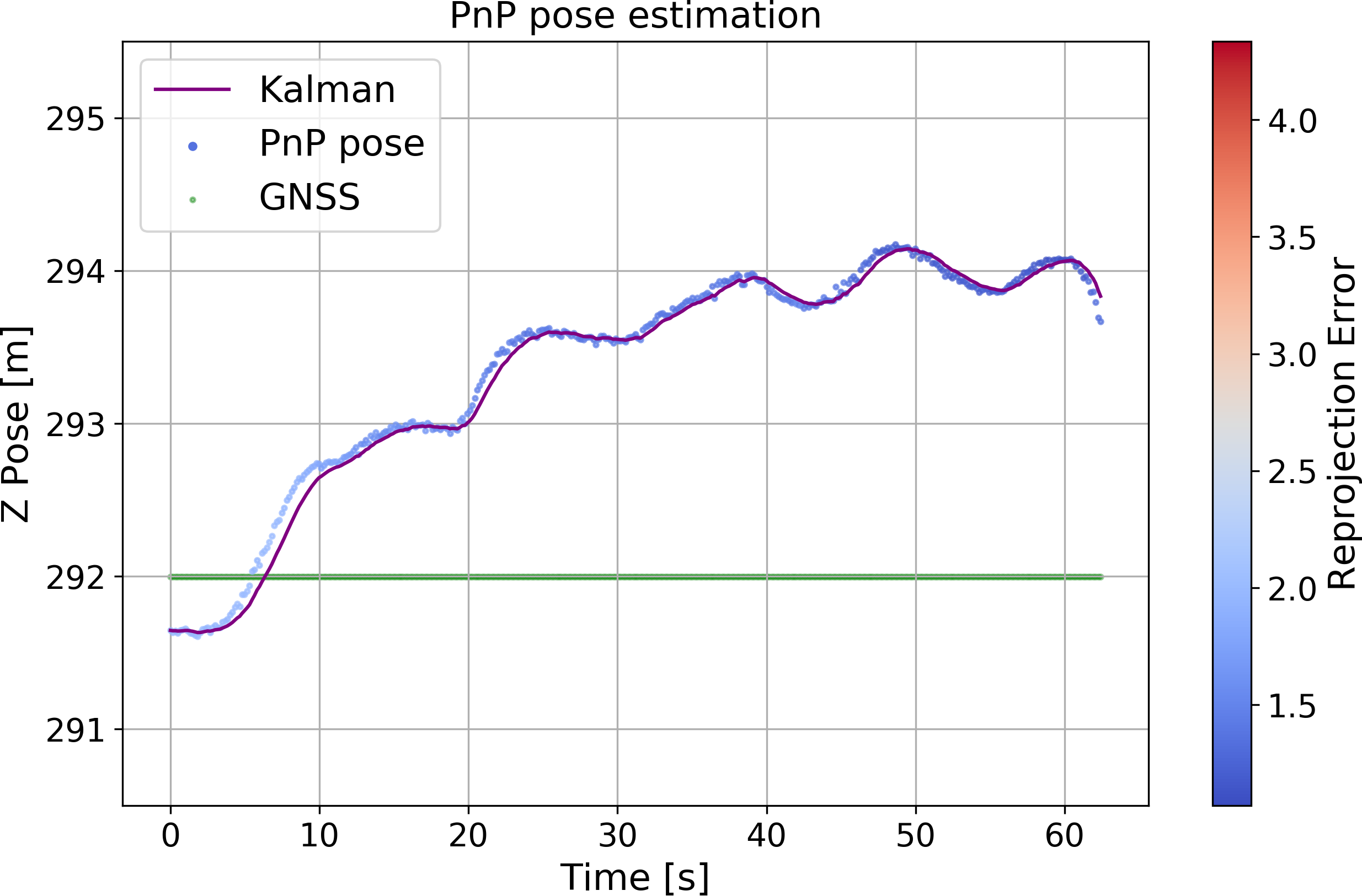}
  \caption{}
  \label{fig:PPBedgeZ}
\end{subfigure}


\begin{subfigure}[t]{.41\textwidth}
  \centering
  \includegraphics[width=1\textwidth]{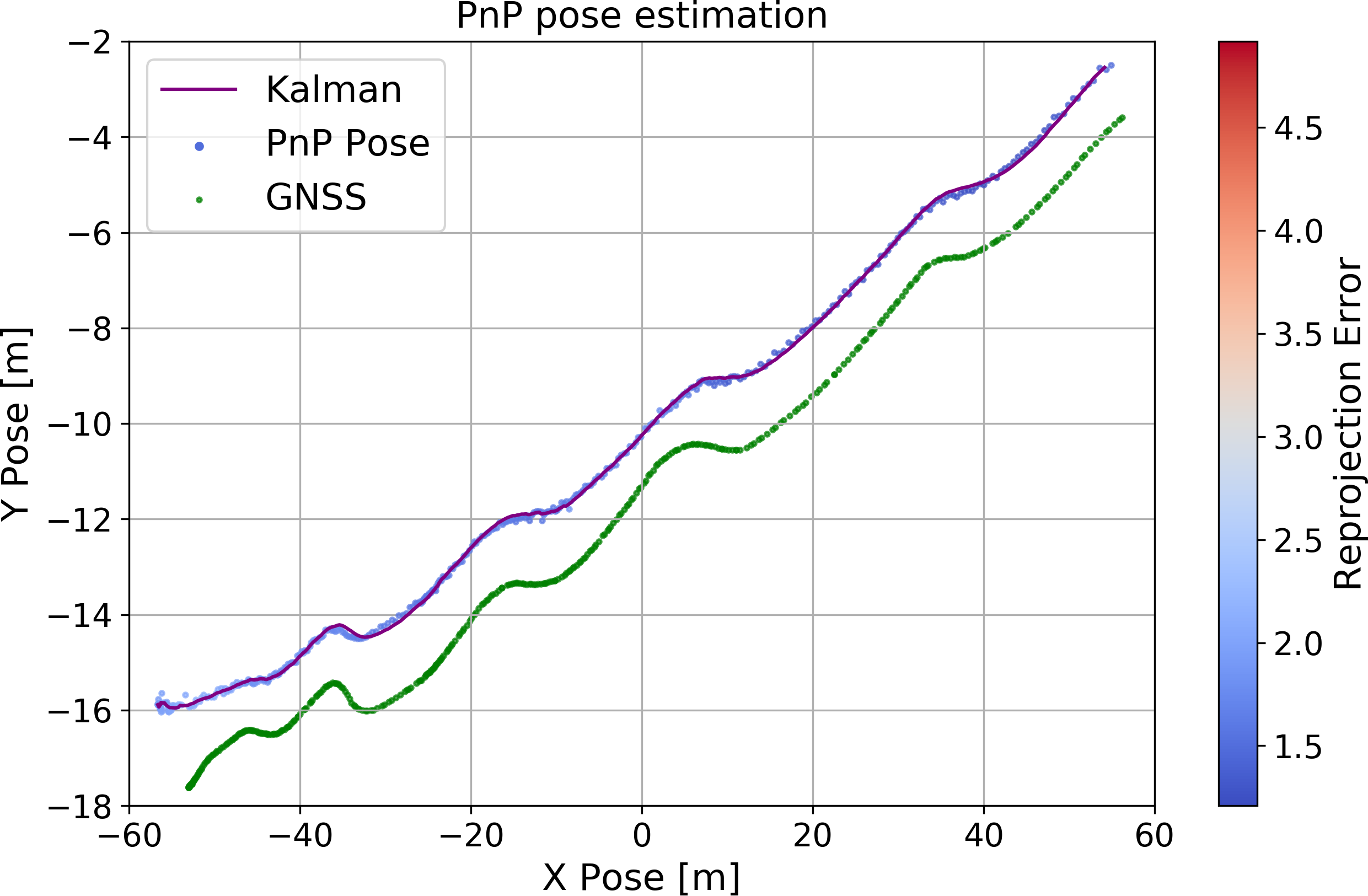}
  \caption{}
  \label{fig:PPBunetX}
\end{subfigure}\hspace{3mm}%
\begin{subfigure}[t]{.41\textwidth}
  \centering
  \includegraphics[width=1\textwidth]{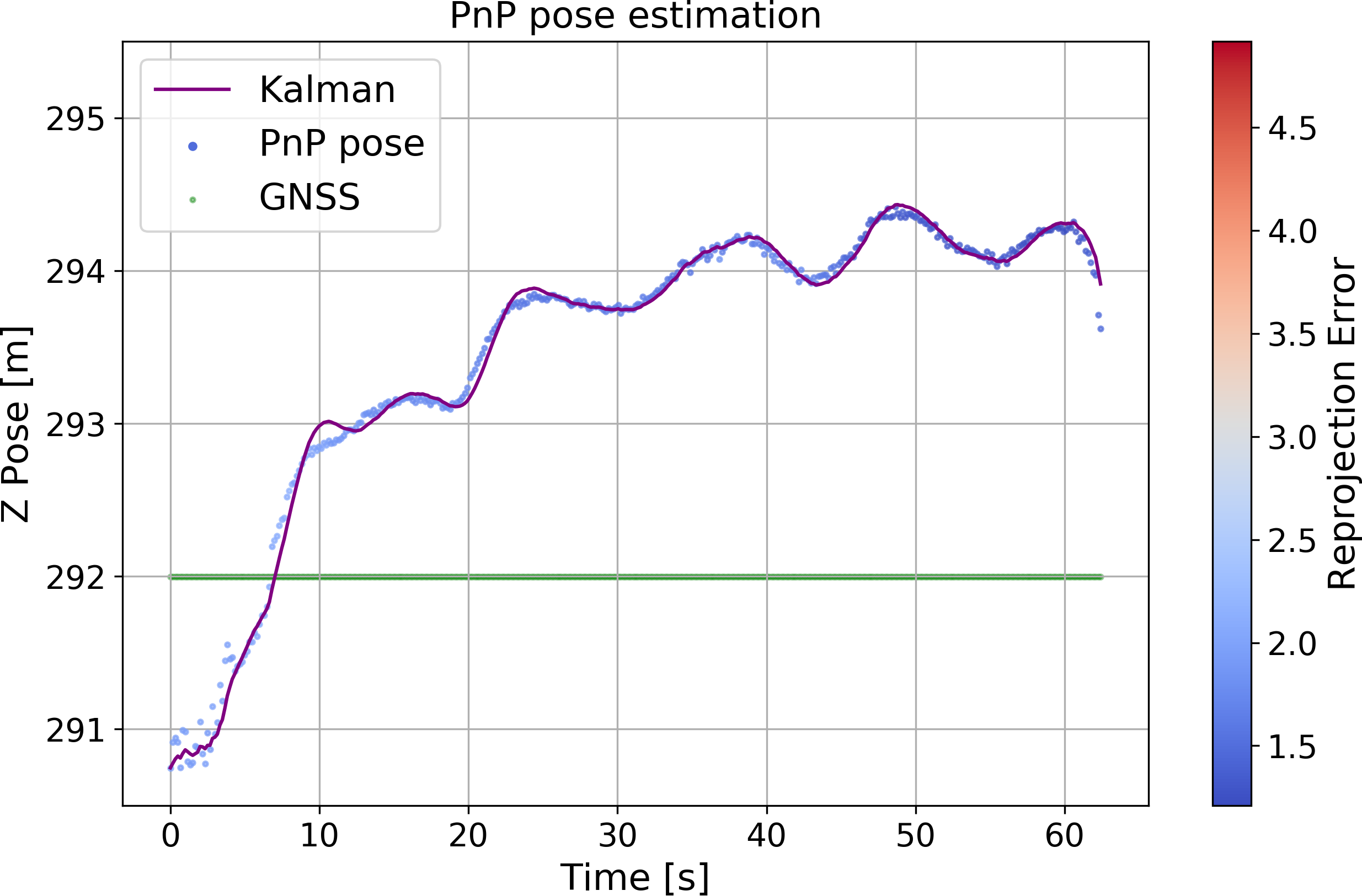}
  \caption{}
  \label{fig:PPBunetZ}
\end{subfigure}


\begin{subfigure}[t]{.41\textwidth}
  \centering
  \includegraphics[width=1\textwidth]{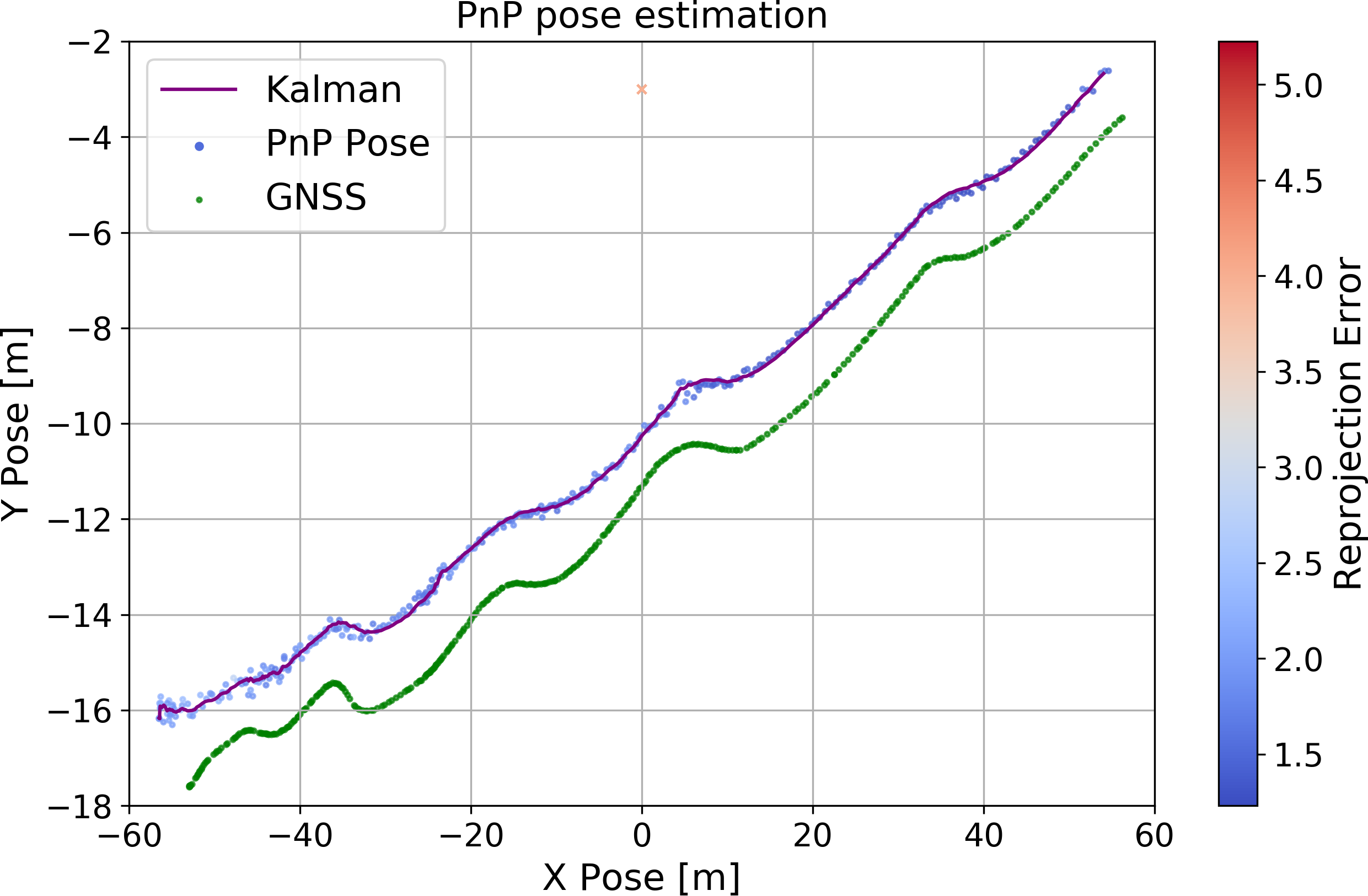}
  \caption{}
  \label{fig:PPByoloX}
\end{subfigure}\hspace{3mm}%
\begin{subfigure}[t]{.41\textwidth}
  \centering
  \includegraphics[width=1\textwidth]{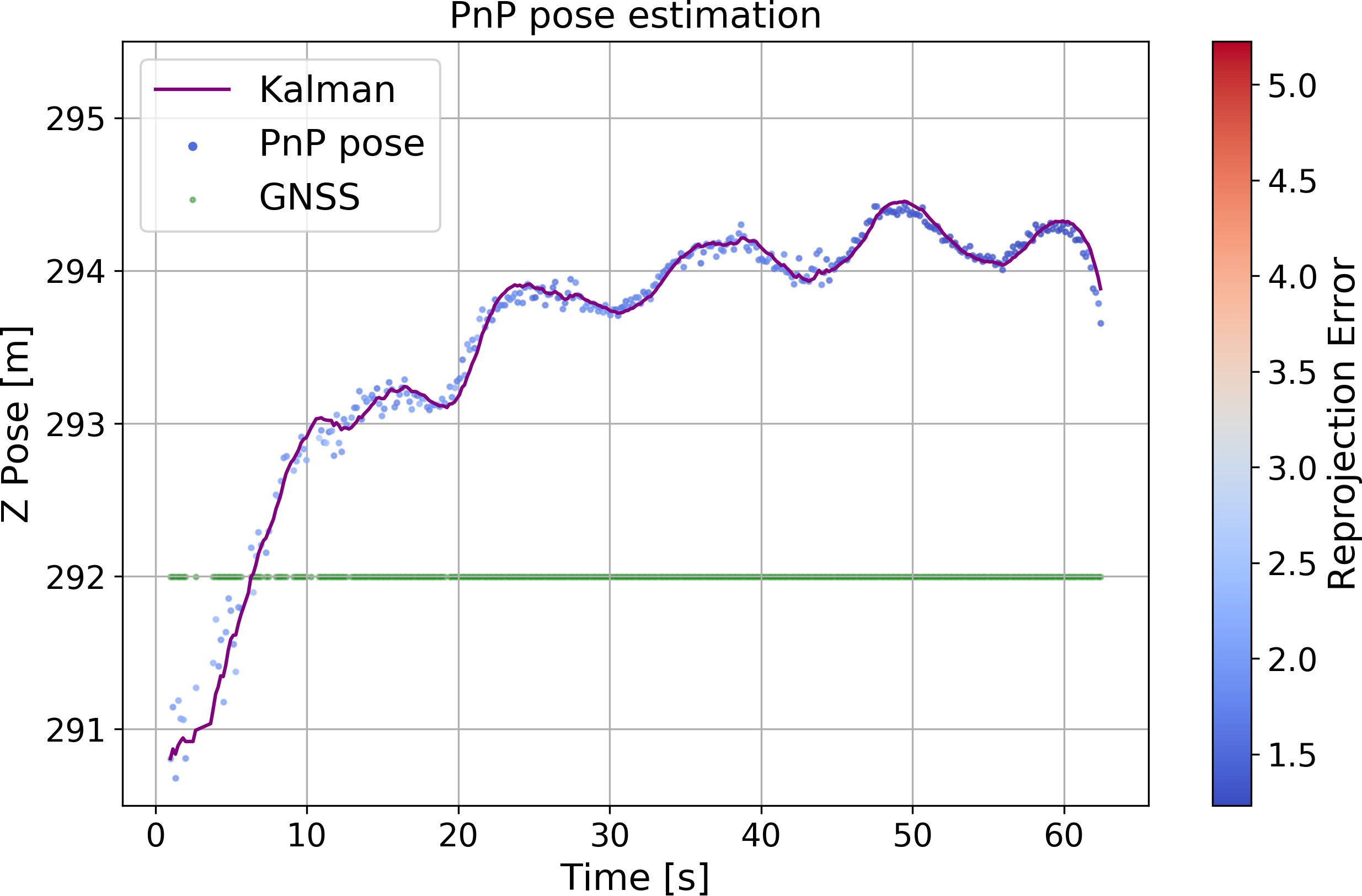}
  \caption{}
  \label{fig:PPByoloZ}
\end{subfigure}

\caption{The PnP output for PPB using the \textit{H-Alt} model. The reprojection error is illustrated using a color map. Deviations exceeding the distance threshold $th_d$ are indicated by crosses. Figures \textbf{(a)} and \textbf{(b)} present results of the edge-based method. Figures \textbf{(c)} and \textbf{(d)} present results for U-Net. Figures \textbf{(e)} and \textbf{(f)} present results for YOLO.}
\label{fig:axes_repr_HAlt_PPB}
\end{figure}


\newpage

\subsection{SfM Model}
\label{apdx_flights_sfm}

\begin{figure}[thpb]
\centering
\begin{subfigure}[t]{.41\textwidth}
  \centering
  \includegraphics[width=1\textwidth]{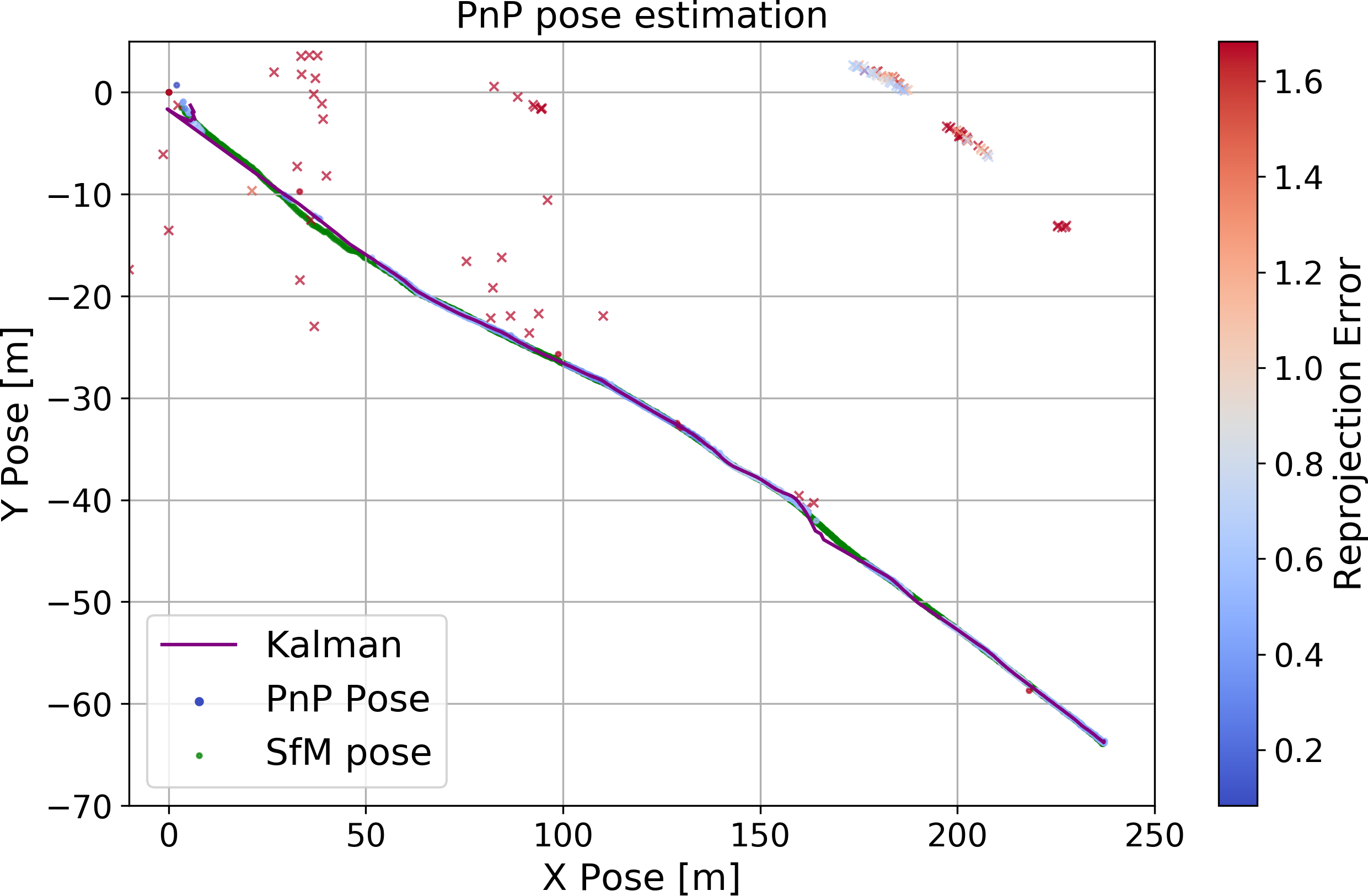}
  \caption{}
  \label{fig:SedgeXY_PPA}
\end{subfigure}\hspace{3mm}%
\begin{subfigure}[t]{.41\textwidth}
  \centering
  \includegraphics[width=1\textwidth]{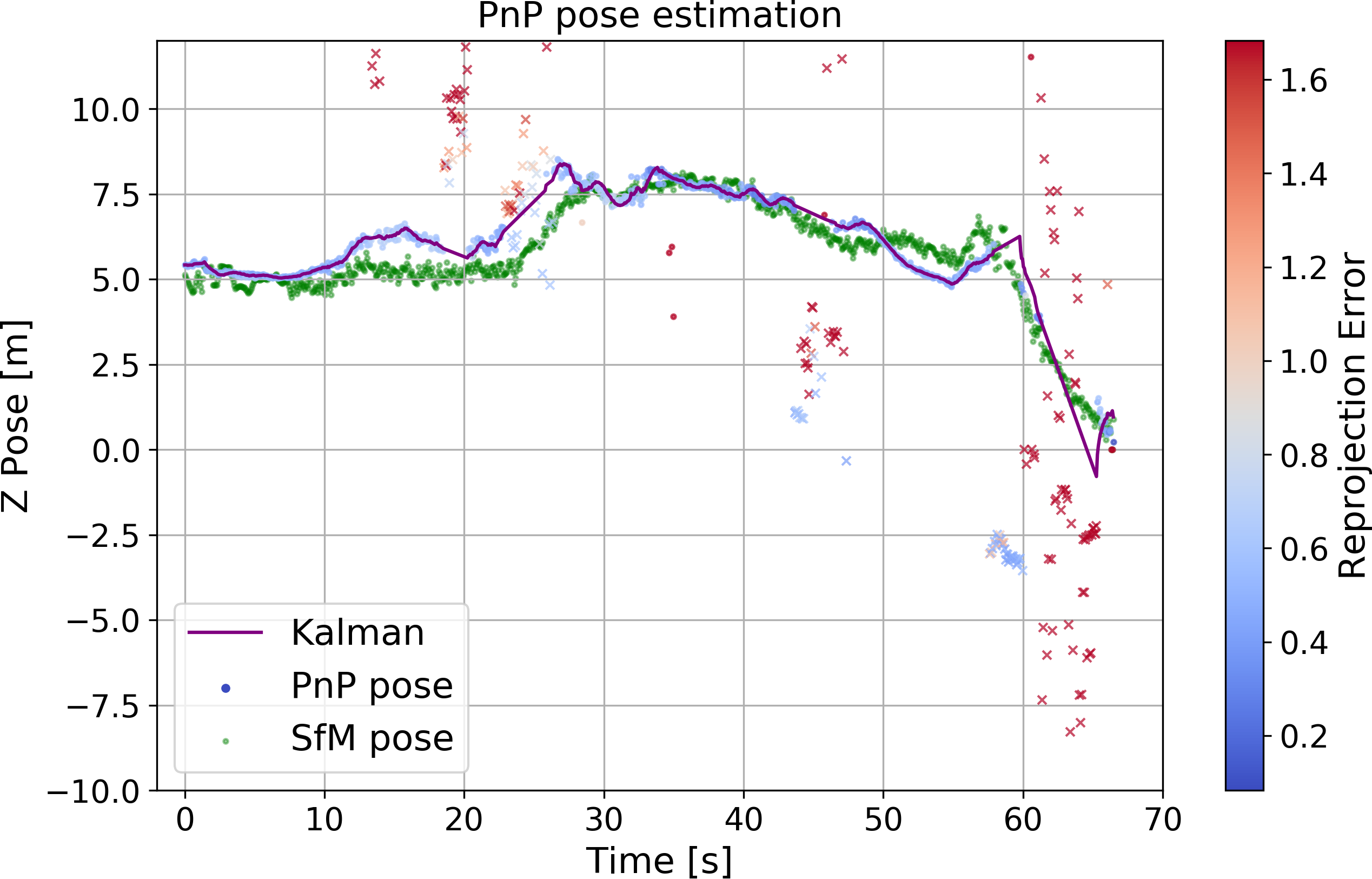}
  \caption{}
  \label{fig:SedgeZ_PPA}
\end{subfigure}


\begin{subfigure}[t]{.41\textwidth}
  \centering
  \includegraphics[width=1\textwidth]{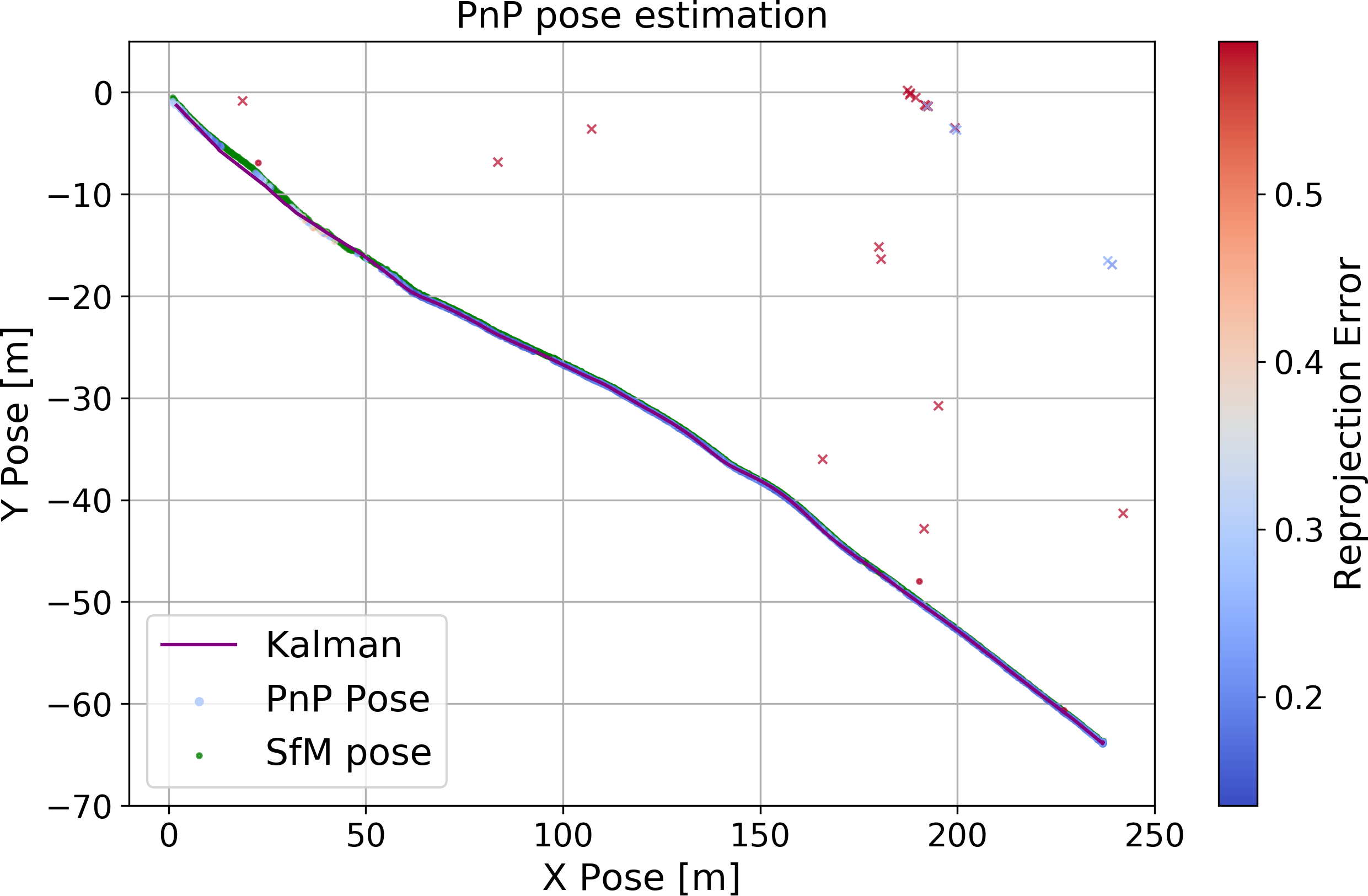}
  \caption{}
  \label{fig:SunetXY_PPA}
\end{subfigure}\hspace{3mm}%
\begin{subfigure}[t]{.41\textwidth}
  \centering
  \includegraphics[width=1\textwidth]{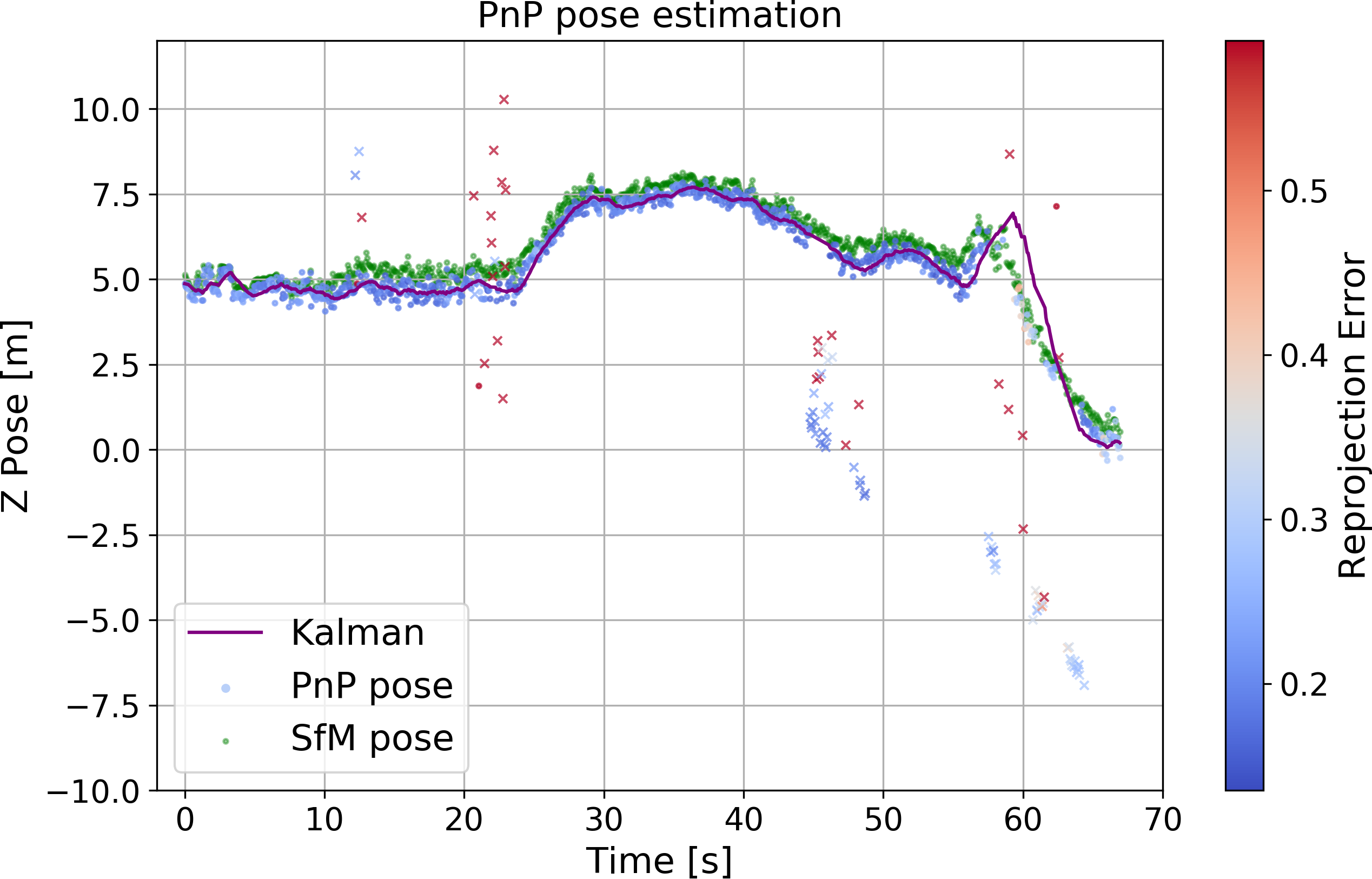}
  \caption{}
  \label{fig:SunetZ_PPA}
\end{subfigure}


\begin{subfigure}[t]{.41\textwidth}
  \centering
  \includegraphics[width=1\textwidth]{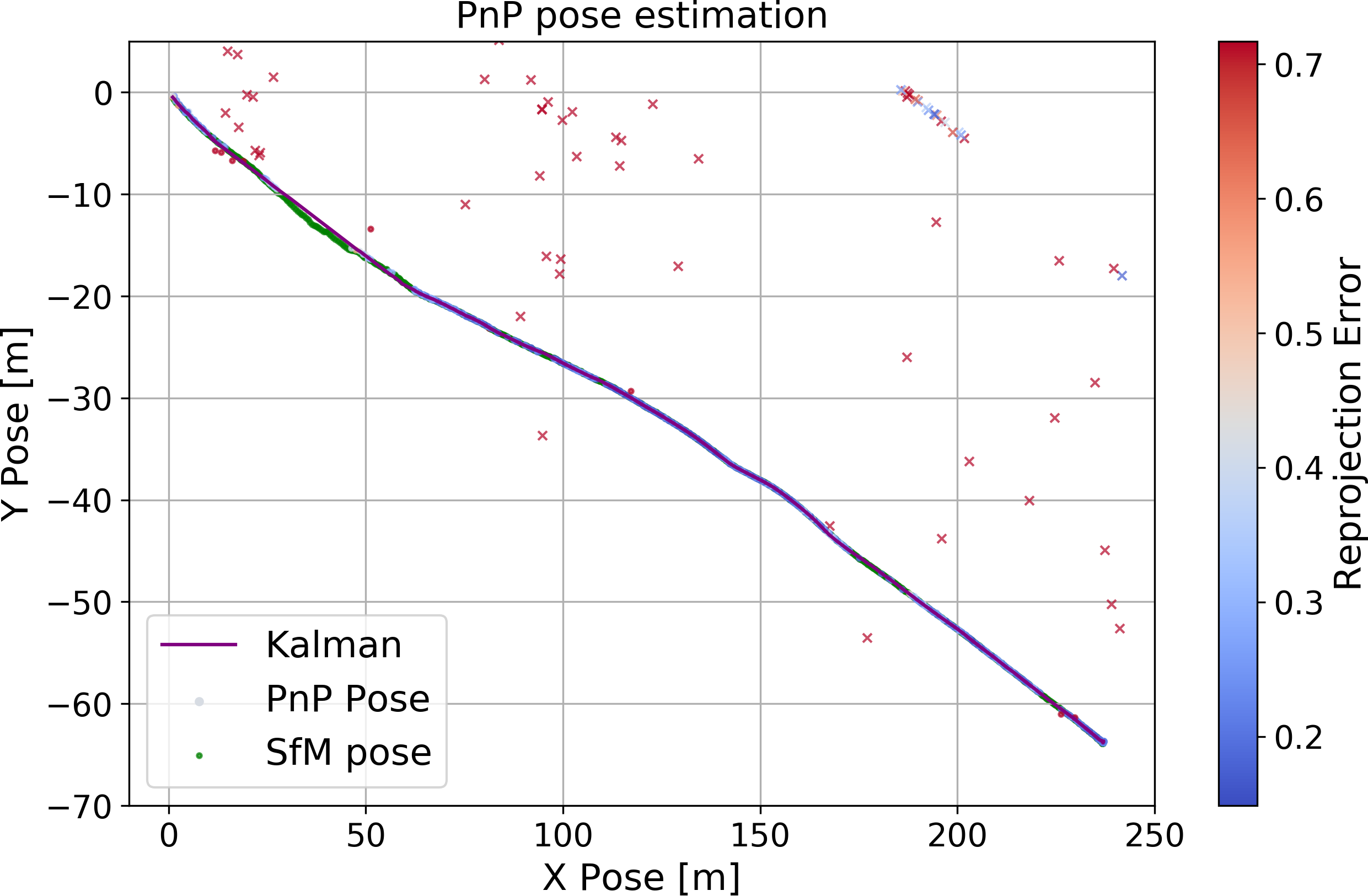}
  \caption{}
  \label{fig:SyoloXY_PPA}
\end{subfigure}\hspace{3mm}%
\begin{subfigure}[t]{.41\textwidth}
  \centering
  \includegraphics[width=1\textwidth]{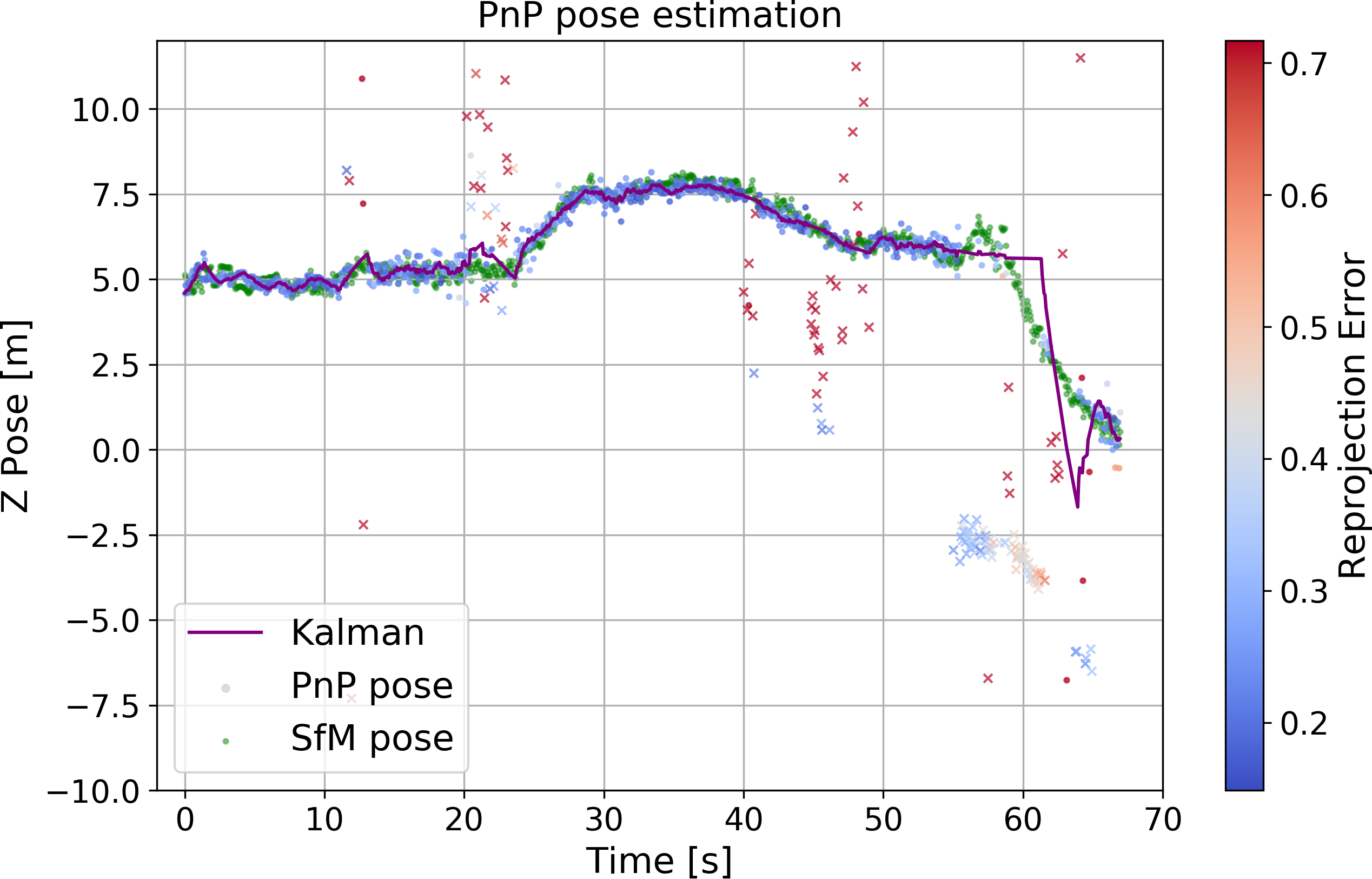}
  \caption{}
  \label{fig:SyoloZ_PPA}
\end{subfigure}


\caption{PPA2, \textit{SfM} model. The reprojection error is  illustrated using a color map. Figures \textbf{(a)} and \textbf{(b)} present results of the edge-based method. Figures \textbf{(c)} and \textbf{(d)} present results for U-Net. Figures \textbf{(e)} and \textbf{(f)} present results for YOLO.}
\label{fig:Saxes_repr_SfM_PPA}
\end{figure}

\begin{figure}[thpb]
\centering
\begin{subfigure}[t]{.41\textwidth}
  \centering
  \includegraphics[width=1\textwidth]{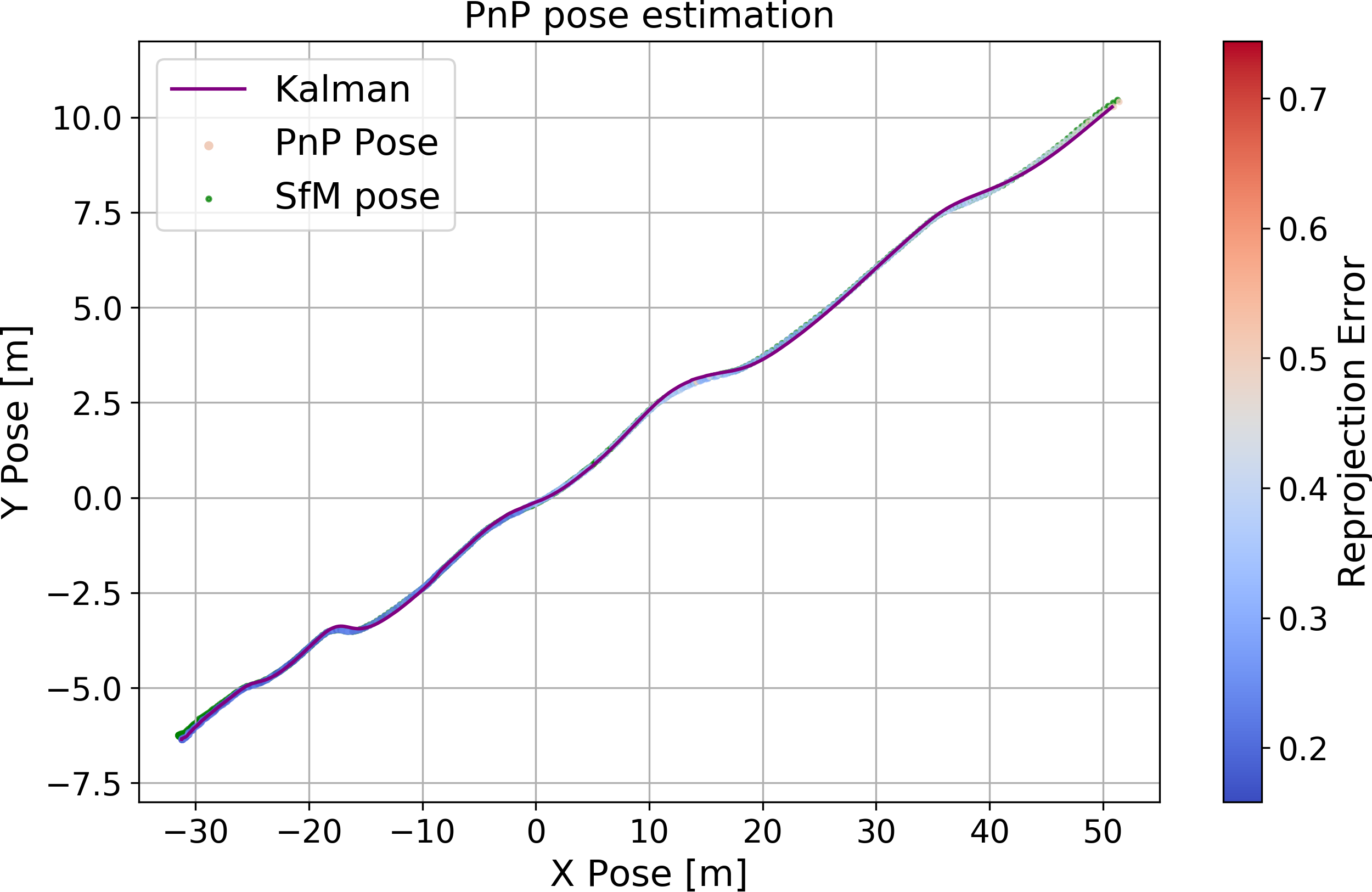}
  \caption{}
  \label{fig:SedgeXY_PPB}
\end{subfigure}\hspace{3mm}%
\begin{subfigure}[t]{.41\textwidth}
  \centering
  \includegraphics[width=1\textwidth]{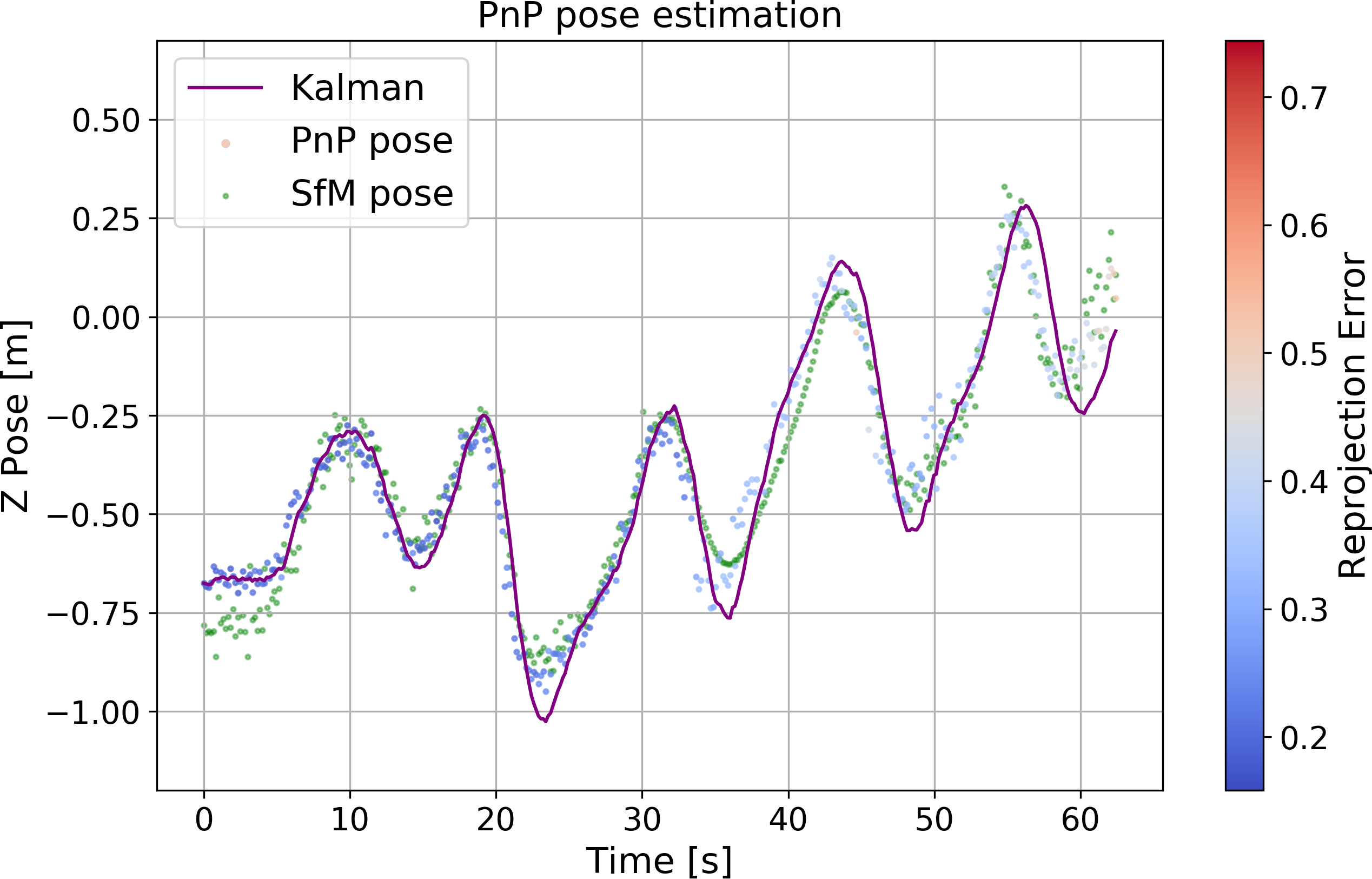}
  \caption{}
  \label{fig:SedgeZ_PPB}
\end{subfigure}

\begin{subfigure}[t]{.41\textwidth}
  \centering
  \includegraphics[width=1\textwidth]{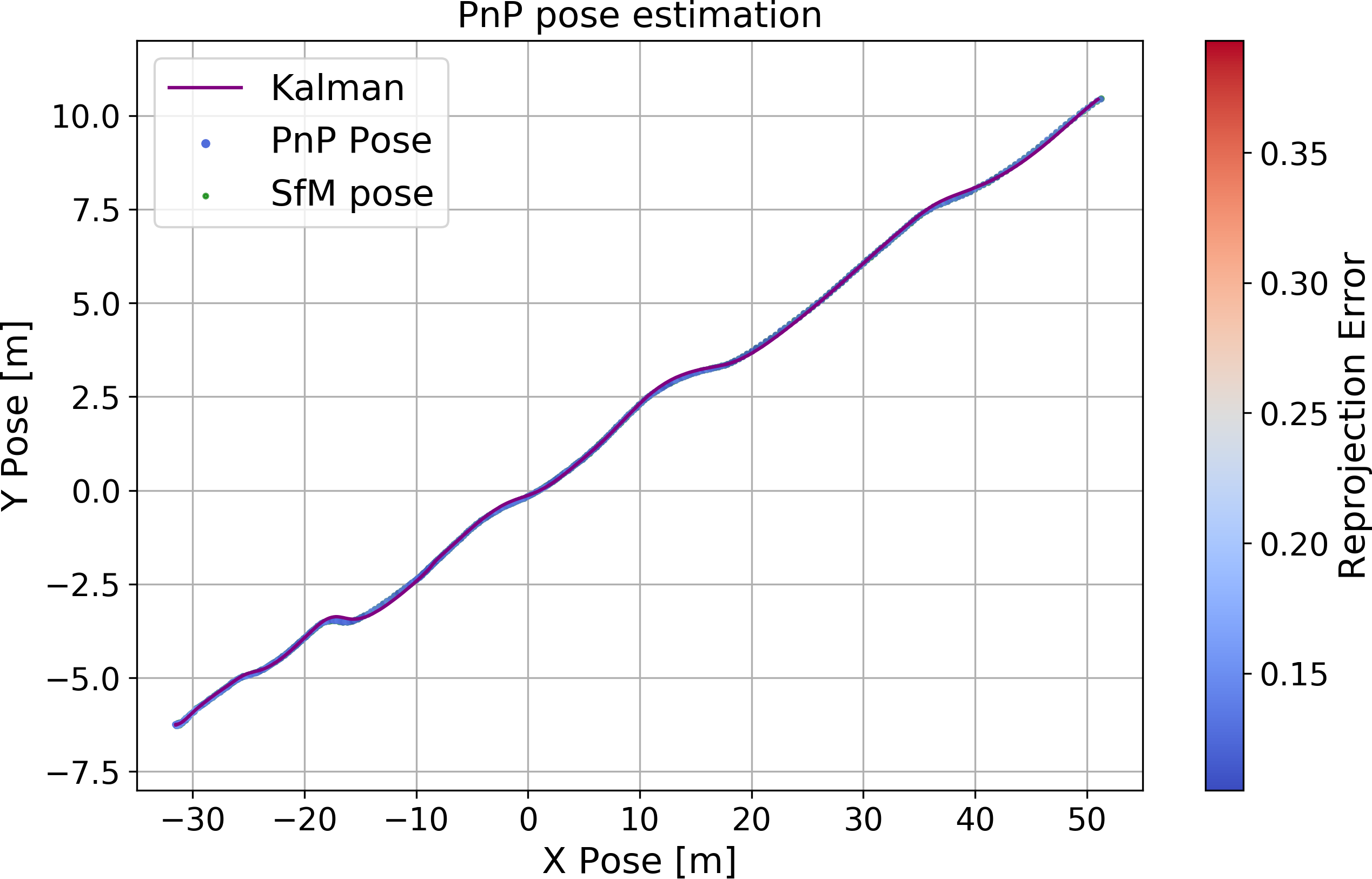}
  \caption{}
  \label{fig:SunetXY_PPB}
\end{subfigure}\hspace{3mm}%
\begin{subfigure}[t]{.41\textwidth}
  \centering
  \includegraphics[width=1\textwidth]{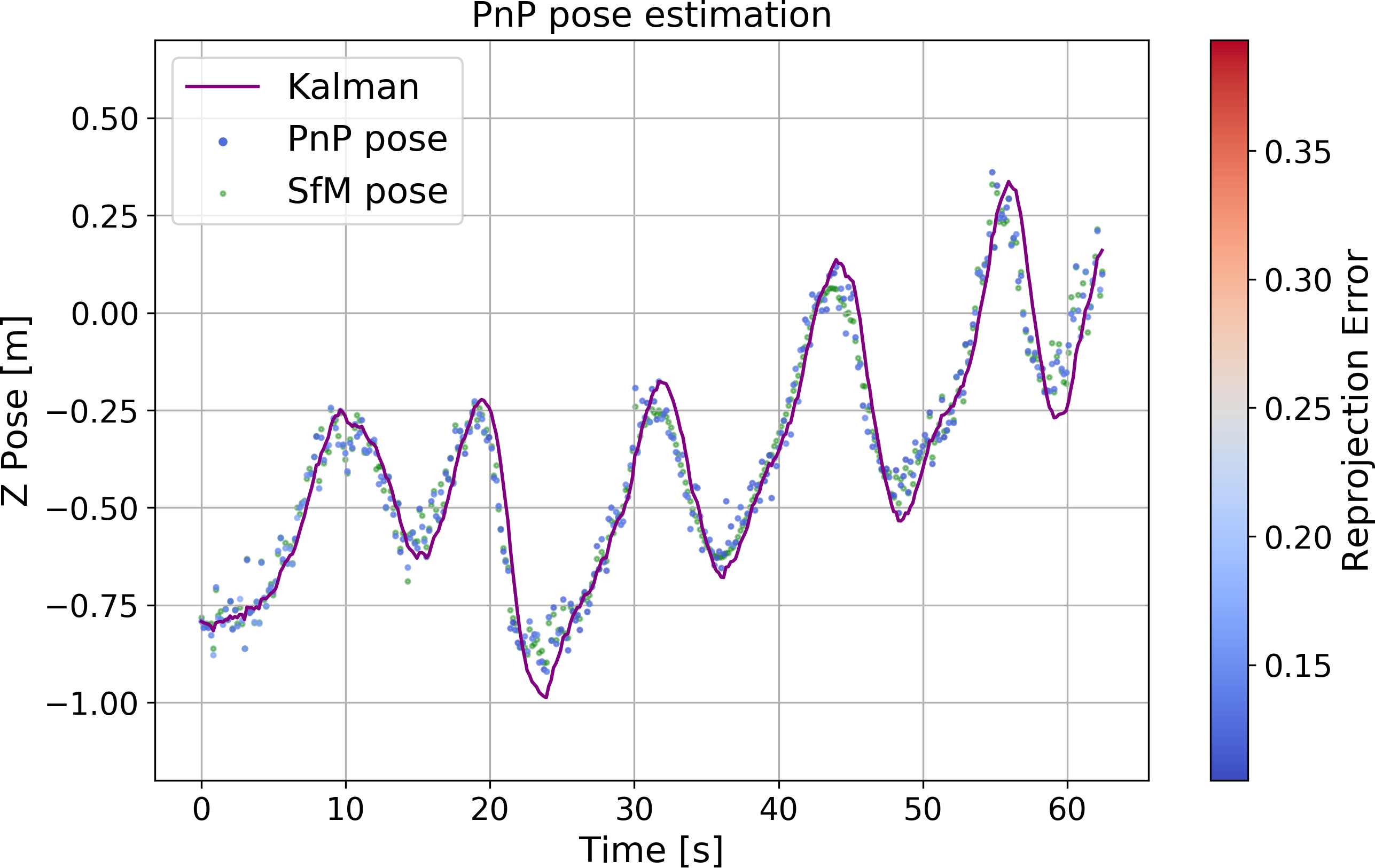}
  \caption{}
  \label{fig:SunetZ_PPB}
\end{subfigure}

\begin{subfigure}[t]{.41\textwidth}
  \centering
  \includegraphics[width=1\textwidth]{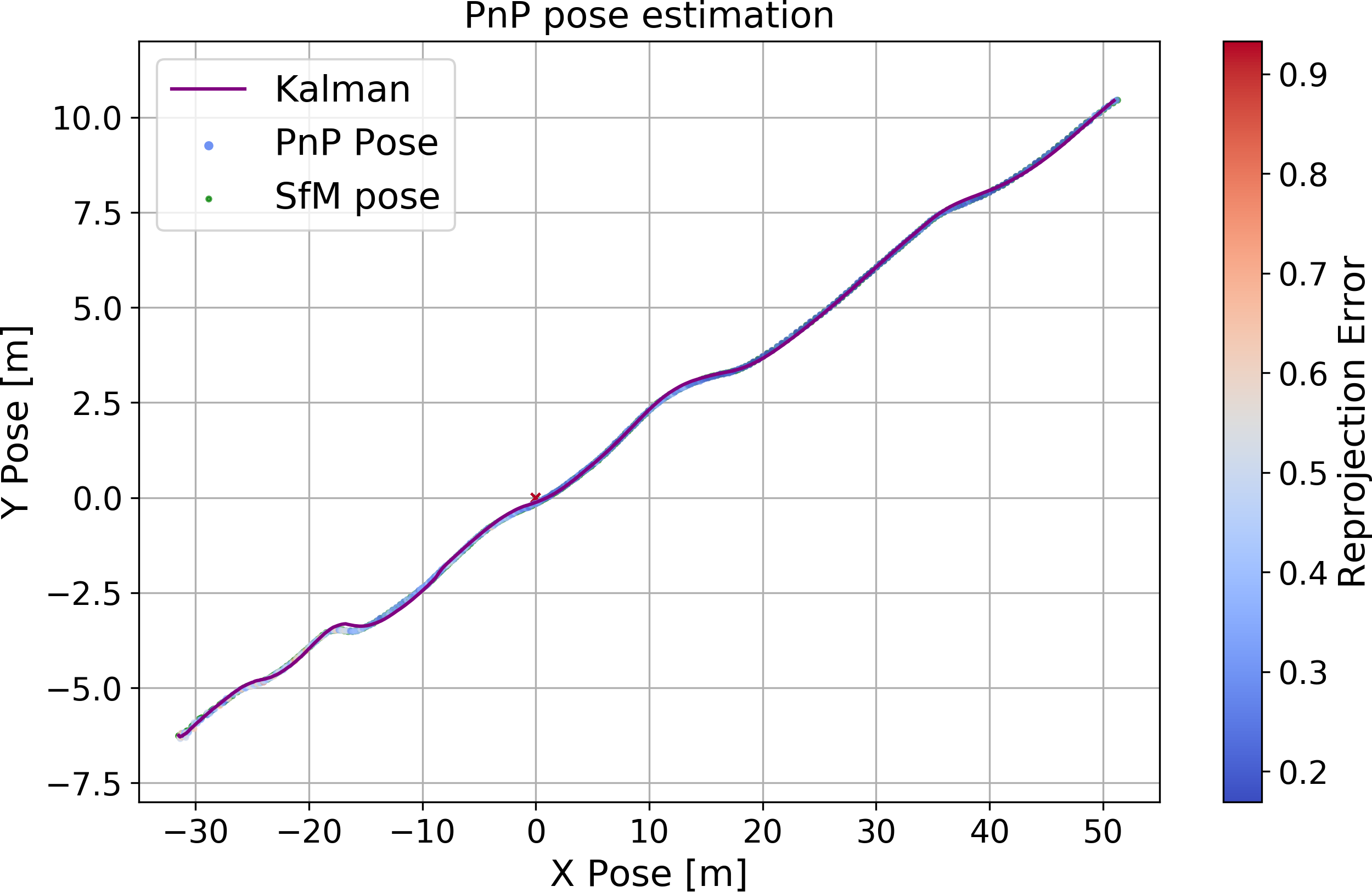}
  \caption{}
  \label{fig:SyoloXY_PPB}
\end{subfigure}\hspace{3mm}%
\begin{subfigure}[t]{.41\textwidth}
  \centering
  \includegraphics[width=1\textwidth]{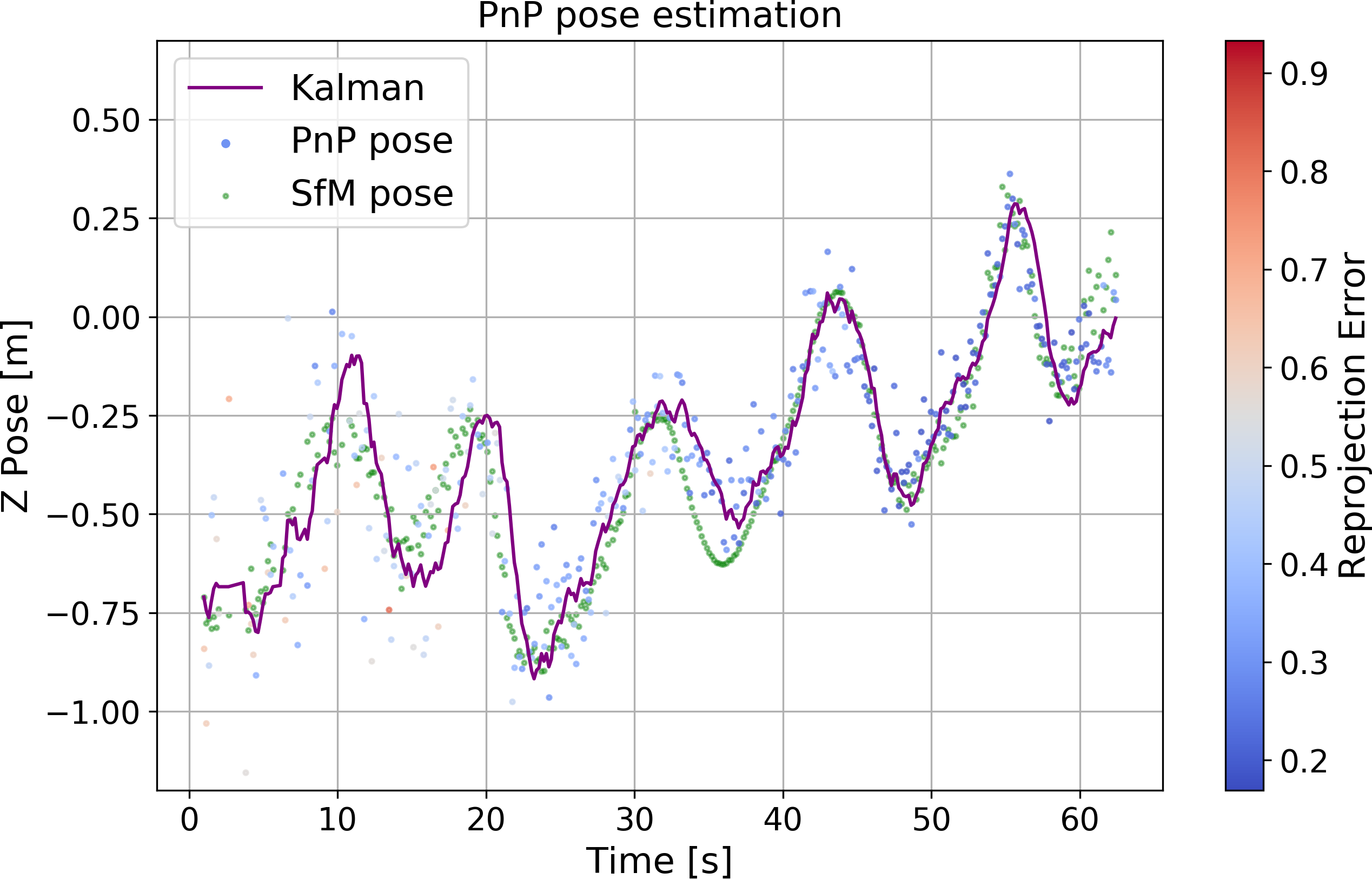}
  \caption{}
  \label{fig:SyoloZ_PPB}
\end{subfigure}

\caption{PPB, \textit{SfM} model. The reprojection error is  illustrated using a color map. Figures \textbf{(a)} and \textbf{(b)} present results of the edge-based method. Figures \textbf{(c)} and \textbf{(d)} present results for U-Net. Figures \textbf{(e)} and \textbf{(f)} present results for YOLO.}
\label{fig:Saxes_repr_SfM_PPB}
\end{figure}

\end{document}